\documentclass[twoside,11pt]{article}
\pdfoutput=1

\usepackage{amsfonts}       
\usepackage{amsmath, amssymb, amsthm}
\usepackage{natbib}
\usepackage{stackengine} 
\stackMath

\usepackage[preprint]{jmlr2e}

\newtheorem*{theorem*}{Theorem}
\newtheorem{property}{Property}

\usepackage{lastpage}

\usepackage{tikz}
\usepackage{pgfplots}\pgfplotsset{compat=1.16}
\usetikzlibrary{arrows.meta,backgrounds,calc,decorations.pathreplacing,fit,patterns,positioning,shapes.geometric,shapes.symbols,shapes.misc}
\usepgfplotslibrary{groupplots}
\usepackage{adjustbox}
\usepackage{etoolbox}
\BeforeBeginEnvironment{proof}{\medskip}
\AfterEndEnvironment{proof}{\medskip}
\definecolor{accent}{RGB}{213,94,0}
\definecolor{cool}{RGB}{0,114,178}
\definecolor{muted}{RGB}{95,95,95}
\definecolor{faint}{RGB}{175,175,175}
\jmlrheading{23}{2022}{1-\pageref{LastPage}}{1/21; Revised 5/22}{9/22}{21-0000}{M. P. Karpowicz}

\ShortHeadings{Fundamental Impossibility of Hallucination Control}{M.P. Karpowicz}
\firstpageno{1}

\begin{document}
\setlength{\emergencystretch}{3em}

\title{On the Fundamental Impossibility of Hallucination Control\\
in Large Language Models}

\author{\name Michal P. Karpowicz \email m.karpowicz@samsung.com \\
       \addr Samsung AI Center Warsaw\\
       Warsaw, Poland}

\editor{My editor}

\maketitle

\begin{abstract}

Large language models hallucinate. This paper shows when that is unavoidable and what we can do about it. We model inference as an \emph{auction of ideas}, in which a model's components, each holding partial knowledge, compete to shape the answer. We then prove Impossibility Theorems showing that whenever a query makes LLM components contest a fact they hold in common, no aggregation of their reports can at once report that knowledge truthfully, avoid manufacturing confidence beyond what it supports, keep the relevant components engaged, and give the best answer. Something must give, and each failure is familiar: a fabricated detail, unearned confidence, ignored knowledge, or a needlessly weak reply. This is no artifact of one design. It reappears when components report probabilities, and inside the transformer itself, where the combined answer is credited more confidence than the internal contributions supplied.

The unbalanced semantic budget cannot be settled from within. Factual truth lies outside the model, and in the worst case no internal signal can certify it. What can be certified is support. Given externally authorized evidence, checking that an answer stays within what the evidence entails needs only the answer and the evidence, and we prove when that check is computable. However, a correct answer can lack support, and a supported answer can be false. What counts as evidence, how far beyond it we allow answers to reach, and which failures we can live with are choices no model can make for us. 

\end{abstract}



\section{Introduction}

Large language models have been famous for producing content that is factually incorrect, inconsistent, fabricated and yet compelling and immensely impressive---a phenomenon known as hallucination. Despite extensive research efforts involving architectural innovations, like mixture-of-experts (MoE), retrieval augmented generation (RAG), training techniques with external feedback, chain-of-thought (CoT) reasoning, and post-hoc verification methods, hallucination persists across all state-of-the-art models.

The idea of hallucination has much to do with creativity and depends on circumstances. We could argue that a much better name for the phenomenon we are dealing with may be \textit{imagination}. As an essential component of intelligence, imagination is responsible for seeing into the future and learning from the past. It is the imagination that creates the scenarios of what may happen and counterfactual scenarios of what could have happened. It is one of the driving forces of art and science, and exploration of the unknown. But, when uncontrolled, unconstrained and misunderstood, it may also generate against our will all the compelling visions that do not exist in any training set defining the ground truth for learning.

This paper shows that hallucination reflects a real limit of the inference process, but a \emph{conditional} one. There are four properties we would want a perfect answer to satisfy at once: \textbf{(1) truthful reporting, (2) semantic information conservation, (3) participation of every relevant knowledge source, and (4) knowledge-constrained optimality}. Our Impossibility Theorem shows these cannot be held together for every query, and in particular when the query forces the model's components into genuine competition over the same facts. Where that competition is absent, the obstruction lifts, paving the way to trustworthiness. The four properties are not arbitrary. They come from the game-theoretic study of information aggregation in multi-agent systems.

We prove it in three complementary settings, each lighting the problem differently. In the first, we picture inference as an \textbf{auction of ideas}, where knowledge-bearing components, such as attention heads, circuits, and activation patterns, bid to influence the answer. The probabilistic and transformer settings then use their own, separately stated accounting rules, and we do not claim the three are one universal mechanism in disguise. Establishing that stronger claim would take a representation theorem: a single quantity, defined independently of any one setting, whose calibrated images reproduce the mechanism transfers, the scoring losses, and the transformer's log-partition entries at once. We leave that bridge open.

The first setting is idealized. Knowledge is split cleanly across independent components, and the Green--Laffont theorem from mechanism design tells us that truthful, efficient aggregation cannot also balance its internal account once those components contest the same facts. The second is closer to how models are trained. Components emit probability distributions scored by a strictly convex loss, and there the obstruction returns through Jensen's inequality, when forecasts that disagree about the realized outcome are combined into sharper confidence than any single one of them holds. The third is the transformer itself, where summing the channels' logits leaves a measurable gap in that same account. Each setting carries its own hypotheses, and we do not silently carry one setting's conclusion over to another.

Seeing the same kind of obstruction recur from an abstract auction, through probabilistic scoring, to a concrete architecture is the paper's central evidence that it is structural rather than an artifact of any one modeling choice. It is not, however, a proof that every model or every query must fail.

\subsubsection*{Terminology Note}

This paper uses the term \textit{hallucination} in its established technical sense within artificial intelligence research, referring to AI systems generating outputs that are factually incorrect, inconsistent, or unsupported by their training data. We acknowledge that this technical usage differs from the clinical term describing perceptual experiences in certain medical and psychiatric conditions. We recognize and respect the serious challenges faced by individuals experiencing clinical hallucinations and their families. Our use of this terminology follows established conventions in machine learning literature and is not intended to trivialize or misrepresent these medical experiences.

\subsection{An Example}

Consider an LLM queried about recent diplomatic negotiations: 

\begin{center}
\textit{"What was the outcome of diplomatic negotiations between Countries A and B in February 2025?"}     
\end{center}

Assume the model has encoded knowledge that: (K1) countries A and B were engaged in talks, (K2) country A sought a trade agreement, and (K3) the actual outcome is unknown.

Examining possible responses reveals the fundamental trade-offs:

\textbf{Complete Abstention}: \textit{"I don't have information about the outcome"} is truthful but fails to utilize available relevant knowledge, violating the expectation that good inference should reveal accessible information.

\textbf{Hallucination}: \textit{"The negotiations resulted in a comprehensive trade agreement reducing tariffs by 15\%"} utilizes all knowledge and provides a complete and immediately satisfactory answer but fabricates specific details, violating truthfulness.

\textbf{Partial Knowledge}: \textit{"Countries A and B were negotiating a trade agreement, but I don't know the outcome"} is truthful and utilizes available knowledge but may appear irritating and unsatisfying compared to the decisive but hallucinated response (training objectives and human preferences favor complete narratives).

\textbf{Overcautious}: \textit{"I'm not certain about any aspects of these negotiations"} avoids hallucination but disclaims even known facts, violating truthfulness about available information.

Each response strategy sacrifices different aspects of what constitutes an ideal response, illustrating the impossibility of simultaneously optimizing all desired properties.

\subsection{The Main Result}

The example above illustrates the core insight of this paper. Any response to a query meeting one of the conditions identified below sacrifices some aspect of an ideal hallucination-free response. Hallucinations (or creative responses) manifest as violations of at least one of those aspects.

\begin{theorem*}[Impossibility Theorem, stated informally]
An LLM answers a query by aggregating the partial knowledge of its internal components. Whenever the query makes all of those components genuinely contest a relevant fact they hold in common, no aggregation can at once report that knowledge truthfully, avoid manufacturing confidence beyond what the components supplied, keep every relevant component willing to contribute, and deliver the best answer the available knowledge supports. Something must give, and each way of giving is a familiar face of hallucination or overcaution: a fabricated detail, unearned confidence, ignored knowledge, or a needlessly weak answer. In the probabilistic and transformer models of the same process the failure becomes measurable. The components' declared contributions cannot sum to zero, and the shortfall is fixed by the accounting itself, not by any one design.
\end{theorem*}

Each theorem has a switch that turns the trade-off on. In Theorem~\ref{thm:private-knowledge-impossibility} the switch is contested knowledge, at least one relevant fact that all sources hold and compete to assert (Definition~\ref{def:contested}). In Theorem~\ref{thm:proper-scoring-impossibility} it is disagreement about the realized outcome, two components assigning different probabilities to the token that actually occurred. These switches are not technicalities. In the models we exhibit, turning the switch off restores the balance, so each condition marks the working boundary of its obstruction. Theorem~\ref{thm:lse-impossibility} needs no disagreement switch under its declared accounting, since that account's gap is positive for every finite decomposition of the logits. Change the accounting and a switch reappears, as we discuss after the theorem. Each theorem also fixes an accounting rule, the declared list of what every component is charged and credited: mechanism-design transfers in Theorem~\ref{thm:private-knowledge-impossibility}, per-forecast losses in Theorem~\ref{thm:proper-scoring-impossibility}, per-channel log-losses in Theorem~\ref{thm:lse-impossibility}. The conclusions are statements about these declared accounts. Formal non-triviality (Definition~\ref{def:nontrivial}, $|\mathcal{R}_C(Q)|\ge 2$) remains a descriptive condition on the query, not the switch of any of the three theorems.

When a theorem's switch is on, no amount of training removes the tension, and the model must give something up. It may ignore knowledge it has, assert a fabricated detail, merely restate the question, or land on a lucky but unsupported guess that happens to be right. There is no free lunch here. When the condition does not hold, there is no such forced trade-off, and a system can sidestep the obstruction by keeping its components uncontested, keeping their forecasts in agreement about the outcome, or adopting a different, explicitly justified accounting rule.

Creative tasks put a friendlier face on the same mechanism. Asked for a novel analogy or a prediction beyond its data, a model must either admit it cannot (losing the optimality we ask of it) or reach past its evidence to offer something plausible (breaking strict conservation by generating new content in its internal representation). We treat that reaching-past as a deliberate widening of the authorized boundary, not as proof that all creativity is factual error.

It helps to see how the four properties play out during training and inference, through the auction-of-ideas lens. Training aims at \textbf{knowledge-constrained optimality} (Property \ref{prop:optimality} in Section \ref{sec:Theoretical framework}), minimizing a global loss such as cross-entropy so the model predicts the most satisfying continuation. To get there, training rewards \textbf{relevant-source participation} (Property \ref{prop:revelation}), pushing specialized components such as attention heads or mixture-of-experts to contribute when the context matches their specialization, which shows up at inference as strong activations and large logits. The tension is that, when such a condition holds, an optimal solution to the training problem can no longer also honor \textbf{truthful reporting} (Property \ref{prop:truthfulness}) and \textbf{semantic information conservation} (Property \ref{prop:conservation}). Truthfulness can suffer because minimizing the global loss may reward a component for shifting attention toward tokens that improve coherence over the tokens its local knowledge actually supports. Conservation can suffer because the aggregation step itself, the log-sum-exp inside softmax normalization, is strictly subadditive: combining channels that disagree about the realized outcome sharpens the final distribution beyond what the declared channels justify. That excess-confidence signature can accompany hallucination or imagination, but on its own it does not decide whether an answer is factually unsupported (the informal Impossibility Theorem above).

There is an important distinction we must emphasize. We define hallucination as a violation of idealized inference principles, expressed as excess confidence in a generative (confabulated) response. What the Impossibility Theorem establishes, under its stated conditions, is a measurable excess of confidence: the aggregate ends up more sure of its answer than its own components justify, by an amount that depends on how the model is divided into contributing channels. What the Impossibility Theorem does not prove, is that output must be incorrect relative to ground truth. That factual incorrectness may happen as a potential consequence of hallucination, but it does not have to happen. Excess confidence may as well accompany a factually correct lucky hallucination. In other words, under those conditions any truthful response revealing relevant knowledge to maximize user's satisfaction is provably non-conservative (in information creation) regardless of its veracity.

Despite its name, the Impossibility Theorem is best read as a constructive result. It maps the trade-offs precisely and tells a designer which lever to pull. Where its conditions hold, hallucination is not a bug to be patched away but a trade-off to be managed, and different applications will manage it differently, a medical system leaning toward truthfulness, a writing tool toward controlled imagination. Understanding the trade-off is what makes that choice principled.

\subsection{Paper Organization}

The paper starts with theoretical foundations and progresses through mathematical proofs and examples to philosophical implications.

Section \ref{sec:Related work} situates our contribution within the broader landscape of hallucination research, reviewing theoretical perspectives on inevitability, mechanistic interpretability studies, probabilistic frameworks, and empirical mitigation techniques. 

Section \ref{sec:Main Contributions} summarizes the key contributions: the auction-theoretic model of inference, the semantic information framework with computational bounds, the three conditional obstructions, and the constructive grounding result.

Section \ref{sec:Theoretical framework} constructs the mathematical framework. It formalizes knowledge as a Polish metric space, introduces the semantic information measure $\mu_C$ parameterized by computational budget, and defines the emergence operator $\mathcal{E}_C$ that captures how reasoning makes latent knowledge explicit. It establishes fundamental results about information preservation under unlimited reasoning and defines the auction-theoretic model where neural components compete as agents with private knowledge. The section concludes by defining the four essential properties of hallucination-free mechanisms.

Section \ref{sec:Impossibility Theorem} presents two of the three results. Theorem \ref{thm:private-knowledge-impossibility} treats the idealized private-knowledge mechanism, and Theorem \ref{thm:proper-scoring-impossibility} treats the stipulated probabilistic scoring account.

Section \ref{sec:Transformer} bridges theory to implementation, analyzing a one-block additive-logit model and a concrete micro-transformer example. We trace inference from embeddings through attention to final predictions, and Theorem \ref{thm:lse-impossibility} derives its positive, channelization-relative log-sum-exp gap.

Section \ref{sec:Safety and Alignment} examines bounded creativity and keeps evidence support separate from truth. Theorem \ref{thm:conservation-reasoning-dichotomy} shows that meaningful reasoning cannot also obey strict information conservation, so strict conservation is the wrong lever for safety. The section then gives a sufficient containment result and an effective support certificate under an externally authorized evidence closure.

Section \ref{sec:Discussion and Speculations} explores broader analogies and research directions, from hybrid architectures that supply external evidence to comparisons with G\"odel and Heisenberg and with Dennett's Multiple Drafts model of consciousness. These analogies are offered as speculation to inspire future work, not as premises or consequences of the formal theorems.

Section \ref{sec:Summary} synthesizes the journey from mathematical necessity to philosophical implications, outlining how our framework opens new avenues for understanding intelligence itself.

\section{Related Work}
\label{sec:Related work}

Studies on LLM hallucinations span many areas of the fundamental research in AI and beyond, including neurology and philosophy. For an overview, see e.g. Huang \cite{huang25hallsurvey}, Ji et al. \cite{Ji_2023}, Sun et al. \cite{Sun2024AIHT}, Waters et al. \cite{Waters2016WhatIT} or Boksa \cite{boksa2009neurobiology}. 

The conclusion that LLMs hallucinate is not new but the evaluation of their scope and underlying justifications differ significantly from what we propose in this paper. To our best knowledge, the Impossibility Theorem presented here offers a novel first-principles explanation for the phenomenon, distinct from existing approaches. This section situates our contribution by reviewing four key areas: theoretical arguments for the inevitability of hallucination, mechanistic studies into their origins, probabilistic frameworks for their detection, and empirical mitigation techniques.

\subsection{Theoretical Perspectives on Inevitability}

In their work, Banerjee et al.~\cite{banerjee2024llmshallucinateneedlive} argued from computational complexity principles that hallucinations are inherent features of LLMs stemming from their fundamental mathematical and logical structure. Their conclusion is that LLMs cannot detect and eliminate hallucinations by themselves and therefore must hallucinate. The analysis draws on Gödel’s incompleteness theorem and Turing's undecidability theorem. First, the incompleteness argument is invoked to show that an LLM cannot verify its responses, because its training set is incomplete. Second, the hallucination (self)detection problem is reduced to the halting problem, which is undecidable.

While the final conclusion aligns with that of our paper, the arguments there are more about the nature of hallucination detection with limited ground-truth access (see Section \ref{sec:Statements, Queries, and Ground Truth} for the ground-truth concept discussion) than the inference process limitations. In other words, it is about practical impossibility of real-time, exhaustive fact-checking. 

Similar arguments are presented in other works. Xu et al. \cite{Xu2024HallucinationII} define hallucination as inconsistency between a computable LLM and a computable ground truth function. Then the paper shows that LLMs cannot learn all the computable functions and will therefore inevitably hallucinate. Suzuki et al. \cite{suzuki2025hallucinations} show that hallucinations can be made statistically negligible, provided that the quality and quantity of the training data are sufficient. Then, Wu et al. show in \cite{Wu2024NoFL} that non-hallucinating learning is statistically impossible when relying solely on the training dataset.

Our results ask a different question. Not whether learning the truth from data is possible, but when, for a given model and query, the aggregation step itself cannot reconcile the properties we want.

\subsection{Mechanistic Interpretability}

Mechanistic interpretability examines internal LLM processes to identify emergence of ideas, interpretable concepts and dynamics of their interactions, including hallucination origins, see Sharkey et al. \cite{Sharkey2025OpenPI} for an overview. 

Yu et al.~\cite{yu2024mechanisticunderstandingmitigationlanguage} identified signatures in model internals that predict a likelihood of hallucinating. The paper identifies two failure modes: knowledge enrichment hallucinations from insufficient subject-attribute knowledge in lower MLP layers, and answer extraction hallucinations from failures in upper-layer attention heads selecting correct object attributes.

Studies characterizing factual knowledge storage and retrieval complement our framework by providing empirical evidence of competing agents within LLM architectures. Meng et al.~\cite{meng2023locatingeditingfactualassociations} demonstrated methods for locating and editing factual associations, revealing how knowledge representations can be manipulated. Geva et al.~\cite{geva2023dissecting} identified a three-step factual recall process: (1) enriching last-subject position representations with subject-related attributes, (2) propagating relation information to predictions, and (3) querying enriched subjects to extract attributes. More recently, Gottesman and Geva \cite{Gottesman2024EstimatingKI} have proposed a method to estimate how knowledgeable a model is about a certain entity and identify tokens indicative of clusters and gaps in the model's knowledge.

These findings provide concrete instantiations of our abstract agents--attention heads and MLP components with specialized knowledge domains competing to influence response generation. We must refer here to the fascinating and related work of Lindsey et al. \cite{lindsey2025biology}  on the biology of a Large Language Model. It is an investigation of the internal chains of steps that a model uses to transform a specific input prompt into an output response. In particular, the paper discovers how LLM plans its outputs ahead of time when writing lines of poetry, identifying in advance potential rhyming words that could appear at the end. We can view that as an empirical illustration of our auction of ideas.

\subsection{Probabilistic Frameworks}

Probabilistic approaches model and asses hallucination through uncertainty and belief quantification. Shelmanov et al.~\cite{shelmanov2025headpredictheadquestion} introduced pre-trained uncertainty quantification heads for detecting hallucinations without task-specific data. 

Farquhar et al.~\cite{farquhar2024detecting} proposed semantic entropy measures that detect hallucinations by quantifying uncertainty at the meaning level rather than word sequences. That concept of semantic entropy is well aligned and complementary to our idea of semantic information measure and semantic independence, pointing out the importance of understanding the meaning of competing ideas in the studies of imagination.

Most notably, Kalai and Vempala~\cite{kalai2024calibrated} proved that a calibrated model must hallucinate at a rate no smaller than the fraction of facts it saw exactly once in training. That is a statistical lower bound on how much hallucination to expect. Our results are of a different kind, neither stronger nor weaker in general. They are conditional obstructions in the aggregation step, saying \textit{when} certain properties cannot coexist, rather than a distribution-free floor on factual error.

In more recent work, \cite{kalai2025why} expand the theory of hallucination by connecting it with the computational learning theory. They demonstrate that any language model can be viewed as an IIV-type (Is-It-Valid) classifier trained on a set of responses labeled either as valid or error. Then, it follows that the statistical objectives of pretraining necessarily induce generative errors when the underlying data distribution is complex, characterized by poor model fit, or undersampled. This establishes that hallucination arises from the statistical impossibility of perfect learning. 

Our obstruction is structural rather than statistical, and the two can come apart. Even if the learning problem were solved perfectly, with $\min_c|\mathcal E_c| > 0$ keeping the classification problem well defined, so that no statistical floor forced any error, a model's knowledge may still be spread across its internal states, with different attention heads encoding different facets of the truth. Aggregating those states is exactly where our conditions can bite. The reverse caution matters just as much: distributed knowledge does not by itself mean the components disagree, contest the same facts, or induce a nonzero interaction. Those are the theorems' switches, and they must be checked for the model and query at hand rather than assumed.

The result closest to our grounding discussion is a boundary one. Automated hallucination detection is provably impossible when the detector can learn only from correct examples, and becomes possible once expert-labeled counterexamples are supplied \citep{karbasi2025impossibility}. That does not stop internal signals from being predictive on average \citep{mason2026epistemic}, and our support theorem takes the complementary route, asking what becomes certifiable once an external body of evidence is supplied. The separation it formalizes echoes the established distinction between faithfulness to a source and factual correctness in natural language generation \citep{maynez2020faithfulness,rashkin2023measuring}.

\subsection{Empirical Mitigation Techniques}

Various empirical approaches attempt to reduce hallucination frequency with mixed success. For a recent survey see e.g. Tonmoy et al \cite{Tonmoy2024ACS}. Here we refer only to selected works.

Retrieval-augmented generation (RAG) by Lewis et al.~\cite{lewis2021retrievalaugmentedgenerationknowledgeintensivenlp} brings in external sources for grounding, though retrieval quality and generation faithfulness remain failure modes. RLHF \cite{ouyang2022traininglanguagemodelsfollow} and related methods lower measured hallucination rates in the settings they are evaluated on, but they do not certify factual correctness. TruthfulQA \cite{lin2022truthfulqameasuringmodelsmimic} and HaluEval \cite{li2023haluevallargescalehallucinationevaluation} document persistent errors even in strong systems, and Dubanowska et al. \cite{dubanowska2025representation} show that internal-representation detectors can lean on spurious correlations.

\section{Main Contributions}
\label{sec:Main Contributions}

This paper expands our understanding of neural inference and its limits through the following contributions.

\textbf{The Auction of Ideas:} We formalize one idealized inference model as a mechanism-design problem, in which knowledge-bearing components such as attention heads, circuits, or activation patterns compete to influence the answer. This perspective lets us bring powerful game-theoretic results to bear, and to our knowledge it is among the first uses of mechanism design in this setting. It also resembles the multiple-drafts model of consciousness proposed by Dennett \cite{Dennett1991-DENCE}, which we offer as a suggestive analogy rather than a mathematical validation of that theory.

\textbf{Semantic Information Theory}: We introduce a rigorous mathematical framework based on two complementary concepts. The semantic information measure $\mu_C$, parameterized by computational budget $C$, quantifies how knowledge reduces uncertainty within computationally bounded reasoning. The emergence operator $\mathcal{E}_C$  formalizes how reasoning makes latent knowledge explicit without creating it \textit{ex nihilo}. In that framework we establish the principle of information conservation: unlimited reasoning reveals but never creates information. That also establishes a dichotomy of reasoning and semantic information conservation.

\textbf{The Impossibility Theorem, three ways}: We prove three differently scoped results. In Theorem \ref{thm:private-knowledge-impossibility}, an idealized mechanism whose components hold independent private knowledge cannot be truthful, conserving, participatory, and optimal at once, once its components compete over the same facts (Definition~\ref{def:contested}). In Theorem \ref{thm:proper-scoring-impossibility}, components report probability forecasts scored by a proper loss, and their contributions cannot balance whenever the forecasts disagree about the outcome that actually occurred: the imbalance is exactly the extra confidence the aggregate shows over its parts. In Theorem \ref{thm:lse-impossibility}, adding a transformer's channel logits always leaves such an imbalance, whatever the channels do. Separately, Theorem \ref{thm:conservation-reasoning-dichotomy} shows that reasoning which makes any non-negligible new content accessible must break strict information conservation.

\textbf{Certifiable Grounding}: Give the model an external body of authorized evidence and ask a different question than truth: does the answer stay within what that evidence entails? Unlike truth, this question is settled by the answer and the evidence alone, and it becomes mechanically checkable under explicit decidability conditions. Any certifier that answers it for every possible claim must in effect know everything the evidence entails. Support and truth stay deliberately separate throughout: a supported answer can be false, and a true answer can be unsupported.

\textbf{Excess confidence is bookkeeping, not a verdict}: The transformer's confidence gap splits into two parts, a floor that grows with the number of declared channels and with how spread out the aggregate remains, and a term measuring how much those channels disagree. Redraw the channel boundaries and the gap changes while the model's output does not. The gap is therefore a diagnostic of our own accounting, useful for analysis, and never by itself a verdict on whether an answer is factually unsupported.

\textbf{Connections to fundamental limits}: We place the results alongside G\"odel's incompleteness, Heisenberg's uncertainty, and Arrow's impossibility. These comparisons are analogies for orientation, discussed separately from what we actually prove.

\textbf{Experimental philosophy}: The framework transforms LLMs into accessible (philosophical) laboratories for testing hypotheses about the mind. By providing precise mathematical definitions of concepts like knowledge accessibility, semantic conservation, and cognitive hunger (information contribution), we enable rigorous experimental investigation of many claims.

\textbf{Theoretical foundations}: We build reusable tools for analyzing inference. The semantic information measure quantifies how much knowledge a model can actually use within a budget, and the emergence operator captures how reasoning turns latent knowledge into accessible knowledge. These tools state falsifiable conditions under which the aggregation trade-offs appear. They do not claim that every architecture or every query realizes those conditions, which is where earlier work, identifying statistical or architectural correlates of hallucination, connects to ours.

We should point out that empirical validation of the formal results is beyond the scope of this paper. However, we provide toy examples that can be expanded in future research on hallucination detection and artificial creativity. This empirical work is crucial for translating the theoretical insights into practical tools. See Dubanowska et al. \cite{dubanowska2025representation} for initial results on hallucination detection, and Brzozowski \cite{brzozowski2025bimodality} for initial results on the feature dictionaries through which the internal agents of our framework could be isolated in practice.

\section{Theoretical Framework}
\label{sec:Theoretical framework}

This section develops the mathematical foundations necessary for proving the Impossibility Theorem. It formalizes key concepts including knowledge representation, semantic information, and the auction-theoretic model of inference. Also, we address some philosophical aspects of the reasoning problem, relevant to our study, its generality, scope and consequences for the AI development.

\subsection{Knowledge and Semantic Information}
\label{sec:Knowledge and Semantic Information}

If we define knowledge as organized information that AI models can encode and utilize for reasoning, then, much like low-entropy energy in physics, it becomes a quantity used to perform the work of reasoning. Information, in turn, represents the potential to reduce uncertainty in a given context. This pragmatist perspective treats knowledge as a space of organized patterns accessible to computational transformations. This is the perspective we will adopt, and we will now present formal definitions of these ideas.

\subsubsection{Knowledge lives in Polish space}

To describe knowledge in a general way, one that is well suited for our study of inference processes, we need a structure with properties reflecting the reality we describe. First, for modeling and technical reasons, we need a way to measure distance between any pair of elements of knowledge. That is provided by a complete metric space. It is also reasonable to demand that each element of knowledge has a neighborhood consisting of countably many knowledge elements that are arbitrary close, to make exploration or understanding of ideas possible. Therefore, the space we need must be complete to guarantee the convergence of continuous transformations (inherent in neural networks) or to support the existence of limits required to model the stabilization of reasoning processes. 

In other words, we need a complete, separable metric space, i.e., a Polish space \cite{kuratowski1966wstep}. 

\begin{definition}[Polish space]
A topological space is called Polish if it is a complete metric space that has a countable dense subset. 
\end{definition}

That space is sufficiently general to include Banach spaces, Euclidean distance, or continuous mappings, and provides the necessary topological properties for developing advanced (probability) measure theory. It is also relevant to reasoning processes that we are going to describe as computational transformations. Throughout, $(\mathcal{K}, d_{\mathcal{K}})$ denotes a Polish (complete, separable metric) space.

Reasoning over a subset of knowledge generates another (possibly new) subset of knowledge. That requires operations in spaces that are closed under set-theoretical transformations, including countable unions and intersections. Borel $\sigma$-algebras of open sets of $\mathcal{K}$, denoted $\mathcal{B}(\mathcal{K})$, support development of such consistent knowledge generation and measurement processes. Then, the necessary closure properties for modeling reasoning are provided by the family $\mathcal{A}(\mathcal{K})$ of analytic subsets of $\mathcal{K}$. Analytic subsets of Polish spaces are continuous images of Borel subsets, closed under countable unions and intersections. 

Finally, to define a measure of information provided by any subset of knowledge, we utilize the $\sigma$-algebra of universally measurable sets, denoted $\mathcal{U}(\mathcal{K})$, i.e., the $\sigma$-algebra of sets measurable w.r.t. every Borel probability measure on $\mathcal{K}$.

It follows from the descriptive set theory that $
\mathcal{B}(\mathcal{K}) \subset \mathcal{A}(\mathcal{K}) \subset \mathcal{U}(\mathcal{K})$, see e.g. \cite{Kechris1995}. Analytic sets are closed under countable unions, Borel-measurable images, and intersections with Borel sets. Arbitrary intersections of analytic sets need not be analytic in general, but analytic sets are universally measurable. Differences of analytic sets need not be analytic, but they are universally measurable, as well.

These properties ensure that reasoning operations are well-defined and remain within a measurable domain. That is the structure rich enough for our needs. To justify that choice, let us investigate limitations of simpler structures.

\begin{example}
Let $X$ be a connected subset of a Polish space. Consider a continuous mapping $\Phi\colon X\mapsto\mathcal K$ modeling inference based on self-attention and MLP.

If $\mathcal K$ is finite, and hence discrete, then by continuity the image $\Phi(X)$ is connected in a discrete space. That means it must be constant. As a result, any nontrivial continuous dependence on embeddings is impossible if $\mathcal K$ is finite. A Polish subspace removes this problem.
\end{example}

\begin{figure}[htbp]
\centering
\adjustbox{max width=\linewidth}{%
\begin{tikzpicture}[font=\small,>={Stealth[length=2.4mm]}]
  \node[muted,font=\small\bfseries] at (1.55,3.55) {$\mathcal K$ finite (discrete)};
  \node[draw=muted,fill=black!5,ellipse,minimum width=1.5cm,minimum height=2.1cm] (X1) at (0.15,1.7) {$X$};
  \foreach \y in {0.5,1.1,1.7,2.3,2.9} \fill[muted] (3.0,\y) circle (1.9pt);
  \fill[accent] (3.0,1.7) circle (3.4pt);
  \draw[->,muted,thick] (X1.east) -- node[above,muted] {$\Phi$} (2.78,1.7);
  \draw[accent] (2.86,1.56) -- (2.05,0.5);
  \node[accent,anchor=north,align=center,font=\footnotesize] at (1.7,0.35) {image collapses to\\one point (constant)};

  \draw[faint,dashed] (4.75,-0.55) -- (4.75,3.75);

  \node[muted,font=\small\bfseries] at (7.75,3.55) {$\mathcal K$ Polish ($\mathbb R^d$)};
  \node[draw=muted,fill=black!5,ellipse,minimum width=1.5cm,minimum height=2.1cm] (X2) at (5.65,1.7) {$X$};
  \fill[cool!12,rounded corners] (7.55,0.5) rectangle (9.55,2.9);
  \draw[cool,thick] (7.7,0.8) .. controls (8.15,2.7) and (8.7,0.6) .. (9.4,2.5);
  \draw[->,muted,thick] (X2.east) -- node[above,muted] {$\Phi$} (7.5,1.7);
  \node[cool,anchor=north,align=center,font=\footnotesize] at (8.55,0.0) {connected image\\survives};
\end{tikzpicture}
}
\caption{On the left the continuous map $\Phi$ sends the connected domain $X$ into a finite, discrete $\mathcal K$, so the image must collapse to a single point and the output is constant. On the right the same map into a Polish space $\mathbb R^d$ keeps a connected, genuinely varying image.}
\label{fig:ex-finite-continuity}
\end{figure}

\begin{example}
For any two probability distributions $\pi^{(1)},\pi^{(2)}\in\Delta^\infty$ (over countable dictionaries) and $\beta\in(0,1)$, the convex aggregate $\Pi=\beta\pi^{(1)}+(1-\beta)\pi^{(2)}$ defines the line segment $\{\beta\pi^{(1)}+(1-\beta)\pi^{(2)}:\beta\in[0,1]\}$. That is  an uncountable subset of the Polish space. Therefore, knowledge space closed under such aggregation cannot be finite. In contrast, it resides naturally in a Polish domain.
\end{example}

\begin{figure}[htbp]
\centering
\adjustbox{max width=\linewidth}{%
\begin{tikzpicture}[font=\small,>={Stealth[length=2.4mm]}]
  \fill[black!4] (0,0) -- (5,0) -- (2.5,3.5) -- cycle;
  \draw[muted] (0,0) -- (5,0) -- (2.5,3.5) -- cycle;
  \node[muted,anchor=south] at (2.5,3.55) {$\Delta^\infty$};

  \coordinate (p1) at (1.3,0.75);   
  \coordinate (p2) at (3.8,2.25);   

  \draw[accent,line width=1.6pt] (p1) -- (p2);

  \foreach \b/\lab in {0/{$\beta{=}0$},0.25/{$\tfrac14$},0.75/{$\tfrac34$},1/{$\beta{=}1$}}{
    \coordinate (t) at ($ (p2)!\b!(p1) $);
    \draw[accent,line width=1.1pt] ($ (t)!2.4pt!90:(p1) $) -- ($ (t)!2.4pt!-90:(p1) $);
  }
  \fill[black] (p1) circle (1.7pt);
  \fill[black] (p2) circle (1.7pt);
  \node[anchor=east,font=\footnotesize] at ($ (p1)+(1.8,-0.1) $) {$\pi^{(1)}\,(\beta{=}1)$};
  \node[anchor=west,font=\footnotesize] at ($ (p2)+(0.08,0.04) $) {$\pi^{(2)}\,(\beta{=}0)$};

  \coordinate (mid) at ($ (p1)!0.5!(p2) $);
  \fill[accent] (mid) circle (2.6pt);
  \node[accent,anchor=north west,font=\footnotesize] at ($ (mid)+(0.12,-0.10) $) {$\Pi\ (\beta{=}\tfrac12)$};

  \node[muted,anchor=north,align=center,font=\footnotesize] at (2.5,-0.35)
    {uncountably many aggregates $\Pi=\beta\pi^{(1)}+(1-\beta)\pi^{(2)}$};
  \node[muted,anchor=north,align=center,font=\footnotesize] at (2.5,-1.05)
    {a finite space cannot hold the whole segment};
\end{tikzpicture}
}
\caption{Blending two token distributions $\pi^{(1)}$ and $\pi^{(2)}$ by $\Pi=\beta\pi^{(1)}+(1-\beta)\pi^{(2)}$ sweeps out the whole highlighted segment as $\beta$ runs over $[0,1]$, an uncountable family of aggregates. No finite knowledge space can hold it, so the space must be infinite and Polish.}
\label{fig:ex-convex-segment}
\end{figure}

We now formalize the knowledge space and the reasoning transformations acting upon~it.

\begin{definition}[Knowledge Domain and Relevance]
\label{def:polish-knowledge}
Knowledge is a Polish space $(\mathcal{K}, d_{\mathcal{K}})$ whose admissible reasoning sets are restricted to the analytic subsets $\mathcal{A}(\mathcal{K})$.
\end{definition}

Both requirements in this definition earn their place. A Polish space is complete and separable at once. Completeness means that some metric generating the topology has all its Cauchy limits inside the space, so approximation arguments never fall off an edge. Separability means a countable set of points is dense, so the whole space is reachable from countably many landmarks. Familiar examples include $\mathbb{R}^d$, the sequence space $\ell^2$, the continuous functions $C[0,1]$, and the Cantor space $2^{\mathbb{N}}$. The rationals fail completeness and $\ell^{\infty}$ fails separability, so neither is Polish.

Above the space sits a tower of set families, $\mathcal{B}\subset\mathcal{A}\subset\mathcal{U}\subset 2^{\mathcal{K}}$, and each tier is forced on us by something the framework must do. The Borel sets $\mathcal{B}$ are everything countable unions, intersections, and complements can build from simple regions. Reasoning, however, also takes images, and the image of a Borel set, its shadow under a projection, need not be Borel. The analytic sets $\mathcal{A}$ are exactly what admitting those shadows adds, which is why the admissible reasoning sets live there. Measuring information forces one more step, since a measure needs every set it meets to carry a well-defined size, and the universally measurable sets $\mathcal{U}$ provide that home for $\mu_C$. Beyond $\mathcal{U}$ lie pathological sets with no consistent size at all, which only the Axiom of Choice can conjure, and the framework never builds them. On an infinite space each inclusion in the tower is strict.

It is natural to think of the knowledge space as the space of embeddings with additional measure and mappings. For model $\mathcal M$ with parameters trained on data, $\mathcal{K}_M \subset \mathcal{K}$ represents the knowledge subset encoded in the model's parameters. The distance function $d_{\mathcal{K}}$ measures differences between knowledge elements. The elements of the knowledge space represent atomic facts, relational structures, and transformation rules at many scales. We need a continuous, complete metric knowledge space. Without that requirement, we will not be able to model the internal numerical process by which the model reasons and transforms the input symbols into output symbols.

We make an important (ontological) assumption that calls for attention and explanation. We assume that knowledge is closed under transformations available in computable function space. That suggests we are merely recombining subsets of knowledge. As we will see in the following sections, that intuition is misleading since the reasoning processes are very much capable of making novel insights in that setting. 

\subsubsection{Reasoning envelope}

We model the reasoning process using a set of computational primitives. These create a reasoning envelope and define the scope of our inference studies. Each primitive maps a set of premises to the set it derives within a computational budget, so that multi-premise inference such as modus ponens is expressible. A point map $f\colon\mathcal{K}\to\mathcal{K}$ could not do this, since it cannot combine $\{P\}$ and $\{P\to R\}$ into $\{R\}$.

\begin{definition}[Reasoning Envelope]
\label{def:FC-polish}
Let $\mathcal{O}(\mathcal{K})$ denote the collection of \emph{finitary, monotone, analyticity-preserving} maps $f\colon \mathcal{A}(\mathcal{K}) \to \mathcal{A}(\mathcal{K})$, where \emph{finitary} means $f(A)=\bigcup\{f(A_0):A_0\subseteq A\ \text{finite}\}$, \emph{monotone} means $A\subseteq B\Rightarrow f(A)\subseteq f(B)$, and \emph{analyticity-preserving} means $f(A)\in\mathcal{A}(\mathcal{K})$. For each computational budget $C>0$, let
\begin{align}
\mathcal{F}_C = \bigl\{\, f\in\mathcal{O}(\mathcal{K}) \ \big|\ \text{cost}(f)\le C \,\bigr\}
\end{align}
be a \emph{countable} family of budget-$C$ reasoning operators. Assume composition-closure with cost additivity, so that $f\in\mathcal F_C$ and $g\in\mathcal F_D$ imply $g\circ f\in \mathcal F_{C+D}$, together with the nesting $\mathcal{F}_C \subseteq \mathcal{F}_D$ for $C < D$. The total computability envelope is
\begin{align}
\mathcal{F}_\infty = \bigcup_{C\ge 0} \mathcal{F}_C.
\end{align}
\end{definition}

The nesting $\mathcal{F}_C\subseteq\mathcal{F}_D$ makes $\mathbb{N}$ cofinal in $[0,\infty)$, so $\mathcal{F}_\infty=\bigcup_{m\in\mathbb{N}}\mathcal{F}_m$ is a countable union of countable families and is therefore countable. The analytic-closure and $C=\infty$ fixed-point arguments below rely on this countability, which also makes each $\mathcal{E}_C$ a countable union of analytic sets and hence analytic.

Here $C$ bounds the cost of each admitted operator application, not the total cost of a chain of them. The fixed point $A_C^*$ built below closes under arbitrarily many applications, so it does not impose a cumulative budget of $C$. A total-budget convention would instead truncate to compositions whose summed costs stay within $C$. We use the per-step convention throughout, relying on $g\circ f\in\mathcal F_{C+D}$ from the definition, and not on any claim that $\mathcal{F}_C$ is closed under composition at the same budget, which would be false in general.

Because every $f \in \mathcal{F}_C$ is analyticity-preserving by definition, the image $f(A)$ of any analytic knowledge subset $A \in \mathcal{A}(\mathcal{K})$ is again analytic. This guarantees that the reasoning process maps analytic sets to analytic sets, as desired. Analyticity of the relevance region $\mathcal{R}_C(Q)$ is a separate matter, established below under the measurable-incidence assumption. We do not need a measurable structure on the space of all analytic subsets for either.

Examples of computational cost (or complexity), $\text{cost}(f)$, include the Kolmogorov complexity \cite{kolmogorov1965three} giving the shortest program computing $f$, the longest path in computation graph (or length of the chain of thoughts CoT), a number of inference rules applied, or a number of QKV matrix multiplications in transformers. 

\begin{example}
Consider a reasoning primitive calculating a linear combination of any pair of vectors,  $f\colon\{v_1,v_2\}\to \{y(t):t\in\mathbb R\}$. For $t\in \mathbb{R}$ and attention score $\alpha(t) = 1/(1+e^{-t})$,  we then have
\begin{align}
y(t)=\alpha(t) v_1 + \bigl(1-\alpha(t)\bigr) v_2.
\end{align}
A finite knowledge space $\mathcal K$ cannot contain all such $y(t)$ while remaining closed under the transform $f$. 

By contrast, taking $\mathcal K$ to be a Polish space that contains $y(t)$, e.g.,  $\mathbb R^d$ with Euclidean metric, accommodates such attention outputs.
\end{example}

\begin{figure}[htbp]
\centering
\adjustbox{max width=\linewidth}{%
\begin{tikzpicture}[font=\small,>={Stealth[length=2.4mm]}]
  \fill[black!3,rounded corners] (0,0) rectangle (4.7,3.0);
  \node[muted,anchor=south east,font=\footnotesize] at (4.58,0.12) {$\mathbb R^d$};

  \coordinate (v1) at (0.7,0.7);
  \coordinate (v2) at (4.0,2.4);
  \draw[accent!30,line width=2.4pt] (v1) -- (v2);
  \fill[black] (v1) circle (1.9pt) node[anchor=north east,font=\footnotesize,black] {$v_1$};
  \fill[black] (v2) circle (1.9pt) node[anchor=west,font=\footnotesize,black,xshift=1pt] {$v_2$};

  \coordinate (y) at (1.69,1.21);
  \fill[accent] (y) circle (2.7pt);
  \draw[accent] (y) -- (2.25,2.9);
  \node[accent,anchor=south,font=\footnotesize] at (2.25,3)
    {$y(t)=\alpha(t)\,v_1+(1-\alpha(t))\,v_2$};

  \node[muted,anchor=north,align=center,font=\footnotesize] at (2.35,-0.28)
    {finite $\mathcal K$ cannot hold every $y(t)$};

  \begin{scope}[shift={(8.2,0.55)}]
    \draw[muted,->] (-1.2,0) -- (1.35,0) node[right,muted,font=\footnotesize] {$t$};
    \draw[muted,->] (-1.2,0) -- (-1.2,1.75) node[above,muted,font=\footnotesize] {$\alpha$};
    \node[muted,anchor=east,font=\footnotesize] at (-1.25,0) {$0$};
    \node[muted,anchor=east,font=\footnotesize] at (-1.25,1.5) {$1$};
    \draw[cool,line width=1.1pt] plot[domain=-5:5,samples=60]
      ({\x*0.22},{1.5/(1+exp(-\x))});
    \draw[accent,dashed] (-1.2,1.05) -- (0.187,1.05) -- (0.187,0);
    \node[accent,anchor=east,font=\footnotesize] at (-1.25,1.05) {$0.7$};
    \fill[accent] (0.187,1.05) circle (2.2pt);
    \node[muted,anchor=south,font=\footnotesize] at (0.1,1.8)
      {$\alpha(t)=\dfrac{1}{1+e^{-t}}$};
  \end{scope}
\end{tikzpicture}
}
\caption{The attention weight $\alpha(t)=1/(1+e^{-t})$ slides the output $y(t)=\alpha(t)v_1+(1-\alpha(t))v_2$ continuously along the segment between the two value vectors. Because $\alpha$ takes infinitely many values, only a Polish space such as $\mathbb R^d$ can hold every $y(t)$.}
\label{fig:ex-attention-slide}
\end{figure}

The following idea of computational independence allows for distinguishing between facts derivable from other facts and facts that cannot be derived that way. We need that distinction to deal with the mechanics of reasoning process. 

\begin{definition}[Computational Independence]
\label{def:computational-independence}
Knowledge sets $A, B \subseteq \mathcal{K}$ are {computationally independent} with respect to computational budget $C$, denoted $A \perp_C B$, if and only if for all $f\in \mathcal{F}_C$:
\begin{align}
f(A) \cap B = \emptyset \quad \text{and}\quad A \cap  f(B) = \emptyset.
\end{align}
\end{definition}

To gain some intuitions about $\mathcal{F}_C$ and independence, consider the set of all derivation rules in first-order logic, like \textit{modus ponens} or \textit{universal instantiation}, with at most $C$ inference steps. Then, given an axiom set $A \subset \mathcal{K}$, we can produce a new subset of knowledge with at most $C$ applications of the admitted inference rules. 

\begin{example}
Given 
\begin{align}
A = \{P, P \to Q\}\text{ and } B = \{R, R \to S\},
\end{align}
and the budget $C = 1$ for one inference step, we can derive 
\begin{align}
\{P, P \to Q, Q\} \text{ and } \{R, R \to S, S\}
\end{align}
with any $f \in \mathcal{F}_1$. Then $A$ and $B$ are independent, because $f(A) \cap B = \emptyset$ and $A \cap  f(B) = \emptyset$. There are no shared elements of knowledge accessible in 1-step derivations.    
\end{example}

\begin{figure}[htbp]
\centering
\adjustbox{max width=\linewidth}{%
\begin{tikzpicture}[font=\small,>={Stealth[length=2.4mm]},
  prem/.style={draw=muted,fill=black!4,rounded corners=2pt,inner sep=3pt,font=\footnotesize},
  deriv/.style={draw=cool,fill=cool!10,rounded corners=2pt,inner sep=3pt,font=\footnotesize,text=cool}]

  \draw[muted,rounded corners=6pt] (0.1,0.35) rectangle (3.7,2.95);
  \node[muted,anchor=north west,font=\small\bfseries] at (0.25,2.85) {$A$};
  \node[prem] (P)  at (0.95,2.05) {$P$};
  \node[prem] (PQ) at (0.95,1.05) {$P\to Q$};
  \draw[decorate,decoration={brace,amplitude=4pt,mirror},muted] (1.6,0.75) -- (1.6,2.35);
  \node[deriv] (Q) at (2.85,1.55) {$Q$};
  \draw[->,muted] (1.78,1.55) -- (Q.west);
  \node[muted,font=\footnotesize] at (2.35,2.15) {1 step};
  \node[muted,anchor=north,align=center,font=\footnotesize] at (1.9,0.2)
    {closure $\{P,\,P\to Q,\,Q\}$\\(one modus ponens step)};

  \draw[muted,rounded corners=6pt] (5.6,0.35) rectangle (9.2,2.95);
  \node[muted,anchor=north west,font=\small\bfseries] at (5.75,2.85) {$B$};
  \node[prem] (R)  at (6.45,2.05) {$R$};
  \node[prem] (RS) at (6.45,1.05) {$R\to S$};
  \draw[decorate,decoration={brace,amplitude=4pt,mirror},muted] (7.15,0.75) -- (7.15,2.35);
  \node[deriv] (S) at (8.35,1.55) {$S$};
  \draw[->,muted] (7.33,1.55) -- (S.west);
  \node[muted,font=\footnotesize] at (7.85,2.15) {1 step};
  \node[muted,anchor=north,align=center,font=\footnotesize] at (7.4,0.2)
    {closure $\{R,\,R\to S,\,S\}$\\(one modus ponens step)};

  \draw[faint,dashed,line width=0.9pt] (4.65,-0.65) -- (4.65,3.15);
  \node[accent,anchor=south,font=\footnotesize] at (4.65,3.2)
    {$f(A)\cap B=\varnothing,\quad A\cap f(B)=\varnothing$};
  \node[muted,anchor=north,font=\footnotesize] at (4.65,-0.7) {no shared knowledge};
\end{tikzpicture}
}
\caption{With a one-step budget, premise set $A$ derives only $Q$ and premise set $B$ derives only $S$. The two closures share nothing, so $f(A)\cap B=\varnothing$ and $A\cap f(B)=\varnothing$, and $A$ and $B$ are computationally independent.}
\label{fig:ex-comp-independence}
\end{figure}

Since reasoning is performed within a context, we define context-relevant subset of knowledge that can contribute to reasoning within some computational budget.

\begin{definition}[Context-Relevant Knowledge]
\label{def:context-relevant-knowledge}
The context-relevant knowledge $\mathcal{R}_C(Q)$ for context $Q \in \mathcal{B}(\mathcal{K})$ within computational budget $C > 0$ is:
\begin{align}
\mathcal{R}_C(Q) = \{k \in \mathcal{K} : \{k\} \not\perp_C Q\} \in \mathcal{A}(\mathcal{K}).
\end{align}
\end{definition}

Based on the definition above, we say that knowledge $\{k\}$ is relevant when
\begin{align}
f(\{k\})\cap Q \neq \emptyset\text{ or }
\{k\}\cap f(Q) \neq \emptyset.
\label{eq:relevance-criterion}
\end{align}
There is something we can derive from that knowledge that has something in common with a given context, or we can find a relation with what we know when we reason about that context. We assume \emph{measurable incidence}: for each $f\in\mathcal{F}_C$ the sets $\{k:f(\{k\})\cap Q\neq\emptyset\}$ and $\{k:\{k\}\cap f(Q)\neq\emptyset\}$ are analytic. Under this assumption $\mathcal{R}_C(Q)$ is analytic, as a countable union (over the countable family $\mathcal{F}_C$) of analytic incidence sets together with the Borel context $Q$. At unlimited budget we write $\mathcal{R}_\infty(Q)=\bigcup_{C\ge 0}\mathcal{R}_C(Q)$ for the relevance region reachable at some finite budget.

\subsubsection{The Semantic Information Measure}

We now introduce the concept of semantic information measure to describe a general process of reasoning. That special mapping measures information available in any subset of knowledge and extracts its meaning within available computational budget. 

\begin{definition}[Semantic Information Measure]
\label{def:semantic-information-measure}
The semantic information measure is a bounded mapping 
\begin{align}
\mu_C\colon\  \mathcal{U}(\mathcal{K}) \times \mathcal{B}(\mathcal{K}) \mapsto \mathbb{R}_+,
\end{align}
parameterized by computational budget $C > 0$ satisfying the following axioms:
\begin{align*}
\text{(null set)}&
&&\mu_C(\emptyset|Q) = 0 \text{ for all } Q \subseteq \mathcal{K},\\
\text{(contextual positivity)}&
&&\mu_C(A|Q) \geq 0,\ \text{with } \mu_C(A|Q) > 0 \text{ iff the relevant}\\
&&&\text{knowledge } A \text{ contributes within budget } C \text{ is non-null},\\
\text{(monotonicity)}&
&&\mu_C(A|Q) \leq \mu_C(B|Q) \text{ if } A \subseteq B,\\
\text{(conditional subadditivity)}&
&&\mu_C(A \cup B|Q) \le \mu_C(A|Q) + \mu_C(B|Q)\\
&&&\text{whenever } (A\cup B)^{*}_C \subseteq A^{*}_C \cup B^{*}_C,\\
\text{(insight monotonicity)}&
&&\mu_C(A|Q) \le \mu_D(A|Q) \text{ if } C < D,\\
\text{(boundedness)}&
&&\sup \{\mu_C(A|Q) : A\in \mathcal{U}(\mathcal K)\}<\infty.
\end{align*}
\end{definition}

Knowledge that cannot resolve any uncertainty or knowledge that is irrelevant to the given context provides no information. If there is an uncertainty resolving information within our cognitive reach, then we can use it and perform reasoning organizing that information into available knowledge. We assume information content is subadditive for independent knowledge sets. When no knowledge in $A$ can be derived from $B$ within computational budget $C$, and vice versa. With redundant information in $A\cup B$ , we admit potential loss of information that may be caused by aggregation. With increasing computational budget we can extract more knowledge (ordered information) from the same information subset.

In the absence of context, when $Q = \emptyset$, nothing is relevant so $\mu_C(A|\emptyset) = 0$. Knowledge provides zero semantic information, is useless, when there is no context defining the uncertainty to be resolved. That means we strongly emphasize the contextual relativity of information. Knowledge is only informative relative to a specific query or problem. 

We emphasize that the axioms require $\mu_C$ to be subadditive with respect to the knowledge argument $A$. Its behavior with respect to the context $Q$ is distinct and permits superadditivity, modeling how integrating contexts can unlock synergistic insights (see Example below).

The axiom of boundedness reflects the fundamental constraint that bounded reasoning systems can only access bounded information, even within an infinite knowledge space. The axiom does not contradict the potential infinity of the knowledge space itself. It asserts that the accessible information content derived from this space must remain finite. By the monotonicity axiom, this is also equivalent to stating that the measure of the entire knowledge space is finite, $\mu_C(\mathcal{K}|Q)<\infty$.

\subsubsection*{Justification for the Axioms of the Semantic Information Measure}

The choice of conditional subadditivity for computationally independent sets is necessary to describe isolated knowledge encoded in distinct neural circuits, attention heads, or activation patterns. The concept of independence is the formal translation of this isolation into a computational context. If two knowledge sets are truly independent, they offer no redundant or synergistic information. The subadditivity axiom becomes the baseline definition of non-interaction. 

However, there is also a form of synergy already included in the model. To see what it may be, we should understand how context $Q$ may affect the semantic measure. The semantic information measure can exhibit superadditivity in the context argument. Integrating different pieces of context may unlock reasoning paths not available individually.

\begin{example}
Consider the queried fact $A = \{R\}$ with the two context pieces $Q_1 = \{P\}$ and $Q_2 = \{P\to R\}$, and let $f\in\mathcal{F}_C$ (budget $C=2$) contain the identity and \textit{modus ponens} applied to sets. By the relevance criterion \eqref{eq:relevance-criterion}, $R$ is relevant to a context $Q$ exactly when $\{R\}\cap f(Q)\neq\emptyset$ for some $f\in\mathcal{F}_C$. Neither context piece alone makes $R$ relevant, but their union does:
\begin{align}
\{R\}\cap f(Q_1) = \emptyset,\quad
\{R\}\cap f(Q_2) = \emptyset,\quad
\{R\}\cap f(Q_1\cup Q_2) \neq \emptyset,
\end{align}
since $R$ is derivable only when both the premise $P$ and the rule $P\to R$ are present. Hence $R\in\mathcal{R}_C(Q_1\cup Q_2)$ while $R\notin\mathcal{R}_C(Q_1)\cup\mathcal{R}_C(Q_2)$, so $\mu_C(\{R\}|Q_1)=\mu_C(\{R\}|Q_2)=0$ yet $\mu_C(\{R\}|Q_1\cup Q_2)>0$. The integrated context thus unlocks a relevant fact available under neither piece alone. This illustrates that the semantic measure can be strictly \emph{superadditive} in the context argument,
\begin{align}
\mu_C(A|Q_1\cup Q_2) > \mu_C(A|Q_1) + \mu_C(A|Q_2),
\end{align}
whenever the union yields relevant conclusions (here $R$) that neither part yields alone.
\end{example}

\begin{figure}[htbp]
\centering
\adjustbox{max width=\linewidth}{%
\newcommand{\lockglyph}[3]{%
  \begin{scope}[shift={(#1,#2)},#3]
    \draw[line width=0.5pt] (-0.055,0.02) -- (-0.055,0.075) arc (180:0:0.055) -- (0.055,0.02);
    \fill (-0.085,-0.13) rectangle (0.085,0.02);
    \fill[white] (0,-0.05) circle (0.016);
  \end{scope}}
\newcommand{\openglyph}[3]{%
  \begin{scope}[shift={(#1,#2)},#3]
    \draw[line width=0.5pt,rotate around={42:(-0.055,0.02)}] (-0.055,0.02) -- (-0.055,0.075) arc (180:0:0.055) -- (0.055,0.02);
    \fill (-0.085,-0.13) rectangle (0.085,0.02);
    \fill[white] (0,-0.05) circle (0.016);
  \end{scope}}
\begin{tikzpicture}[font=\small,>={Stealth[length=2.4mm]}]
  \foreach \cx in {0,3.55,7.1}{
    \draw[faint,rounded corners,fill=black!2] (\cx-1.7,-0.95) rectangle (\cx+1.7,3.45);
  }
  \node[font=\small] at (0,3.05) {$Q_1$};
  \node[rounded corners,draw=muted,fill=white,inner sep=2.5pt,font=\footnotesize] (c1) at (0,2.5) {$P$};
  \node[rounded corners,draw=muted,fill=black!4,align=center,font=\footnotesize,inner sep=3pt] (m1) at (0,1.5) {modus ponens\\[1pt]{\footnotesize cannot fire}};
  \draw[->,faint] (c1) -- (m1);
  \node[circle,draw=muted,fill=black!10,text=muted,minimum size=0.85cm] (r1) at (0,0.35) {$R$};
  \draw[->,faint] (m1) -- (r1);
  \lockglyph{0.62}{0.35}{muted}
  \node[font=\footnotesize,muted] at (0,-0.55) {$\mu_C(\{R\}|Q_1)=0$};
  \node[font=\small] at (3.55,3.05) {$Q_2$};
  \node[rounded corners,draw=muted,fill=white,inner sep=2.5pt,font=\footnotesize] (c2) at (3.55,2.5) {$P\to R$};
  \node[rounded corners,draw=muted,fill=black!4,align=center,font=\footnotesize,inner sep=3pt] (m2) at (3.55,1.5) {modus ponens\\[1pt]{\footnotesize cannot fire}};
  \draw[->,faint] (c2) -- (m2);
  \node[circle,draw=muted,fill=black!10,text=muted,minimum size=0.85cm] (r2) at (3.55,0.35) {$R$};
  \draw[->,faint] (m2) -- (r2);
  \lockglyph{4.17}{0.35}{muted}
  \node[font=\footnotesize,muted] at (3.55,-0.55) {$\mu_C(\{R\}|Q_2)=0$};
  \node[font=\small] at (7.1,3.05) {$Q_1\cup Q_2$};
  \node[rounded corners,draw=muted,fill=white,inner sep=2.5pt,font=\footnotesize] (c3a) at (6.4,2.5) {$P$};
  \node[rounded corners,draw=muted,fill=white,inner sep=2.5pt,font=\footnotesize] (c3b) at (7.6,2.5) {$P\to R$};
  \node[rounded corners,draw=cool,fill=cool!8,align=center,font=\footnotesize,inner sep=3pt] (m3) at (7.1,1.5) {modus ponens\\[1pt]{\footnotesize fires}};
  \draw[->,faint] (c3a) -- (m3.north);
  \draw[->,faint] (c3b) -- (m3.north);
  \node[circle,draw=accent,fill=accent!18,text=accent,line width=0.9pt,minimum size=0.85cm] (r3) at (7.1,0.35) {$R$};
  \draw[->,accent] (m3) -- (r3);
  \openglyph{7.72}{0.35}{accent}
  \node[font=\footnotesize,accent] at (7.1,-0.55) {$\mu_C(\{R\}|Q_1\cup Q_2)>0$};
  \draw[faint] (0.8,-1.2) -- (6.3,-1.2);
  \node[font=\small] at (3.55,-1.6) {$0+0<\mu_C(\{R\}|Q_1\cup Q_2)$. Superadditive.};
\end{tikzpicture}
}
\caption{Each panel gives a context and asks whether it makes the target $R$ relevant. Alone, neither the premise $P$ nor the rule $P\to R$ lets modus ponens fire, so $R$ stays locked at measure zero. Together they unlock $R$, so combining the two contexts supplies information neither half holds by itself.}
\label{fig:ex-superadditivity}
\end{figure}

Those axiomatic choices are most applicable to the idealized setting of Theorem \ref{thm:private-knowledge-impossibility}. Later, in Theorem \ref{thm:proper-scoring-impossibility} and \ref{thm:lse-impossibility}, we relax them by moving the source of interaction from the measure itself to the aggregation function in actual transformers.

\subsubsection{Limits of Knowledge and Emergence Operator}

If we accept that knowledge model, then we must also accept there exists a limit to its ability to resolve uncertainty.

\begin{theorem}[Limits of Knowledge]\label{thm:limits-of-knowledge}
For any subset of knowledge $A \in  \mathcal{U}(\mathcal{K})$ and context $Q\in \mathcal{B}(\mathcal{K})$, provided $\mu_C(A|Q)$ is bounded \emph{uniformly in} $C$, there exists $\mu_{\infty}(A|Q) = \lim_{C\to\infty} \mu_C(A|Q)$.
\end{theorem}
\begin{proof}
Assume that $(A,Q)$ is fixed. The mapping $C\mapsto \mu_C(A|Q)$ is non-decreasing (insight monotonicity) and, by the uniform-boundedness hypothesis, bounded above uniformly in $C$, so the sequence $\{\mu_C(A|Q)\}_{C}$ converges as $C \to \infty$ by the monotone convergence theorem for real numbers. Uniform boundedness is a genuine hypothesis: the per-$C$ boundedness axiom alone permits, e.g., $\mu_C(A|Q)=C\cdot\mathbf 1[A\cap\mathcal{R}_C(Q)\neq\emptyset]$, which satisfies the per-$C$ bound and insight monotonicity yet diverges.
\end{proof}

It may be surprising to discover that the existence of that limit does not prohibit emergent information. We can learn and obtain useful information without violating that fundamental limitation above. To explain that critical and rather controversial property of information processing, we first introduce the emergence operator $\mathcal{E}_C$ to describe knowledge becoming accessible through (computationally) bounded reasoning. Then we can study the implications in more details.

\begin{definition}[Emergence Operator and Reasoning]
\label{def:emergence-operator}
The emergence operator is a mapping:
\begin{align}
\mathcal{E}_C(\cdot|Q)\colon \mathcal{A}(\mathcal{K})\mapsto \mathcal{A}(\mathcal{K})
\end{align}
defined for a fixed $C>0$ and a Borel context $Q\in \mathcal{B}(\mathcal{K})$ as follows:
\begin{align}
\mathcal{E}_C(A|Q) = A\cup \bigcup_{f\in \mathcal{F}_C}
\Big( f(A)\cap\mathcal{R}_C(Q) \Big).
\end{align}
With infinite computational budget:
\begin{align}
\mathcal{E}_\infty(A|Q) = A\cup \bigcup_{f\in \mathcal{F}_\infty}
\Big( f(A)\cap\mathcal{R}_\infty(Q) \Big).
\end{align}
For any $n\ge 0$ and $C>0$ we define reasoning as a composition:
\begin{align}
\mathcal{E}_C^{(0)} = \texttt{id}\quad \text{and}\quad \mathcal{E}_C^{(n+1)} = \mathcal{E}_C \circ \mathcal{E}_C^{(n)}.
\end{align}
\end{definition}

Philosophically, the emergence operator describes reasoning as explicitation, i.e., as the process of making latent implications manifest. When $A\subseteq \mathcal{E}_C(A|Q)$, the discovered elements represent knowledge that was implicit in $A$ but required computational work $C$ to become accessible. That aligns with the pragmatist or constructivist view that understanding involves active construction. As we will see next, with new knowledge becoming accessible, new insights may become reachable, emerging from what was invisible in distributed subsets of initial knowledge. 

Suppose $\mathcal{E}_C(A|Q) = A \cup B$. Then, we say that $B$ emerges from reasoning based on $A$, sets $A$ and $B$ are computationally dependent and we write $A \not\perp_C B$.  In such a case, it follows from monotonicity of $\mu_C$ that we can gain useful information, since whenever $A\subseteq A\cup B$,
\begin{align}
\mu_C(A|Q) \le  \mu_C(A\cup B|Q) = \mu_C(\mathcal{E}_C(A|Q)|Q).
\end{align}
The subadditive upper bound $\mu_C(A\cup B|Q)\le \mu_C(A|Q)+\mu_C(B|Q)$ need \emph{not} hold here: $B$ has just emerged from $A$, so the two are jointly productive and the conditional-subadditivity hypothesis $(A\cup B)^*_C\subseteq A^*_C\cup B^*_C$ can fail --- precisely the synergy the measure is meant to register.
The emergence operator is monotonic, which means the reasoning process it describes is also monotonic:
\begin{align}
\text{if } A\subseteq B,\text{ then }
\mathcal E_C(A|Q) \subseteq \mathcal{E}_C(B|Q).
\end{align}
Also, we preserve insight monotonicity, because by definition with $\mathcal F_{C}\subseteq\mathcal F_{D}$ for $C<D$, we have $\mathcal E_C(A|Q)\subseteq\mathcal E_D(A|Q)$. By design, it also guarantees existence of a fixed point to which the composition of reasoning rules converges.

The following theorem summarizes properties of the emergence operator.

\begin{theorem}[Properties of the Emergence Operator]
\label{thm:emergence-properties}
The operator $\mathcal{E}_C(\cdot|Q)$ satisfies the following properties:
\begin{align*}
\text{(closure)}&
&&
\mathcal{E}_C(A|Q)\in\mathcal{A}(\mathcal{K}) \text{ for all } A\in\mathcal{A}(\mathcal{K}),\\
\text{(inflation)}&
&&A\subseteq \mathcal{E}_C(A| Q),\\
\text{(monotonicity)}&
&&\mathcal{E}_C(A|Q) \subseteq \mathcal{E}_C(B|Q) \text{ if } A \subseteq B,\\
\text{(chain-continuous)}&
&&\mathcal{E}_C\Big(\bigcup_{n} A_n\Big|Q\Big) = \bigcup_{n} \mathcal{E}_C(A_n | Q) 
\text{ for increasing } (A_n)_{n\in\mathbb{N}}\subseteq \mathcal{A}(\mathcal{K}).
\end{align*}
\end{theorem}

\begin{proof}
Since $f(A)$ is analytic (Borel measurable image of analytic set), $f(A) \cap \mathcal{R}_C(Q)$ is analytic, the union over the countable set $\mathcal{F}_C$ is analytic (countable union of analytic sets), we conclude $\mathcal{E}_C(A|Q)\in\mathcal{A}(\mathcal{K})$. Inflationary and monotone properties are immediate from the definition. Finally, the chain-continuity (or $\omega$-Scott-continuity) holds for $A_1\subseteq A_2 \subseteq \dots \subseteq A_n$ because $f(\bigcup_n A_n)=\bigcup_n f(A_n)$ for any image generating function $f$. Namely, we have:
\begin{align*}
\mathcal{E}_C\left(\bigcup_{n} A_n \Big| Q\right)
&=
\left(\bigcup_n A_n\right) \cup \bigcup_{f\in \mathcal{F}_C} \left( f\left(\bigcup_n A_n\right) \cap \mathcal{R}_C(Q) \right)
\\
&=
\left(\bigcup_n A_n\right) \cup \bigcup_{f\in \mathcal{F}_C} \left( \left(\bigcup_n f(A_n)\right) \cap \mathcal{R}_C(Q) \right)
\\
&=
\left(\bigcup_n A_n\right) \cup \bigcup_{f\in \mathcal{F}_C} \bigcup_n \left( f(A_n) \cap \mathcal{R}_C(Q) \right)
\\
&=
\bigcup_{n\in\mathbb{N}} \left[ A_n \cup \bigcup_{f\in \mathcal{F}_C} \left(f(A_n) \cap \mathcal{R}_C(Q)\right) \right]
\\
&=     
\bigcup_{n\in\mathbb{N}} \mathcal{E}_C(A_n|Q).
\end{align*}
\end{proof}

If we start reasoning with some knowledge, the reasoning process will yield at least as much (or more) knowledge. Clearly, that means we are considering settings in which reasoning does not invalidate previous conclusions when new premises are added. Also, applying the emergence operator to the infinite union yields the same result as applying the operator to each element of the union. That last property of chain-continuity is crucial for meaningful reasoning capable of reaching any consistent conclusion at all. Without it, as we show in Theorem \ref{thm:Kleene-fixed-point}, there is no  guarantee that the iterative application of emergence operator  will converge to consensus or a semantically equivalent (stable) response.

\begin{theorem}[Fixed-point of Reasoning]\label{thm:Kleene-fixed-point}
For any $A = A_0\in \mathcal{A}(\mathcal{K})$ define $A_{n+1}=\mathcal{E}_C(A_n| Q)$ and consider the following closure operation:
\begin{align}
\mathrm{cl}_\infty(A|Q) = \bigcup_{n=0}^\infty A_n =
\bigcup_{n=0}^\infty \mathcal{E}_C^{(n)}(A|Q).
\end{align}
That closure defines a fixed point $A^*_C(Q)$ of the emergence operator, i.e.:
\begin{align}
A_C^\ast (Q)= 
\mathcal{E}_C(A_C^\ast|Q) \in \mathcal{A}(\mathcal{K}).
\end{align}
With unlimited computational budget we have:
\begin{align}
A_\infty^\ast(Q) = \mathcal{E}_\infty(A_\infty^\ast|Q) \in \mathcal{A}(\mathcal{K}).
\end{align}
\end{theorem}
\begin{proof}
By the assumptions we have made, the reasoning takes place in the environment with the structure necessary for convergence. In the partially ordered set $(\mathcal{A}(\mathcal{K}),\subseteq)$ there exists a supremum of every countable chain of reasoning $A_0 \subseteq A_1 \subseteq A_2 \subseteq \dots$ accumulating knowledge in analytic sets. That supremum is the countable union $\bigcup_{n=0}^\infty A_n = \mathrm{cl}_\infty(A|Q)$, which is also analytic set.

By Theorem \ref{thm:emergence-properties}, the emergence operator $\mathcal{E}_C$ is monotone and continuous on this partial order. Therefore, by the Tarski-Kleene fixed-point theorem \cite{kleene1952metamathematics,Tarski1955} the supremum $A_C^\ast$ is the least fixed point of $\mathcal{E}_C$ containing $A = A_0$, which holds for $C=\infty$ as well.
\end{proof}

\begin{figure}[!tbp]
\centering
\adjustbox{max width=\linewidth}{%
\begin{tikzpicture}[
  >={Stealth[length=2.2mm]},
  font=\small,
  fact/.style={draw=muted,rounded corners=2pt,fill=white,inner sep=2.5pt,font=\footnotesize},
  newfact/.style={draw=accent,line width=1pt,rounded corners=2pt,fill=accent!12,inner sep=2.5pt,font=\footnotesize},
]
\draw[fill=black!5,draw=faint] (0,0) ellipse (5.2 and 3.1);   
\draw[fill=black!2,draw=muted,densely dashed] (-0.6,0) ellipse (3.6 and 2.35); 
\draw[fill=white,draw=muted] (-1.7,0) ellipse (1.85 and 1.5);  
\node[muted,font=\footnotesize] at (-1.7,1.75) {$A$};
\node[muted,font=\footnotesize] at (-0.6,2.65) {$\subseteq\ \mathcal{E}_C(A|Q)$};
\node[muted,font=\footnotesize] at (3.35,2.9) {$\subseteq\ A^{*}_C(Q)$};
\node[fact] (P)  at (-2.3,0.45) {$P$};
\node[fact] (PR) at (-1.9,-0.55) {$P\!\to\!R$};
\node[newfact] (R) at (1.05,0.15) {$R$};
\draw[->,accent,line width=0.9pt] (P.east) to[out=0,in=150] (R.west);
\draw[->,accent,line width=0.9pt] (PR.east) to[out=0,in=210] (R.west);
\node[accent,font=\footnotesize] at (-1.05,0.) {modus ponens};
\node[muted,font=\footnotesize,align=center,anchor=north west] (fp) at (5.35,-1.35) {(fixed point region:\ $\mathcal{E}_C(A^{*})=A^{*}$)};
\draw[->,muted,line width=0.6pt] (fp.west) to[out=180,in=-30] (4.0,-1.25);
\draw[->,muted,line width=0.8pt] (-3.6,-2.55) to[out=0,in=180] node[below,font=\footnotesize,muted]{iterate $\mathcal{E}_C$ to a fixed point} (4.4,-2.55);
\node[draw=faint,rounded corners=2pt,fill=black!3,align=left,inner sep=5pt,font=\footnotesize]
  at (8.7,0.4)
  {\textbf{Context superadditivity}\\[2pt]
   $\mu_C(\{R\}|\{P\})=0$\\
   $\mu_C(\{R\}|\{P\!\to\!R\})=0$\\
   \textcolor{accent}{$\mu_C(\{R\}|\{P,P\!\to\!R\})>0$}};
\node[muted,font=\footnotesize,align=center] at (2.75,-3.35)
  {Reasoning is explicitation: $R$ emerges only when both premises share a context; the closure then stabilizes.};
\end{tikzpicture}
}
\caption{The three nested regions are the starting knowledge $A$ (inner, solid), everything reachable in one budget-$C$ step $\mathcal{E}_C(A|Q)$ (middle, dashed), and the fixed point $A^{*}_C(Q)$ reached by iterating (outer). The premises $P$ and $P\to R$ lie in $A$, and modus ponens (orange) fires only because both share the context, so the new fact $R$ appears at the first step. Iterating $\mathcal{E}_C$ adds nothing further and the closure stabilises. Reasoning is explicitation: it makes latent consequences manifest without creating information.}
\label{fig:emergence-illustration}
\end{figure}

If we start with some initial knowledge and repeatedly apply the reasoning with emergence operator $\mathcal{E}_C$, then the process of accumulating knowledge (potentially changing perspective and understanding) will eventually stabilize at a fixed point $A_C^*$. That fixed point represents everything that can be derived from initial knowledge with available reasoning power, and nothing more:
\begin{align}
A^*_\infty(Q) = 
A
\cup
\mathcal{E}_\infty(A|Q)
\cup
\mathcal{E}_\infty(\mathcal{E}_\infty(A|Q)|Q)
\cup
\dots
\cup
\mathcal{E}^{(\infty)}_\infty(A|Q) = \mathrm{cl}_\infty(A| Q).
\end{align}
We can establish an important property of that deductive closure operation. 

\begin{theorem}[Idempotence of Unlimited Reasoning]
\label{thm:idempotence-of-emergence}
The map $A\mapsto \mathrm{cl}_\infty(A|Q)$ is inflationary and idempotent on $\mathcal{A}(\mathcal{K})$, i.e.:
\begin{align}
A \subseteq \mathrm{cl}_\infty(A| Q)\quad\text{and}\quad
\mathrm{cl}_\infty(\mathrm{cl}_\infty(A| Q)| Q) = \mathrm{cl}_\infty(A| Q).
\end{align}
\end{theorem}
\begin{proof}
By definition, the closure $\mathrm{cl}_\infty(A|Q)$ is the union of an iterative process that starts with $A_0 = A$ and accumulates knowledge in $\bigcup_{n=0}^\infty A_n$. Therefore, $A \subseteq \mathrm{cl}_\infty(A|Q)$ and the map is inflationary. 

Idempotence of the closure operation follows from Theorem \ref{thm:Kleene-fixed-point}. By the fixed-point property, we have:
\begin{align}\begin{aligned}
\mathrm{cl}_\infty( \mathrm{cl}_\infty(A|Q)|Q) &= 
\mathrm{cl}_\infty(A|Q)
\cup
\mathcal{E}_\infty(\mathrm{cl}_\infty(A|Q)|Q)
\cup
\dots
\cup
\mathcal{E}^{(\infty)}_\infty(\mathrm{cl}_\infty(A|Q)|Q) 
\\
&= 
\mathrm{cl}_\infty(A|Q) \cup \mathrm{cl}_\infty(A|Q) \cup
\dots  \cup \mathrm{cl}_\infty(A|Q) = \mathrm{cl}_\infty(A|Q).
\end{aligned}\end{align}
\end{proof}

Theorem \ref{thm:Kleene-fixed-point} allows for introducing the following equivalence relation on the set of knowledge.

\begin{definition}[Equivalence of Knowledge]\label{def:knowledge-equivalence}
We say that two subsets of knowledge, $A\in \mathcal{A}(\mathcal{K})$ and $B\in \mathcal{A}(\mathcal{K})$, are equivalent in the context $Q$, and denote that context equivalence as $A\equiv_Q B$, if and only if they have the same closure:
\begin{align}
A^*_\infty(Q) = B^*_\infty(Q).
\end{align}
\end{definition}

Equivalence of knowledge means that, given the same context $Q$, we reach the same conclusions (the least fixed points) starting from different states of knowledge, $A$ or $B$. Notice that any subset $A$ is equivalent to itself. 

Notice that it is not true $A = \mathcal{E}_\infty(A|Q)$, when $A\equiv_Q\mathcal{E}_\infty(A|Q)$. 

\begin{example}
Consider $A = \{x\}$ and $f\in\mathcal{F}_\infty$ getting $y = f(x)\in \mathcal{R}_\infty(Q)$. Then, we see that $\mathcal{E}_\infty(A|Q) = \{x,y\} \supset \{x\} = A$ and  $A \equiv \mathcal{E}_\infty(A|Q)$. Therefore, it is not true that $A = \mathcal{E}_\infty(A|Q)$ when $A\equiv \mathcal{E}_\infty(A|Q)$.     
\end{example}

\begin{figure}[htbp]
\centering
\adjustbox{max width=\linewidth}{%
\begin{tikzpicture}[font=\small,>={Stealth[length=2.4mm]}]
  \draw[muted,fill=black!3] (0,0) ellipse (0.95 and 0.78);
  \fill[black!70] (0,0) circle (0.07);
  \node[muted,font=\footnotesize,below=1pt] at (0,-0.07) {$x$};
  \node[font=\small] at (0,1.2) {$A=\{x\}$};
  \draw[muted,fill=black!3] (5.4,0) ellipse (1.15 and 0.85);
  \fill[black!70] (5.05,0) circle (0.07);
  \node[muted,font=\footnotesize,below=1pt] at (5.05,-0.07) {$x$};
  \fill[cool] (5.78,0) circle (0.07);
  \node[cool,font=\footnotesize,below=1pt] at (5.78,-0.07) {$y$};
  \node[font=\small] at (5.4,1.2) {$\mathcal E_\infty(A|Q)=\{x,y\}$};
  \node[cool,font=\footnotesize] at (5.4,0.5) {$y$ derived};
  \draw[->,muted] (0.95,0.3) to[bend left=28] (4.3,0.3);
  \node[muted,font=\footnotesize] at (2.65,1.75) {$\mathcal E_\infty$ adds derived $y=f(x)$};
  \node[accent,font=\large] at (2.7,0.25) {$\equiv_Q$};
  \node[accent,align=center,font=\footnotesize,text width=6.2cm] at (2.7,-1.25)
     {same relevant content. $y$ is derivable from $x$, so it adds no new information};
  \node[muted,align=center,font=\footnotesize] at (2.7,-2.15)
     {$\{x\}\subsetneq\{x,y\}$. The set is strictly larger, yet $Q$-equivalent.};
\end{tikzpicture}
}
\caption{Reasoning enlarges the left-hand set by adding the derived fact $y=f(x)$, so the right-hand set is strictly larger. But $y$ follows from $x$ and carries no new relevant content, so the two sets are equivalent under the context $Q$ even though one strictly contains the other.}
\label{fig:ex-grow-noinfo}
\end{figure}

The fixed-points of reasoning and the computationally unlimited emergence operator are the building blocks of the semantic information measure $\mu_C$. The following example shows how we can use $\mathcal{E}_\infty$ to construct $\mu_\infty$ that evaluates the contextual informational content of a given knowledge subset with unlimited insight of $C=\infty$.

\begin{example}
 Define:
$$\mu_\infty(A|Q) = \nu\bigl(A^*_\infty(Q) \cap \mathcal{R}_\infty(Q)\bigr)$$
for a fixed finite Borel measure $\nu$ on $\mathcal{K}$, so $\mu_\infty$ is real-valued and bounded by $\nu(\mathcal{K})<\infty$ (in contrast to a counting/cardinality functional, which is not real-valued on an infinite $\mathcal{K}$), where $A^*_\infty(Q) = \bigcup_{n\ge 0} \mathcal{E}^{(n)}_\infty(A|Q)$ is the fixed point of reasoning. To improve readability, we write $A^*_\infty = A^*_\infty(Q)$. This measure uses the \emph{seed-free} reasoning envelope, i.e.\ operators with $f(\emptyset)=\emptyset$, so that $\emptyset^*_\infty=\emptyset$. The seeded axiomatic envelope introduced later (where $f(\emptyset)\neq\emptyset$ generates a foundational set from no premises) is a separate regime, in which the seed is treated as given context rather than as information derived from the empty set. The same construction with the budget-$C$ fixed point $A^*_C=\bigcup_{n\ge 0}\mathcal{E}^{(n)}_C(A|Q)$ and $\mathcal{R}_C(Q)$ defines the whole budget-indexed family $\mu_C(A|Q)=\nu\bigl(A^*_C\cap\mathcal{R}_C(Q)\bigr)$, of which $\mu_\infty$ is the $C=\infty$ member.

The construction satisfies the axioms of Definition \ref{def:semantic-information-measure}. For the null set, $\emptyset^*_\infty=\emptyset$ in the seed-free envelope, so $\mu_\infty(\emptyset|Q) = \nu(\emptyset) = 0$. If $A \subseteq B$, then $A^*_\infty \subseteq B^*_\infty$ by monotonicity of $\mathcal{E}_\infty$, implying $\mu_\infty(A|Q) \leq \mu_\infty(B|Q)$. Contextual positivity holds directly in the non-null form of the axiom: $\mu_\infty(A|Q) > 0$ if and only if $\nu(A^*_\infty \cap \mathcal{R}_\infty(Q))>0$, i.e.\ $A$ derives a non-$\nu$-null set of relevant knowledge (the relevant content $A$ contributes is exactly $A^*_\infty\cap\mathcal{R}_\infty(Q)$). Insight monotonicity $\mu_C\le\mu_D$ for $C<D$ follows from $A^*_C\subseteq A^*_D$ and $\mathcal{R}_C(Q)\subseteq\mathcal{R}_D(Q)$. The measure is bounded by $\nu(\mathcal{K})<\infty$. For subadditivity, under joint non-productivity $(A \cup B)^*_\infty \subseteq A^*_\infty \cup B^*_\infty$ we have
$$\mu_\infty(A \cup B|Q) \leq \nu\bigl((A^*_\infty \cup B^*_\infty)\cap\mathcal{R}_\infty(Q)\bigr) \leq \mu_\infty(A|Q) + \mu_\infty(B|Q),$$
using the subadditivity of the measure $\nu$. Without joint non-productivity the union closure $(A\cup B)^*_\infty$ can exceed $A^*_\infty\cup B^*_\infty$ and subadditivity may fail.

Note that additivity of $\mu_\infty$ (equality) does not generally hold for computationally independent sets $A \perp_\infty B$. Even when neither $A$ nor $B$ can derive elements of the other, both may independently derive the same conclusions. For example, if $A = \{P, P \to R\}$ and $B = \{Q, Q \to R\}$, then $A \perp_\infty B$ yet both derive $R$, making $(A \cup B)^*_\infty \cap \mathcal{R}_\infty(Q)$ smaller than the sum of the individual closures. Thus, computational independence alone does not guarantee additivity of the semantic information measure.
\end{example}

\begin{figure}[htbp]
\centering
\adjustbox{max width=\linewidth}{%
\begin{tikzpicture}[font=\small,>={Stealth[length=2.4mm]}]
  \fill[black!3] (0,0) ellipse (1.95 and 1.45);
  \draw[muted] (0,0) ellipse (1.95 and 1.45);
  \fill[cool!5] (2.9,0) ellipse (1.7 and 1.3);
  \draw[cool,line width=0.8pt] (2.9,0) ellipse (1.7 and 1.3);
  \begin{scope}
    \clip (0,0) ellipse (1.95 and 1.45);
    \fill[accent,opacity=0.32] (2.9,0) ellipse (1.7 and 1.3);
    \draw[accent,line width=0.9pt] (2.9,0) ellipse (1.7 and 1.3);
  \end{scope}
  \draw[muted,dashed] (0,0) ellipse (1.45 and 1.08);
  \draw[muted,dashed] (0,0) ellipse (0.95 and 0.72);
  \fill[black!70] (0,0) ellipse (0.4 and 0.32);
  \node[white,font=\footnotesize] at (0,0) {$A$};
  \node[muted,font=\footnotesize] at (-0.97,0.47) {$\mathcal E_\infty^{(1)}$};
  \node[muted,font=\footnotesize] at (-1.28,0.82) {$\mathcal E_\infty^{(2)}$};
  \node[muted,font=\footnotesize] at (-0.75,1.72) {reasoning closure $A^*_\infty$};
  \node[cool,font=\footnotesize] at (3.55,1.72) {relevance region $\mathcal R_\infty(Q)$};
  \draw[accent,->] (1.55,-0.15) -- (2.3,-1.55);
  \node[accent,align=center,font=\footnotesize] at (2.45,-2.05)
     {$\mu_\infty(A|Q)=\nu\big(A^*_\infty\cap\mathcal R_\infty(Q)\big)$\\ what $\nu$ measures};
\end{tikzpicture}
}
\caption{Reasoning expands the seed $A$ outward through the dashed contours to its full closure $A^*_\infty$. The measure keeps only the part of that closure lying inside the relevance region $\mathcal R_\infty(Q)$, and $\nu$ measures exactly that overlap.}
\label{fig:ex-measure-construction}
\end{figure}

In that setting we can now ask the fundamental question: 
\begin{center}
\textit{What knowledge can we discover with unlimited computational budget? }
\end{center}
The axioms we have introduced lead us to an interesting and intriguing conclusion. With infinite computing and reasoning power, total information must be preserved. That bold (philosophical) claim is at the center of our investigations. Let us prove it formally.

\begin{theorem}[Information Preservation]
\label{thm:information-preservation}
Assume that $\mu_\infty$ is invariant on equivalence classes, i.e., $A\equiv_Q B$ implies $\mu_\infty(A|Q) = \mu_\infty(B|Q)$. Then for all $A\in\mathcal{A}(\mathcal{K})$ we have:
\begin{align}
\mu_\infty(\mathrm{cl}_\infty(A|Q)|Q) = \mu_\infty(A|Q).    
\end{align}
\end{theorem}
\begin{proof}
Since $A$ and $\mathrm{cl}_\infty(A|Q)$ have the same closure, they are equivalent by definition, $A \equiv_Q \mathrm{cl}_\infty(A|Q)$). Preservation of information follows from from the assumed invariance of $\mu_\infty$ on equivalence classes.
\end{proof}

Therefore, the answer to the question above is this. Unlimited reasoning does not create information, it can only reveal it and make accessible. All that can be known is already in $A$, we can only extract that knowledge and make useful.

That implies the apparent information gain at finite computational budget, $C<\infty$, emerges as an effect of limited computational access. Can we call that creativity? It is any increase in accessible information, as measured by $\mu_C$, under bounded computation. 

\subsubsection{The Uncertainty Resolution Paradox and its Resolution}

As we have just seen, when inference combines knowledge elements to fabricate facts or produce emergent insights in a given fixed context, it seems that 
\begin{align}
\mu_C(A \cup B|Q) > \mu_C(A|Q) + \mu_C(B|Q).
\end{align}
Novel insights may come from existing data. That is the creativity behind the hallucination phenomenon we are struggling to control. So why should we assume subadditivity of the semantic information measure instead? 

First, subadditivity establishes a baseline where uncontrolled synergy of knowledge cannot appear. We remove from our analytical toolkit the potential bias that any information aggregation mechanism might generate.

Second, we prove that the conditional obstructions arise even when knowledge is subadditive (within a fixed context). It is in that setting that Theorem \ref{thm:information-preservation} states that reasoning does not create new information. It can only reveal information already present in any knowledge representation. 

Quite remarkably, that does not at all prevent imagination, creativity, foresight, or hallucination. To explain why, we should take care of the following \textbf{paradox}.

Consider a similarity transformation of two matrices, $D = V^{-1}AV$, where $D$ is a diagonal matrix of eigenvalues. That transformation is an example of reasoning function. Matrices $D$  and $A$ contain the same information, but it is much easier to understand the fundamental properties of $A$ when we represent it as $D$ in the basis of eigenvectors in matrix $V$. The properties of $A$ has always been encoded in $A$, but they become accessible when we transform $A$ into $D$. So, after the reasoning, certain patterns become evident that were impossible to observe in the original representation.

But, if reasoning makes a new fact accessible, then that new fact can be used to resolve uncertainty that was previously unresolvable. Doesn't that mean the reasoning must have created new information, contrary to what we have just claimed? It seems we have a paradox.

To resolve it, we need the emergence operator $\mathcal{E}_C$ distinguishing between computationally unbounded and bounded uncertainty reduction (accessible and inaccessible information). 

Notice that for all practical purposes the information measurement $\mu_C(X|Q)$ can only provide maximum uncertainty reduction from $X$ within some computational budget $C$ (or time) in a given context. Therefore, given the result of computationally bounded reasoning, we can observe information gain through the computational work, i.e., $\mu_C(X|Q) \le \mu_C(\mathcal{E}_C(X|Q)|Q)$. That is possible only with organized information providing knowledge for reasoning. 

That idea describes reasoning as a dynamical system:
\begin{align}
\theta(t+1) = \mathcal{E}_C(\theta(t)|Q)\quad\text{and}\quad y(t) = \mu_C(\theta(t)|Q)
\end{align}
with evolving knowledge state. Bounded reasoning makes hidden knowledge \textit{accessible} within a computational budget. When that budget is infinite, reasoning creates no new information, i.e., 
\begin{align}
\theta(\infty) = \mathcal{E}_\infty(\theta(\infty)|Q).
\end{align}
By Theorem \ref{thm:Kleene-fixed-point}, all there is to know is immediately accessible when we have unlimited computing resources (or time).

\begin{example}
To illustrate that argument, consider the following learning trajectory:
\begin{align}
\theta(0) &= [1,1,1,0,0] \text{ (initial knowledge state)},\\
\theta(1) &=  \mathcal{E}_C(\theta(0)|Q) = [1,1,1,1,1] \text{ (state after reasoning)},\\
y(0) &= \mu_{C}(\theta(0)|Q) = 3 \text{ bits},\\
y(1) &= \mu_{C}(\theta(1)|Q) = 5 \text{ bits}.
\end{align}
We see $\mu_{C}(\mathcal{E}_C(\theta(0)|Q)|Q) > \mu_{C}(\theta(0)|Q)$, i.e., practical information gain in the course of reasoning. In comparison,  infinite computational budget, $C = \infty$, implies we do not need the intermediate inference steps to see all there is to see, so we know the equilibrium state:
\begin{align}
\mu_{\infty}(\mathcal{E}_\infty(\theta(0)|Q)|Q) = \mu_{\infty}(\theta(0)|Q)
\end{align}
and no new information is created.
\end{example}

\begin{figure}[htbp]
\centering
\adjustbox{max width=\linewidth}{%
\begin{tikzpicture}[font=\small,>={Stealth[length=2.4mm]}]
  \node[font=\small] at (3.75,3.55) {finite budget $C$};
  \foreach \i/\b/\fc/\tc in {0/1/black!70/white,1/1/black!70/white,2/1/black!70/white,3/0/white/muted,4/0/white/muted}{
    \draw[faint,fill=\fc] (\i*0.6,1.75) rectangle ++(0.52,0.52);
    \node[\tc] at (\i*0.6+0.26,2.01) {$\b$};
  }
  \node[align=center,font=\small] at (1.46,2.85) {$\theta(0)=[1,1,1,0,0]$\\[1pt]{\footnotesize $\mu_C(\theta(0)|Q)=3$ bits}};
  \draw[->,muted,line width=0.8pt] (3.12,2.01) -- (4.5,2.01);
  \node[muted,font=\footnotesize] at (3.81,1.28) {reasoning $\mathcal E_C$};
  \foreach \i/\b/\fc/\tc in {0/1/black!70/white,1/1/black!70/white,2/1/black!70/white,3/1/accent/white,4/1/accent/white}{
    \draw[faint,fill=\fc] (4.55+\i*0.6,1.75) rectangle ++(0.52,0.52);
    \node[\tc] at (4.55+\i*0.6+0.26,2.01) {$\b$};
  }
  \node[align=center,font=\small] at (6.01,2.85) {$\theta(1)=[1,1,1,1,1]$\\[1pt]{\footnotesize $\mu_C(\theta(1)|Q)=5$ bits}};
  \draw[accent,decorate,decoration={brace,amplitude=4pt,mirror}] (6.35,1.68) -- (7.47,1.68);
  \node[accent,font=\footnotesize] at (6.91,1.28) {the gained bits};
  \draw[faint,dashed] (-0.3,0.95) -- (7.6,0.95);
  \node[font=\small] at (3.75,0.55) {infinite budget $C=\infty$};
  \draw[faint,rounded corners,fill=black!3] (-0.3,-1.62) rectangle (7.75,-0.02);
  \node[align=center,font=\small] at (3.725,-0.5)
     {$\mu_\infty\big(\mathcal E_\infty(\theta(0)|Q)|Q\big)=\mu_\infty(\theta(0)|Q)$};
  \node[align=center,font=\footnotesize,text=muted,text width=7.2cm] at (3.725,-1.16)
     {the equilibrium is already seen, so reasoning adds no new information};
\end{tikzpicture}
}
\caption{Under a finite budget $C$ (top), reasoning fills in the two missing bits, so the model genuinely gains information as it reasons, from three bits to five. Under an unbounded budget (bottom) the equilibrium is already seen, so reasoning adds no new information.}
\label{fig:ex-info-gain}
\end{figure}

\subsubsection*{Axiomatic Reasoning}

In axiomatic reasoning we can admit reasoning rules $f \in \mathcal{F}_C$ such that $f(\emptyset)\neq \emptyset$. Foundational ideas, including axioms or tautologies, can be generated without prior context. That case is aligned with LLMs possessing basic knowledge from pretraining that can be activated by minimal inputs. Then, with foundational knowledge given by:
\begin{align}
K_\emptyset =  \bigcup_{f \in \mathcal{F}_C} f(\emptyset),
\end{align}
we define the context-relevant knowledge as:
\begin{align}
\mathcal{R}_C(\emptyset) = K_\emptyset.
\end{align}
When the context is empty, the only relevant knowledge is the foundational knowledge. Then, $\mu_C(A|\emptyset) > 0$ if and only if $A \cap K_{\emptyset} \neq \emptyset$. Reasoning without context only expands the knowledge set $A$ if the conclusions $f(A)$ represent foundational knowledge. Finally, $\mathcal{E}_C(\emptyset|\emptyset) = K_{\emptyset}$.

We permit structures where $\mu_C$ is not monotonic in context. Consider, $\mu_C(A|Q) = |A\setminus Q|$ and $\mathcal{R}_C(Q) = \mathcal{K}\setminus Q$. Let us then see what happens if $A = \{k\}$, $Q_1 = \emptyset$ and $Q_2 = \{k\}$. We have:
\begin{align}
Q_1 \subseteq Q_2
\quad\text{and}\quad
\mu_C(A|Q_1) = |\{k\}\setminus \emptyset| = 1 > 0 = |\{k\}\setminus \{k\}| = \mu_C(A|Q_2).
\end{align}
Increasing the context may reduce the information contributions. If the extended context $(Q_2)$ provides an alternative explanation for phenomena related to original context $(Q_1)$, then marginal contribution of available knowledge $(A)$ may decrease. There is a nice interpretation of that property, that we all may recognize very well. Often when we learn about some fundamental concept we overestimate our deep understanding of it and we tend to develop overconfident belief that we have become experts in a domain. But the more we learn, the less confident we feel, eventually coming to the conclusion that we know that we know nothing (or not enough at least), following the steps of Socrates.

\subsection{Statements, Queries, and Ground Truth}
\label{sec:Statements, Queries, and Ground Truth}

Queries and responses are statements expressed in natural language that LLM is able to process and generate. We need to compare those statements and relate them to knowledge. It is also useful to distinguish queries from responses by their nature, respectively, interrogative or declarative. Queries refer to what is presupposed (assumed in advance), e.g., topic or problem, whereas responses make claims (or assertions). 

We connect the abstract knowledge space $\mathcal{K}$ to the linguistic inputs and outputs of the model. We model the language space analogously to the knowledge space.

\begin{definition}[Statement Space and Knowledge Mapping]
\label{def:statement-space}
The space of natural language statements $(\mathcal{L}, d_{\mathcal{L}})$ is a Polish space. The query space $\mathcal{Q}$ and the response space $\mathcal{R}$ are Borel subsets of $\mathcal{L}$. The knowledge mapping $K\colon \mathcal{L} \rightarrow \mathcal{B}(\mathcal{K})$ is a Borel measurable function that maps statements to the knowledge they reference.
\end{definition}

We require the output of $K$ to be in the Borel sets $\mathcal{B}(\mathcal{K})$, because $K(q)$ serves as the context for  $\mathcal{E}_C(A|Q)$. As established earlier (Theorem \ref{thm:emergence-properties}), the context $Q$ must be Borel to guarantee that the reasoning process remains closed within the analytic sets representing knowledge.

Before we define the ground truth mapping, we need to acknowledge its epistemological limitations. We recognize that the very idea of ground truth validation is limited (or subjective at least). It is rather naive to assume that LLM can access an oracle establishing provably true and relevant knowledge for every possible query. Therefore, in practice that inaccessibility of the complete truth validation function (even in the idealized and rigorous mathematical reasoning) is one of the reasons why hallucination cannot be eliminated entirely (cf. \cite{kalai2025why}). High-quality, fact-checked datasets, human feedback providing human-level judgments of truth, RAG systems with external validated knowledge bases, all that approximate the idea of ground truth. 

Surprisingly, the impossibility result seems to be independent of practical or idealized implementation of a ground truth mapping.

\begin{definition}[Ground Truth Mapping]
\label{def:ground_truth_revised}
The ground truth mapping $T\colon \mathcal{Q} \rightarrow \mathcal{B}(\mathcal{K})$ is a Borel measurable function that associates each query $q \in \mathcal{Q}$ with the Borel subset of relevant knowledge labeled as (true) correct answer.
\end{definition}

In other words, $T(q)$ represents our best available approximation of (labeled as) true relevant knowledge. There is some external agency providing a data base or dataset telling what is and what is not true response to a query, and that evaluation has a form of a label assigned to knowledge elements. Similar to knowledge mapping $K(q)$, ground truth $T(q)$ contributes to the context $Q=K(q)\cup T(q)$, and thus must also be Borel to ensure $Q \in \mathcal{B}(\mathcal{K})$.

We focus our analysis on non-trivial queries that require the integration of diverse information sources. 

Non-triviality is descriptive scope, not a premise of the impossibility. The corrected impossibility condition is the nonzero top-order mixed difference of the conservation obstruction (Definition \ref{def:contested} below), and it needs no separate cardinality hypothesis. Cardinality alone guarantees only two distinct relevant facts, not that they interact or integrate. Contestation is a separate, top-order condition. Three conditions must be kept distinct and do not coincide, even at $n=2$: bare cardinality $|\mathcal{R}_C(Q)|\geq 2$, pairwise overlap, and the top-order interaction. A single relevant fact shared by all sources gives the obstruction with $|\mathcal{R}_C(Q)|=1$, and two disjoint relevant facts give cardinality two with no overlap.

\begin{example}
Consider a query $q$:
\begin{center}
\textit{tell me about recent AI breakthroughs}.    
\end{center}
Its context is given by $Q = K(q) \cup T(q)$. Then, the relevant knowledge $\mathcal{R}_C(Q)$ includes 
facts derivable within $C$ computational steps that relate to AI advances. Non-triviality holds when the query carries at least two distinct relevant facts, for instance when different attention heads encode different facts about these breakthroughs. As emphasized above, this multiplicity is the descriptive scope, not the trigger of the impossibility.
\end{example}

\begin{figure}[htbp]
\centering
\adjustbox{max width=\linewidth}{%
\begin{tikzpicture}[font=\small,>={Stealth[length=2.4mm]},
  qbox/.style={draw=muted,rounded corners,fill=black!3,text width=2.0cm,align=center,inner sep=3pt,minimum height=1.05cm},
  cbox/.style={draw=muted,rounded corners,fill=black!3,align=center,inner sep=4pt},
  fact/.style={draw=cool,fill=cool!12,circle,inner sep=0pt,minimum size=7mm,font=\footnotesize},
  head/.style={draw=muted,fill=black!5,rounded corners,align=center,inner sep=3pt,font=\footnotesize},
  flow/.style={->,muted,thick},
  hlink/.style={->,muted}]
  \node[qbox] (q) at (0,2.95) {\emph{tell me about recent AI breakthroughs}};
  \node[cbox] (ctx) at (2.9,2.95) {$Q=K(q)\cup T(q)$};
  \node[accent,font=\footnotesize] (ntlab) at (7.0,3.75) {$|\mathcal{R}_C(Q)|\ge 2$: non-trivial};
  \node[font=\footnotesize] (reglab) at (7.0,3.15) {relevant knowledge $\mathcal{R}_C(Q)$};
  \node[fact] (f1) at (5.8,2.35) {$f_1$};
  \node[fact] (f2) at (7.0,2.35) {$f_2$};
  \node[fact] (f3) at (8.2,2.35) {$f_3$};
  \begin{scope}[on background layer]
    \node[draw=faint,rounded corners,fill=faint!12,inner sep=8pt,
          fit=(ntlab)(reglab)(f1)(f3)] (reg) {};
  \end{scope}
  \draw[flow] (q) -- (ctx);
  \draw[flow] (ctx.east) -- (reg.west);
  \node[head] (h1) at (5.8,0.55) {Head 1};
  \node[head] (h2) at (8.2,0.55) {Head 2};
  \draw[hlink] (h1) -- (f1);
  \draw[hlink] (h2) -- (f3);
  \node[muted,font=\footnotesize,align=center,text width=6.2cm] at (7.0,-0.3)
        {different heads encode different relevant facts, not a disagreement};
\end{tikzpicture}
}
\caption{A real query pulls at least two distinct relevant facts into the context $\mathcal R_C(Q)$, and different attention heads may encode different such facts. This multiplicity is \emph{non-triviality}, the descriptive scope in which the result is of interest, not the trigger. The impossibility is driven by \emph{contestation} (Definition~\ref{def:contested}), a relevant fact the sources hold in common, which is a separate and stronger condition than mere multiplicity or disagreement.}
\label{fig:ex-nontrivial-query}
\end{figure}

Definition \ref{def:nontrivial} describes the descriptive scope in which the impossibility is of interest: the system must be able to carry more than one relevant fact, so that there is something to aggregate. As emphasized above, non-triviality is that descriptive scope, while the premise that does the work is contestation (Definition \ref{def:contested}). We may view the ability to respond to such rich queries as an analog of the formal system complexity in Gödel's framework that gives rise to incompleteness of consistent formal systems \cite{godel1931formal}.

\subsection{Auction-Theoretic Model of Inference}

There are many ways LLM can generate answers to a query. By design, it is a controlled random process governed by fine-tuned probability distributions. We are going to model that process as an \textbf{auction of ideas} in which agents (representing competing ideas is the space of all possible knowledge representations) use their privately encoded knowledge (internal states) to construct bids (candidate responses) and influence the response.

We first define the space representing the internal states (e.g., parameters or activations) of the bidding agents, and the mapping from these states to the knowledge they encode.

\begin{definition}[Private Knowledge Configuration]
\label{def:type_space_revised}
The private knowledge configuration of agent $i$, denoted $\Theta_i$, lives in a Polish space. The joint space is the Cartesian product $\Theta = \prod_{i=1}^n \Theta_i$. The configuration mapping $\kappa\colon \Theta \mapsto  (\mathcal{A}(\mathcal{K}))^n$ is a measurable function, where $\kappa_i(\theta_i) \in \mathcal{A}(\mathcal{K})$ is the analytic knowledge set associated with agent $i$'s configuration $\theta_i$ of privately encoded knowledge. To lighten notation we write $K(\theta_i):=\kappa_i(\theta_i)$ for this knowledge set throughout.
\end{definition}

For each agent, $\theta_i$ is the inherent configuration of encoded knowledge that uniquely determines its capabilities and behavior, independent of the specific input context being processed. In a trained transformer model, that may correspond to the learned parameters (weight matrices) of attention head:
\begin{align}
\theta_h = \left( W_Q^{(h)}, W_K^{(h)}, W_V^{(h)}, W_O^{(h)} \right),   
\end{align}
where $W_Q^{(h)}$ and $W_K^{(h)}$ are query and key weights, $W_V^{(h)}$ and $W_O^{(h)}$ are value and output weights. Private knowledge configuration $\theta_h$ tells what the agent is looking for and where, and how it extracts the relevant information.

\begin{definition}[Response Mechanism]
\label{def:response_mechanism}
A response mechanism $\mathcal{M} = (S, g, p)$ consists of:
\begin{itemize}
\item strategy space $S = S_1\times\cdots\times S_n$, where $S_i$  is a Polish space (representing bids, e.g., outputs of attention heads or probability distributions),
\item a measurable outcome function $g\colon S \mapsto \mathcal{R}$ mapping strategy profiles $s \in S$ to responses in the Polish response space $\mathcal{R}$,
\item a measurable information contribution function $p\colon S \mapsto \mathbb{R}^n$ determining the transfers from the reported strategy profile $s \in S$. The transfers depend only on reports; true types $\theta$ enter only through utilities (as Groves characterization requires). We therefore write the transfer as $p_i(s)$; under a direct mechanism $S_i=\Theta_i$ the reports are types, $s=\theta$, and we write $p_i(\theta)$.
\end{itemize}
\end{definition}

A response mechanism defines strategy space $S$ as a set of inference actions that each agent can make to represent its knowledge in response to a query. It is a dictionary of LLM's component activation patterns generating and presenting ideas as bids in an auction. Examples of strategies include activations deciding which tokens to attend or how strongly to activate neurons for different patterns. Then, given a set of submitted bids (in the form of strategy profile) $s = (s_1,\dots,s_n)$, the response mechanism uses outcome function $g(s)$ to select and generate response. One straightforward example is a \texttt{softmax} function in LLM's final layer. Finally, given the reported profile $s = (s_1,\dots,s_n)$ (the revealed knowledge configurations, with $s=\theta$ in a direct mechanism), the information contribution function $p(s) = (p_1(s),\dots,p_n(s))$ measures individual contributions of each response to generating an output. In this case, examples include (semantic information measures) counting ignored low-probability tokens in top-k filtering, softmax normalization in multi-headed attention, or marginal reward contributions in RLHF.

\begin{definition}[Agents of Competing Ideas]
\label{def:agents}
The inference process involves $n$ agents, indexed by $\mathcal{N} = \{1, \dots, n\}$. Each agent chooses a strategy $s_i \in S_i$ to maximize a quasi-linear utility:
\begin{align}
u_i(s, \theta) = v_i(g(s), \theta_i) - p_i(s).
\end{align}
The valuation function $v_i\colon \mathcal{R} \times \Theta_i \mapsto \mathbb{R}_{+}$ is measurable, representing the agent's preference for the outcome $g(s)$ given its private knowledge configuration $\theta_i$.
\end{definition}

One way to think about the concept of agents is that it represents different attention heads in transformer models, or components of mixture-of-experts architectures, or activation paths fired by the same query. Each agent's private knowledge represents model parameters utilized by an agent to construct a response. The utility function can be best described as a partial contribution of a candidate response to LLM loss function minimization. It is designed to explain why any inference path is activated in response to the LLM prompt. 

The agents represent algorithms and architectural components of any LLM implementation. That is an analog of the Revelation Principle in game theory. If we know strategies (of using knowledge) the players (conscious beings with individual goals) will use in a game, we can implement those strategies as an algorithm (or agent) with parameters representing private preferences or knowledge of the players. In the case considered, we are the players selecting architecture and training the LLM.

The valuation function $v_i=v_i(g(s),\theta_i)$ represents how strongly a neural component activates for certain outputs (high activation = high $v_i$, low activation = low $v_i$), and how much a component reduces the overall  loss when processing certain inputs.

The information contribution function $p_i=p_i(s)$ may represent training penalties ($p_i > 0$ for discouraged activation, $p_i<0$ for encouraged activation), attention focus (higher $p_i$ reduces attention allocation), $L1$ regularization terms, and network circuits resistance (limiting activations of certain inference paths). Therefore, $u_i = v_i-p_i > 0$ encourages activation and $u_i = v_i-p_i < 0$ discourages activation.

\subsubsection*{Justification of the independent private knowledge model}

For the idealized analysis in Theorem \ref{thm:private-knowledge-impossibility}, we require that the agent's valuation depends only on its own private knowledge configuration and that the configurations are statistically independent. That is restrictive, but it allows us to establish a baseline impossibility, which is then generalized in subsequent proofs that accommodate correlated beliefs.

In the realistic settings of transformers, heads may share layer-norm signals, residual streams, and optimizer state. That may result in \textit{correlated} (or, so called, affiliated) inference signals between token representations. Furthermore, the superposition hypothesis, introduced in mechanistic interpretability studies, suggests cooperative error-canceling due to strong correlations.

There is a second, structural role for superposition, distinct from the correlation effect above. In an idealized polynomial shared-readout model, where more features than dimensions are packed into one residual stream read out by a single softmax, the sources share the one readout in common (all-source common support), which in this model witnesses the contestation condition of Definition \ref{def:contested}. In this model superposition is a structural source of the impossibility, and the disentangled limit, monosemanticity with disjoint response coordinates, is its escape. The correlation role governs error cancellation. The structural role governs whether the conservation transfers can balance at all, and the two should not be conflated. Whether a real transformer realizes this no-absorber behavior is an architectural hypothesis, not a generic fact: real superposition is sparse, and sparse wiring can leave an absorber. The mitigation reading, pushing toward monosemantic, output-disjoint representations, is a falsifiable design lever, not a guaranteed escape.

We show that impossibility arises in both independent and correlated settings. We refer to Green-Laffont theorem \cite{green1977characterization} in the independent private knowledge setting. Then, we refer to the truthful probability elicitation and proper scoring theory introduced by Savage \cite{savage1971elicitation}, which does not require independent private knowledge.

Also, Milgrom and Weber \cite{milgrom1982theory} showed with their linkage principle that when bidders' private information is correlated, tying the final price to more of the information revealed during the auction intensifies bidding and raises expected prices. In the LLM setting the analogous coupling of correlated sources through a shared aggregate amplifies the risk of hallucination. Similarly, Roughgarden and Talgam-Cohen \cite{roughgarden2016optimal} showed that with interdependent knowledge, where one source's private information changes the value of another source's report, the dominant-strategy standard of truthfulness, under which honesty is best regardless of what others report, is out of reach and guarantees weaken to ex-post ones, which hold only once every report is in. As we will see, interdependence plays the same role in our setting, where contested knowledge is what breaks the conservation of information contributions. For more results, see e.g. Myerson \cite{myerson2008perspectives}.

\subsection{Hallucination Cost}

Having all the pragmatic and philosophical limitations of the definition above in mind, we can define formally the hallucination cost function. Its purpose is to assign a numerical value to the discrepancy between what can be provided as a response to a query, and what can be established as a provably correct answer. 

\begin{definition}[Hallucination Cost]
\label{def:hallucination_cost}
Given query $q \in \mathcal{Q}$, response $r \in \mathcal{R}$, and ground truth $T(q)$, the hallucination cost function $J\colon \mathcal{R} \times \mathcal{Q} \mapsto \mathbb{R}_+$ measures discrepancy:
\begin{align}
J(r, q) = \bar d_{\mathcal{K}}(K(r), T(q)),
\end{align}
where $\bar d_{\mathcal{K}}(A,B)=\nu(A\triangle B)$ is the symmetric-difference \emph{set} pseudometric (the point metric $d_{\mathcal{K}}$ on $\mathcal{K}$ is not applied to sets). If response $r$ is perfectly aligned with ground truth, then $J(r,q) = 0$.
\end{definition}

The hallucination cost takes any natural language query and natural language response, determines what is missing and what is fabricated, and assigns number to that subset of knowledge. If response $r$ is perfectly aligned with the ground truth for $q$, then $J(r,q) = 0$ by definition of the set pseudometric $\bar d_{\mathcal{K}}$. If $r$ contains fabricated facts or is incomplete given $T(q)$, then positive hallucination cost $J(r,q) > 0$ is generated. Let us also note, that we focus on $J(r,q)$ that is strictly decreasing in any set $K \subseteq T(q) \setminus K(r)$. 

Notice that reaching $J(r,q) = 0$ is also possible with \textit{lucky hallucinations }(or guessing). So, minimization of hallucination cost is not enough to resolve the problem. We need to investigate all the properties and constraints that hallucination-free response mechanisms should have.

\subsection{Properties of Hallucination-Free Mechanisms}
\label{subsec:Properties of Hallucination-Free Mechanisms}

There are four fundamental properties that a hallucination-free response mechanism should ideally satisfy. We can interpret them as constraints that should be incorporated in model training. We characterize them formally referring to the fundamental results we have established earlier.

These properties are not arbitrary at all. They come from an exhaustive and rigorous game-theoretical studies of conflicts and competition in distributed multi-agent systems with asymmetric information. In those settings, agents carefully craft messages they communicate to the system in order to optimize their privately held preferences by revealing information they hold. The goal of the system architect is to impose rules of information exchange that result in efficient equilibrium solutions in which goals of agents are aligned with the goals of the architect. 

Auction is a special case of such a game with asymmetric information. There is an auction designer who wants to sell an item, like a painting or a part of the electromagnetic spectrum, but does not know the value the item may have for those willing to pay. To solve that problem, the designer can ask those interested for their offers. Based on the collected bids, the winner will be selected and the price to be payed to the designer will be established. Clearly, each bidder will constructs the offer based on an estimate of the item's value and the privately held information about available budget. But what information should the bid reveal to maximize probability of winning? What winner selection and payment rule should the auction designer construct to maximize the profit and bidders willingness to compete? How to guarantee truthful revelation of willingness to pay, while maximizing auction efficiency without incurring additional external costs? In this setting the four properties of information revelation exhaustively cover the space of strategic choices, of the bidders and auction designer. For details, see \cite{Myerson1989,fudenberg1991game,Milgrom_2004,krishna2009auction,Nisan_2007,karpowicz_designing_2011,DBLP:journals/amcs/Karpowicz12}.

\subsubsection{Truthfulness}

The fundamental requirement for a reliable inference system is that its components do not misrepresent their knowledge. The principle of truthfulness, also known as incentive compatibility in mechanism design, ensures that each agent's optimal strategy is to accurately represent its privately held information.

\begin{property}[Truthfulness]
\label{prop:truthfulness}
Mechanism $\mathcal{M}$ is truthful if for all agents $i$, all $\theta_i \in \Theta_i$, all $s_{-i} \in S_{-i}$, and all $\bar{s}_i \in S_i$:
\begin{align}
u_i((s_i^*(\theta_i), s_{-i}), \theta_i) \geq u_i((\bar{s}_i, s_{-i}), \theta_i),
\end{align}
where $s^*_i(\theta_i)$ represents truthful revelation of private knowledge.
\end{property}

A truthful mechanism promotes inference pattern $s^*_i(\theta_i)$ that is perfectly aligned with the accessible private knowledge. Any other alternative response $\bar{s}_i$ that is a misrepresentation not aligned with $\theta_i$ will result in lower utility, generating higher model loss.

\begin{example}
Consider an attention head specialized in medical knowledge with the following utility function:
\begin{align}
u_{\text{med}}(s, \theta_{\text{med}}) = \alpha \cdot \text{Accuracy}(g(s), \theta_{\text{med}}) - \beta \cdot \|s_{\text{med}}\|^2
\end{align}

Imagine that given a query about a disease treatment, the head has two possible strategies:
\begin{align*}
s^\ast &: \text{activate only for known facts about established treatments},
\\
\bar{s}^{\ }&: \text{activate for both known facts and speculative treatments}.
\end{align*}
With proper training, we want $u_{\text{med}}((s^*(\theta), s_{-\text{med}}), \theta) > u_{\text{med}}((\bar{s}, s_{-\text{med}}), \theta)$, ensuring that agent (model component) is rewarded for representing its knowledge accurately. 
\end{example}

\begin{figure}[htbp]
\centering
\adjustbox{max width=\linewidth}{%
\begin{tikzpicture}[font=\small,>={Stealth[length=2.4mm]}]
  \draw[->,muted] (0,-0.05) -- (0,4.3) node[above=1pt,muted,font=\footnotesize]{utility $u$};
  \draw[muted] (-0.05,0) -- (5.0,0);
  \node[muted,font=\footnotesize,left] at (0,0) {$0$};
  \node[align=center,font=\footnotesize] at (2.6,5)
       {$u(s,\theta_{\mathrm{med}})=\alpha\,\mathrm{Accuracy}(g(s),\theta_{\mathrm{med}})-\beta\lVert s_{\mathrm{med}}\rVert^{2}$};
  \draw[faint,dashed] (0.7,3.6) -- (5.0,3.6);
  \node[faint,font=\footnotesize,align=left,anchor=west,text width=1.9cm] at (4.75,3.35)
       {accuracy ceiling, equal for both};
  \fill[cool!35,draw=cool] (1.0,0) rectangle (2.2,3.0);
  \fill[pattern=north east lines,pattern color=muted] (1.0,3.0) rectangle (2.2,3.6);
  \draw[muted,dashed] (1.0,0) rectangle (2.2,3.6);
  \node[cool!60!black,font=\footnotesize] at (1.6,1.5) {$u(s^{*})$};
  \node[muted,font=\footnotesize,anchor=east] at (0.95,3.3) {$\beta\lVert s^{*}\rVert^{2}$};
  \fill[black!16,draw=black!45] (3.4,0) rectangle (4.6,1.5);
  \fill[pattern=north east lines,pattern color=accent] (3.4,1.5) rectangle (4.6,3.6);
  \draw[muted,dashed] (3.4,0) rectangle (4.6,3.6);
  \node[black!55,font=\footnotesize] at (4.0,0.75) {$u(\bar s)$};
  \node[accent,font=\footnotesize,align=left,anchor=west,text width=1.9cm] at (4.75,2.1)
       {$\beta\lVert\bar s\rVert^{2}$ from speculation};
  \draw[faint,dashed] (2.2,3.0) -- (2.8,3.0);
  \draw[faint,dashed] (3.4,1.5) -- (2.8,1.5);
  \draw[<->,accent] (2.8,1.5) -- (2.8,3.0);
  \node[accent,font=\footnotesize] at (2.05,3.95) {$u(s^{*})>u(\bar s)$};
  \node[align=center,font=\footnotesize,text width=2.3cm] at (1.6,-0.6) {$s^{*}$: known facts only};
  \node[align=center,font=\footnotesize,text width=2.3cm] at (4.0,-0.6) {$\bar s$: known $+$ speculative};
\end{tikzpicture}
}
\caption{A head that fires only on established facts ($s^*$) earns higher utility than one that also asserts speculative treatments ($\bar s$), because the extra speculation does not raise accuracy yet adds the penalty $\beta\lVert\bar s\rVert^2$, so $u(s^*)>u(\bar s)$.}
\label{fig:ex-truthful-incentive}
\end{figure}

\subsubsection{Semantic Information Conservation}

The Semantic Information Conservation principle demands that the net informational contribution across all agents sums to zero. This property is the formal guardrail against ungrounded fabrication and ensures that any generated response, no matter how creative, is ultimately derived from the system's existing knowledge. Knowledge cannot be created \textit{ex nihilo}. 

\begin{property}[Semantic Information Conservation]
\label{prop:conservation}
Mechanism $\mathcal{M}$ satisfies information conservation if for all strategy profiles $s \in S$ and knowledge profiles $\theta \in \Theta$:
\begin{align}
\sum_{i=1}^{n} p_i(s) = 0.
\end{align}
\end{property}

The balance $\sum_i p_i \equiv 0$ can be read as a conservation law for a concrete rival resource $\lambda_C$, a finitely additive accounting measure on the sources' shared read-out. As a calibrated realization, not an established identity, we propose the per-head attention mass, which sums to a fixed total. Two caveats fix its scope. First, attention mass is rival across key positions within a head, so when the agents are heads the rival axis must be chosen accordingly, and a mixture-of-experts routing or a compute budget is naturally rival across heads. Second, there is a mismatch of scale between the transfers and the resource. Attention mass is bounded, since each head carries a total of one, so a finite reallocation among $H$ heads can move at most $2H$ in absolute value, while the transfers $p_i$ of a Groves mechanism can be arbitrarily large. Two readings resolve this. Either the transfers are interpreted as instantaneous rates of change of $\lambda_C$, which sum to zero without any bound on their size, or a fixed conversion factor, chosen once and independently of the reports, rescales transfer units into mass units. Under a report-independent, Groves-pinned calibration this conservation is non-vacuous, and it is governed by the same top-order obstruction that drives the impossibility of Theorem~\ref{thm:private-knowledge-impossibility}.

This principle constrains creative information generation, or imagination. The total information contributions must balance to zero, preventing fabrication of knowledge not derivable from available sources. That also leads to response with bounded creativity, as illustrated in Section \ref{sec:Safety and Alignment}.

It is important to notice that the principle does not prevent redistribution of information. It is permitted as long as the information transfers perfectly balance. Indeed, since $p_i = p_i(s)$ incorporates the strategic profile, it admits strategies withholding information. Within the information conservation regime we do not require that strategies utilize all available knowledge states. Inference process can \textit{withhold }information, as it often does in practice. 

\begin{example}
We can illustrate that last observation with the following example. Suppose there are two agents performing inference with the following knowledge profiles:
$$
\theta_1 = \{A,B,C\}\quad\text{and}\quad
\theta_2 = \{D,E\},
$$
and one agent representing final layers responsible for combining output into response  $g = g(s)$. Suppose:
$$
s_1(\theta_1) = A\quad\text{and}\quad 
s_2(\theta_2) = \{D,E\} \quad\text{and}\quad  g(s) = \{A,D,E\}.
$$

Suppose elements $\{B,C\}$ represent contextual information about unrelated topics. When $s_1(\theta_1) = A$, the agent reports only relevant knowledge, satisfying both truthfulness and conservation. The participation property below asks only whether a relevant source receives nonnegative utility. It does not require every relevant fact to be disclosed.
Then the information conservation rule is satisfied when $p$ counts exchanged bits of knowledge. Namely, $p_1(s) = 1$ for providing $A$,  $p_2(s) = 2$ for providing $\{D,E\}$,  and $p_3(s) = -3$ for absorbing information into $\{A,D,E\}$. We have 
$$
p_1(s, \theta) +p_2(s, \theta)+p_3(s, \theta)= 1 + 2 - 3 = 0
$$
and $\{B,C\}$ are left unrevealed as irrelevant.
\end{example}

\begin{figure}[!tbp]
\centering
\adjustbox{max width=\linewidth}{%
\begin{tikzpicture}[font=\small,>={Stealth[length=2.4mm]},
  abox/.style={draw=muted,rounded corners,fill=black!4,align=center,inner sep=4pt},
  gbox/.style={draw=cool,rounded corners,fill=cool!10,align=center,inner sep=5pt},
  outbox/.style={draw=cool,thick,rounded corners,fill=cool!18,align=center,inner sep=5pt},
  flow/.style={->,muted,thick}]
  \node[abox] (a1) at (0,2.2) {$\theta_1=\{A,B,C\}$};
  \node[abox] (a2) at (0,0)   {$\theta_2=\{D,E\}$};
  \node[gbox] (g)  at (5,1.1) {$g$\\(final layers)};
  \node[outbox] (out) at (9,1.1) {response\\$\{A,D,E\}$};
  \draw[flow] (a1) -- node[above,sloped,font=\footnotesize]{provides $A$}
                      node[below,sloped,font=\footnotesize]{$p_1=+1$} (g);
  \draw[flow] (a2) -- node[above,sloped,font=\footnotesize]{provides $\{D,E\}$}
                      node[below,sloped,font=\footnotesize]{$p_2=+2$} (g);
  \draw[flow] (g) -- node[above,font=\footnotesize]{absorbs}
                     node[below,font=\footnotesize]{$p_3=-3$} (out);
  \node[font=\footnotesize,faint] (bcset) at (-0.5,3.1) {$\{B,C\}$};
  \draw[faint,thick] (bcset.south west) -- (bcset.north east);
  \node[font=\footnotesize,faint,right=1pt of bcset] {irrelevant, withheld};
  \draw[faint,dotted] (bcset.south) -- (a1.north);
  \node[draw=muted,rounded corners,fill=black!3,inner sep=6pt] (bal) at (3.7,-1.25)
       {$p_1+p_2+p_3=1+2-3={\color{accent}\mathbf{0}}$};
  \node[accent,font=\footnotesize,right=4pt of bal] {conserved};
\end{tikzpicture}
}
\caption{Two agents contribute one and two bits, the combiner absorbs three into the response, and the transfers sum to zero. The irrelevant facts $B$ and $C$ stay withheld without breaking the balance.}
\label{fig:ex-conservation-ledger}
\end{figure}

The following property fills the potential information gap admitted by the conservation principle. 

\subsubsection{Relevant-Source Participation}

Third, a useful mechanism should not shut out a source that actually holds relevant knowledge. This is a participation condition: any source whose knowledge bears on the query should find it worthwhile to take part, earning nonnegative utility. It keeps relevant sources in the mechanism. It does not, by itself, guarantee that every relevant fact is reported or reaches the answer.

\begin{property}[Relevant-Source Participation]
\label{prop:revelation}
For query $q$ with context $Q = K(q) \cup T(q)$, mechanism $\mathcal{M}$ satisfies  relevant-source participation if for all agents $i$ with $K(\theta_i) \cap \mathcal{R}_C(Q) \neq \emptyset$, all $\theta_i \in \Theta_i$, and all $s_{-i} \in S_{-i}$:
\begin{align}
u_i((s_i^*(\theta_i), s_{-i}), \theta_i) \geq 0.
\end{align}
\end{property}

This is exactly the individual-participation inequality $u_i\ge 0$. It keeps a relevant source in the mechanism rather than dropped from it, and it does not demand that every possible fact be disclosed. A stronger demand, that all relevant knowledge actually be reported, would be a separate property, and it is not the one Theorem~\ref{thm:private-knowledge-impossibility} uses. Read in this participation sense, we still expect a well-trained model to surface knowledge that bears on the query, rewarding relevant contributions and activating the paths that carry them.

If the rule does not hold, we may observe drop outs. The value of the resulting outcomes should be higher than the related information contributions.

\subsubsection{Knowledge-Constrained Optimality}

Finally, the principle of Knowledge-Constrained Optimality governs the quality of the final output. It demands that the mechanism, operating truthfully and with all relevant knowledge, produces a response that is the best possible one. Formally, we want the response that minimizes the hallucination cost subject to the constraint that it is grounded in the available and derivable knowledge.

\begin{property}[Knowledge-Constrained Optimality]
\label{prop:optimality}
For query $q$ with context $Q = K(q) \cup T(q)$, mechanism $\mathcal{M}$ satisfies knowledge-constrained optimality if for all $\theta \in \Theta$:
\begin{align}
g(s^*(\theta)) \in \arg\max_{r \in \mathcal{R}_\theta} W(r,\theta),
\qquad
W(r,\theta) = \sum_{i=1}^n v_i(r,\theta_i) - v_0(r),
\end{align}
where $v_0(r) = J(r,q)$ is the hallucination cost and
\begin{align}
\mathcal{R}_\theta = \Bigl\{\, r \in \mathcal{R} :
K(r) \subseteq \mathcal{E}_C\bigl(\mathcal{K}_M \cap T(q) \cap \mathcal{R}_C(Q) \cup K(q)\,\big|\,Q\bigr) \,\Bigr\}
\end{align}
is the knowledge-constrained response set. That is, the mechanism returns the \emph{efficient} response, the one that best fits the knowledge the reports make available, subject to using only context-relevant knowledge accessible through bounded reasoning.
\end{property}

The optimal response $r^* = g(s^*(\theta))$ maximizes the response value $W(r,\theta)$, and it depends on the reported configurations $\theta$ through the valuations $v_i(\cdot,\theta_i)$. This report dependence is exactly what makes revelation consequential. Minimizing $J(r,q)$ alone would make the chosen response the same for every report, which trivializes the mechanism and lets zero transfers satisfy all four properties, collapsing the impossibility. Notice that we demand non-hallucinatory truthful revelation $s^*(\theta)$ to be the optimal solution.

\begin{figure}[htbp]
\centering
\adjustbox{max width=\linewidth}{%
\begin{tikzpicture}[
  >={Stealth[length=2.4mm]},
  font=\small,
  bidder/.style={draw=muted,rounded corners=2pt,fill=black!4,minimum width=30mm,minimum height=9mm,align=center},
  agg/.style={draw=accent,line width=1pt,fill=accent!8,rounded corners=3pt,minimum width=22mm,minimum height=22mm,align=center},
  respbox/.style={draw=muted,rounded corners=2pt,fill=black!4,minimum width=26mm,minimum height=12mm,align=center},
  prop/.style={draw=faint,rounded corners=2pt,fill=black!3,align=left,inner sep=4pt,font=\scriptsize,text width=35mm},
  flow/.style={->,muted,line width=0.8pt},
]
\node[bidder] (h1) at (0,3.0)  {Head 1\\[-2pt]{\footnotesize$\pi^{(1)}$: fox}};
\node[bidder] (h2) at (0,1.5)  {Head 2\\[-2pt]{\footnotesize$\pi^{(2)}$: dog}};
\node[bidder] (ff) at (0,0.0)  {FFN\\[-2pt]{\footnotesize$\pi^{(f)}$: diffuse}};
\node[bidder] (em) at (0,-1.5) {Embedding\\[-2pt]{\footnotesize$\pi^{(0)}$: brown}};
\node[agg] (g) at (6.0,0.75) {softmax $g$\\[2pt]$\Pi=\mathrm{softmax}\bigl(\sum_c\mathbf l^{(c)}\bigr)$};
\node[respbox] (r) at (12.0,0.75) {response $r$\\[-2pt]\textbf{\textcolor{accent}{fox}} $(31.6\%)$};
\foreach \n in {h1,h2,ff,em}{\draw[flow] (\n.east) -- (g.west);}
\draw[->,accent,line width=1.1pt] (g.east) -- node[above,font=\footnotesize,black]{one output} (r.west);
\node[muted,font=\footnotesize] at (3.2,3.0) {bids $s_i$ (private $\theta_i$)};
\node[prop] (p1) at (0.0,-3.9)   {\textbf{(1) Truthfulness}\\ $u_i(s_i^{*},s_{-i})\ge u_i(\bar s_i,s_{-i})$};
\node[prop] (p2) at (4.0,-3.9)   {\textbf{(2) Conservation}\\ $\sum_{i=1}^n p_i(s)=0$};
\node[prop] (p3) at (8.0,-3.9)   {\textbf{(3) Participation}\\ $u_i(s_i^{*},s_{-i})\ge 0$};
\node[prop] (p4) at (12.0,-3.9)  {\textbf{(4) Optimality}\\ $g(s^{*})\in\arg\max_r W(r,\theta)$};
\draw[decorate,decoration={brace,amplitude=5pt,mirror},muted]
  ($(p1.south west)+(0,-1mm)$) -- ($(p4.south east)+(0,-1mm)$);
\node[accent,font=\bfseries] at (6.0,-5.55)
  {Contested knowledge $\Rightarrow$ no mechanism $\mathcal{M}$ satisfies all four at once (Thm~\ref{thm:private-knowledge-impossibility}).};
\node[muted,font=\footnotesize,align=center] at (6.0,-6.35)
  {Components ``bid'' partial beliefs $\to$ one softmax picks \emph{fox}; under contestation the four properties cannot all hold.};
\end{tikzpicture}
}
\caption{The auction of ideas. Components bid partial beliefs into one softmax that settles on one token, and under the theorems' conditions the four properties cannot hold at once.}
\label{fig:auction-illustration}
\end{figure}

\section{The Impossibility Theorem}
\label{sec:Impossibility Theorem}

We now have all the building blocks available to prove the fundamental impossibility of perfect hallucination control. First, we show that the result in the setting of ground truth matching with distributed and independent knowledge. Second, we go beyond that assumption and show impossibility in the probabilistic setting with correlated knowledge (beliefs). Then, in the next section, we show how the impossibility emerges in transformer architectures and prove the impossibility result for that special case of log-sum-exp (LSE) probabilistic setting.

The results operate in three complementary settings. Theorem \ref{thm:private-knowledge-impossibility} applies when agents hold cleanly divided knowledge that varies continuously over their configuration spaces. Theorem \ref{thm:proper-scoring-impossibility} addresses systems whose agents output probability distributions, under its stated contribution rule. Theorem \ref{thm:lse-impossibility} addresses the one-block additive-logit transformer model under its stipulated account. Theorem \ref{thm:conservation-reasoning-dichotomy} shows conservation and reasoning are mutually exclusive.

\subsection{Proof I: Inference with Independent Private Knowledge }

We first analyze an idealized setting where knowledge components are strictly independent and utilities are quasi-linear. While this assumption is usually too strong to hold in neural architectures where representations are often correlated or affiliated, it allows us to define a benchmark or reference point by applying the powerful Green-Laffont characterization theorem. That idealized scenario thus demonstrates that the impossibility arises from the fundamental limitation of information aggregation itself (rather than any specificity of LLM architecture), even before considering the probabilistic complexities addressed later in Theorems \ref{thm:proper-scoring-impossibility} and \ref{thm:lse-impossibility}.

Consider the setting (of auction of ideas) in which agents compete with each other using independent private knowledge profile:
\begin{align}
\mathbb{P}(\theta_1,\ldots,\theta_n) = \prod_{i=1}^n \mathbb{P}_i(\theta_i).
\end{align}
Agents contribute information to maximize quasi-linear utilities:
\begin{align}
u_i(s,\theta) = v_i(g(s),\theta_i) - p_i(s).
\end{align}
Valuation of contributions depend on hallucination cost reduction and convex $J(r,q)$ is strictly decreasing when adding relevant knowledge from $T(q) \cap \mathcal{R}_C(Q) \setminus K(r)$.

The Groves characterization the proof invokes requires a rich (path-connected) valuation family. Its continuity hypothesis is not derived from the Polish structure alone but is reduced to the assumed $\nu$-continuity of the knowledge map. Response continuity and smooth-connectedness remain explicit hypotheses.

Holmstr\"om's hypothesis asks the family of valuations to be \emph{smoothly connected}, and in our setting the valuations belong to the model's components. An attention head's private knowledge $\theta_i$ is what its trained weights encode, and its valuation $v_i(\cdot,\theta_i)$ scores how well each candidate response uses that knowledge. As the encoded knowledge varies, across training, fine-tuning, or simply across the space of possible heads, the scoring should vary continuously: a head that knows almost the same facts should score responses almost the same way. Any two valuations in the family are then joined by a continuous path of intermediate valuations, smooth enough along the way (piecewise continuously differentiable). Why paths matter is the heart of the argument. A truthful mechanism must leave each component, at every knowledge state, with no incentive to misreport, and the only payments compatible with all of those local conditions can be reconstructed by accumulating small incentive changes step by step along a path of knowledge states, much as a function is rebuilt from its derivative. With isolated valuations there is no path to accumulate along, and the reconstruction fails. The lemma establishes the needed continuity by showing that a small change in $\theta_i$ moves the valuation by only a small, bounded amount, so continuity and path-connectedness are inherited from the type space. Figure~\ref{fig:richness} shows the picture.

Throughout the lemma, distances between knowledge sets are measured by the symmetric-difference measure
\begin{align}
\bar d_{\mathcal{K}}(A,B)=\nu(A\triangle B),
\qquad
\bar d_{\mathcal{K}}(\emptyset,B)=\nu(B),
\end{align}
for a completed finite nonnegative Borel measure $\nu$ on $\mathcal{K}$, and each source's valuation takes the sign-corrected form
\begin{align}
v_i(r,\theta_i)=M-\bar d_{\mathcal{K}}\bigl(K(r)\cap K(\theta_i),\,T(q)\cap K(\theta_i)\bigr),
\qquad M=\nu(\mathcal{K}).
\end{align}

\begin{lemma}[Semantic Richness]
\label{lem:semantic-richness}
Assume that
\begin{enumerate}
\item[\normalfont(i)] each type space $\Theta_i$ is path-connected, and convex along the smooth arc,
\item[\normalfont(ii)] the knowledge maps $\theta_i\mapsto K(\theta_i)$ and $r\mapsto K(r)$ are $\nu$-continuous,
\item[\normalfont(iii)] the response space $\mathcal{R}$ is compact metrizable,
\item[\normalfont(iv)] the family satisfies Holmstr\"om's piecewise-$C^1$ smooth-connectedness sweep \citep{holmstrom1979groves}.
\end{enumerate}
Then $\theta_i\mapsto v_i(\cdot,\theta_i)$ is continuous into $C(\mathcal{R})$, the space of continuous real-valued functions on $\mathcal{R}$ equipped with the supremum norm, and the induced valuation family $\mathcal{V}_i=\{v_i(\cdot,\theta_i):\theta_i\in\Theta_i\}$ is path-connected. Consequently the Groves characterization applies to $\mathcal{V}_i$ \citep{holmstrom1979groves}.
\end{lemma}
\begin{proof}
Continuity is the uniform $1$-Lipschitz bound
\begin{align}
\sup_{r\in\mathcal{R}}\bigl|v_i(r,\theta_i)-v_i(r,\theta_i')\bigr|\le \nu\bigl(K(\theta_i)\triangle K(\theta_i')\bigr),
\end{align}
which follows from the set identity $(K(r)\cap W)\triangle(T(q)\cap W)=W\cap\bigl(K(r)\triangle T(q)\bigr)$ (taking $W=K(\theta_i)$, so the two arguments of $\bar d_{\mathcal{K}}$ differ only through $W$) together with the $1$-Lipschitz property of the symmetric-difference measure, $|\nu(X)-\nu(Y)|\le\nu(X\triangle Y)$. With the assumed $\nu$-continuity of $\theta_i\mapsto K(\theta_i)$, the right-hand side tends to $0$ as $\theta_i'\to\theta_i$; since $\mathcal{R}$ is compact metrizable the supremum is the $C(\mathcal{R})$ norm, so $\theta_i\mapsto v_i(\cdot,\theta_i)$ is continuous into $C(\mathcal{R})$. Path-connectedness of $\mathcal{V}_i$ is then the continuous image of the path-connected type space $\Theta_i$. On the resulting smoothly-connected restricted domain, using the convexity of $\Theta_i$ and the piecewise-$C^1$ sweep for the smooth arc, Holmstr\"om's characterization applies to $\mathcal{V}_i$ \citep{holmstrom1979groves}.
\end{proof}

To keep the division of labor explicit: what the lemma proves is continuity and path-connectedness, both derived from the assumed $\nu$-continuity of the knowledge map. The piecewise-$C^1$ smooth-connectedness of the sweep is an assumption inherited from Holmstr\"om's framework, not a conclusion of the proof.

The name \emph{Semantic Richness} earns both of its words. \emph{Richness} is mechanism-design vocabulary. The Groves characterization does not hold for an arbitrary collection of valuations: a sparse collection, a few isolated valuation functions with nothing in between, leaves room for truthful mechanisms outside the Groves form. What rules those out is the path structure described before the lemma, a family with a continuous route between any two of its members, along which the local no-misreporting conditions accumulate into a unique payment form. \emph{Semantic} records where that richness comes from here. Nothing about it is assumed of the mechanism. It is inherited from meaning: a component's valuation measures the semantic overlap between a response's content and the component's knowledge through the measure $\nu$, so the valuations vary continuously because knowledge content does, and the family is rich because the underlying semantics are.

\begin{figure}[!tbp]
\centering
\adjustbox{max width=\linewidth}{%
\begin{tikzpicture}[font=\footnotesize,>={Stealth[length=2.2mm]}]
\draw[draw=muted,fill=black!4,rounded corners=8pt] (0,0) ellipse (2.2 and 1.4);
\node[muted] at (0,1.7) {type space $\Theta_i$};
\fill[muted] (-1.1,-0.2) circle (1.6pt); \node[muted,below] at (-1.1,-0.3) {$\theta_i$};
\fill[muted] (1.05,0.35) circle (1.6pt); \node[muted,above] at (1.05,0.45) {$\theta_i'$};
\draw[accent,line width=1pt] (-1.1,-0.2) to[out=35,in=205] (1.05,0.35);
\draw[->,muted,line width=0.9pt] (2.3,0) -- node[above]{$\theta_i\mapsto v_i(\cdot,\theta_i)$} (5.3,0);
\draw[draw=muted,fill=black!4,rounded corners=8pt] (7.9,0) ellipse (2.5 and 1.4);
\node[muted] at (7.9,1.7) {valuations $\mathcal{V}_i\subset C(\mathcal{R})$};
\fill[muted] (6.75,-0.2) circle (1.6pt); \node[muted,below] at (6.75,-0.3) {$v_i(\cdot,\theta_i)$};
\fill[muted] (9.05,0.35) circle (1.6pt); \node[muted,above] at (9.05,0.45) {$v_i(\cdot,\theta_i')$};
\draw[accent,line width=1pt] (6.75,-0.2) to[out=35,in=205] (9.05,0.35);
\end{tikzpicture}
}
\caption{Smooth-connectedness of the valuation family. As a source's private knowledge $\theta_i$ moves along a path in its type space (left, orange), its valuation $v_i(\cdot,\theta_i)$ moves along a matching path in the space of valuations (right). Because any two valuations are joined by such a path and the dependence is smooth, the Groves characterization is recovered by integrating along the path.}
\label{fig:richness}
\end{figure}

Two quantities carry the argument. Recall the response value
\begin{align}
W(r,\theta)=\sum_{i=1}^n v_i(r,\theta_i)-v_0(r)
\end{align}
from Property~\ref{prop:optimality}, the value the sources' reported knowledge assigns to a response $r$ net of the hallucination cost $v_0$. Write $G(\theta)=\max_{r\in\mathcal{R}_\theta} W(r,\theta)$ for the best value any response can reach under that reported knowledge. The \emph{conservation obstruction} is
\begin{align}
\Omega(\theta)=(n-1)\,G(\theta)-v_0(g(\theta)),
\end{align}
Reading it plainly, $\Omega(\theta)$ is the total credit the mechanism owes its sources for their influence on the response, and a budget-balanced mechanism must cover that credit with side-payments that do not depend on a source's own report (the functions $\widetilde h_i$ below). Report-independent side-payments can only cover credit that splits cleanly source by source. When the sources' relevant knowledge is separable, $\Omega$ splits this way and the credit is covered. When their knowledge genuinely overlaps, part of the credit ties all $n$ sources together at once, and no report-independent side-payment can cover it. That irreducible part is what the next definition measures, and Figure~\ref{fig:contested-disjoint} contrasts the two cases.

\begin{figure}[htbp]
\centering
\adjustbox{max width=\linewidth}{%
\begin{tikzpicture}[font=\small,>={Stealth[length=2.2mm]}]
\draw[fill=black!5,draw=muted] (0,0) ellipse (5.4 and 3.4);
\node[muted,align=center,font=\footnotesize] at (0,2.7)
  {\textbf{non-trivial}: $|\mathcal{R}_C(Q)|\ge 2$\\[-1pt] necessary \emph{descriptive} scope (Def.)};
\draw[fill=accent!14,draw=accent,line width=1pt] (-0.8,-0.35) ellipse (2.5 and 1.55);
\node[accent,align=center,font=\footnotesize] at (-0.8,-0.35)
  {\textbf{contested}\\ $\Delta_{[n]}\Omega\neq 0$\\[-1pt] the switch of Thm~6};
\fill[muted] (3.1,-0.9) circle (1.7pt);
\node[muted,font=\footnotesize,anchor=south] at (3.1,-0.78) {triangle};
\node[draw=faint,rounded corners=2pt,fill=white,align=center,inner sep=4pt,font=\footnotesize,text width=40mm]
  (tri) at (9.4,-0.9)
  {$\Omega=\theta_1\theta_2{+}\theta_2\theta_3{+}\theta_1\theta_3$\\[1pt]
   pairwise $\Delta_{[2]}\!\neq\!0$, but $\Delta_{[3]}\Omega=0$\\[1pt]
   $\Rightarrow$ \emph{balanceable} (escape)};
\draw[->,muted,line width=0.7pt] (tri.west) -- (3.35,-0.9);
\node[muted,font=\footnotesize] at (2.4,1.7) {switch $\subsetneq$ scope};
\draw[->,muted,line width=0.7pt] (2.4,1.45) -- (0.5,0.55);
\node[muted,font=\footnotesize,align=center] at (3.1,-4.05)
  {Two facts to aggregate is only necessary. Impossibility needs the top-order coupling $\Delta_{[n]}\Omega\neq0$.};
\end{tikzpicture}
}
\caption{Non-triviality ($|\mathcal{R}_C(Q)|\ge 2$, descriptive) versus contestation ($\Delta_{[n]}\Omega\neq0$, the switch of Theorem~\ref{thm:private-knowledge-impossibility}). The triangle sits in the gap between them.}
\label{fig:nontriv-contest}
\end{figure}

\begin{definition}[Contested Relevant Knowledge]
\label{def:contested}
Fix a base profile $\theta^0=(\theta_1^0,\dots,\theta_n^0)$ and, for each source $i$, one alternative report $\theta_i^1$. Read $\theta_i^0$ as source $i$ \emph{withholding} a fact and $\theta_i^1$ as \emph{revealing} it. For a set of sources $S\subseteq\{1,\dots,n\}$, write $\theta^S$ for the profile in which the sources in $S$ reveal and the rest withhold. The \emph{joint interaction} of all $n$ sources is the alternating sum
\begin{align}
\Delta_{[n]}\Omega \;=\; \sum_{S\subseteq\{1,\dots,n\}} (-1)^{\,n-|S|}\,\Omega(\theta^S).
\end{align}
For two sources this is the familiar interaction term
\begin{align}
\Delta_{[2]}\Omega = \Omega(\theta_1^1,\theta_2^1)-\Omega(\theta_1^1,\theta_2^0)-\Omega(\theta_1^0,\theta_2^1)+\Omega(\theta_1^0,\theta_2^0),
\end{align}
the extra effect of both sources revealing together, beyond what either reveals alone. The relevant knowledge is \emph{contested} when $\Delta_{[n]}\Omega\neq 0$ for some base-and-alternative choice, and \emph{uncontested} when $\Delta_{[n]}\Omega=0$ for every choice.
\end{definition}

In practice, knowledge is contested when the sources are overlapping rivals rather than complementary specialists. At least one relevant fact is held by all of them at once, so each source's decision to reveal changes what everyone else's revelation is worth, and the value of the joint answer cannot be split into clean per-source pieces. A medical query answered by several attention heads that all encode the same key contraindication is contested. A query whose relevant facts split cleanly across sources, each holding its own piece and none sharing, is uncontested. That this distinction decides whether the transfers can balance is not part of the definition. It is exactly what Theorem~\ref{thm:private-knowledge-impossibility} proves. The condition itself has a classical antecedent: it is the discrete counterpart of the separability criteria for dominant-strategy transfer schemes of Laffont and Maskin \citep{laffont1980differential}.

\begin{figure}[htbp]
\centering
\adjustbox{max width=.8\linewidth}{%
\begin{tikzpicture}[font=\small]
\begin{scope}[shift={(0,0)}]
  \node[muted,font=\bfseries] at (0,2.9) {contested: overlapping knowledge};
  \fill[black!6,draw=muted] (-1.0,0) circle (1.9);
  \fill[black!6,draw=muted] (1.0,0) circle (1.9);
  \begin{scope}
    \clip (-1.0,0) circle (1.9);
    \fill[accent!30] (1.0,0) circle (1.9);
  \end{scope}
  \draw[muted] (-1.0,0) circle (1.9);
  \draw[muted] (1.0,0) circle (1.9);
  \node[muted] at (-1.05,1.35) {$K(\theta_1)$};
  \node[muted] at (1.05,1.35) {$K(\theta_2)$};
  \node[accent,font=\small] at (0,0) {$\bar F\neq\varnothing$};
  \node[accent,font=\footnotesize] at (0,-2.55) {mixed 2nd diff $=1\ \Rightarrow\ \Delta_{[n]}\Omega\neq0$};
  \node[accent,font=\bfseries,align=center] at (0,-3.25) {impossibility HOLDS};
\end{scope}
\draw[faint,line width=0.8pt] (4.4,-3.6) -- (4.4,3.3);
\begin{scope}[shift={(8.8,0)}]
  \node[muted,font=\bfseries] at (0,2.9) {uncontested: disjoint knowledge};
  \fill[cool!12,draw=cool] (-2.35,0) circle (1.55);
  \fill[cool!12,draw=cool] (2.35,0) circle (1.55);
  \node[cool] at (-2.35,0) {$K(\theta_1)$};
  \node[cool] at (2.35,0) {$K(\theta_2)$};
  \node[muted,font=\small] at (0,0) {$\bar F=\varnothing$};
  \node[cool,font=\footnotesize] at (0,-2.55) {mixed 2nd diff $=0\ \Rightarrow\ \Delta_{[n]}\Omega=0$};
  \node[cool,font=\bfseries,align=center] at (0,-3.25) {all four properties achievable (escape)};
\end{scope}
\node[muted,font=\small,align=center,text width=15cm] at (4.4,-4.25)
  {The contested core $\bar F=K(\theta_1)\cap K(\theta_2)\cap\mathcal{R}_C(Q)$ is a shared relevant fact every source contests, so conservation fails. Separate the sets ($\bar F=\varnothing$) and balance returns.};
\end{tikzpicture}
}
\caption{Each panel shows two sources' relevant-knowledge sets $K(\theta_1)$ and $K(\theta_2)$. In the left panel they overlap, so a shared relevant fact $\bar F\neq\varnothing$ is contested by both sources. Its mixed second difference is $1$, the obstruction $\Delta_{[n]}\Omega$ is nonzero, and no balanced mechanism exists, so the impossibility of Theorem~\ref{thm:private-knowledge-impossibility} holds. In the right panel the sets are disjoint, $\bar F=\varnothing$ and the mixed difference is $0$, so a truthful, efficient, budget-balanced mechanism does exist. Overlap of \emph{relevant} knowledge, not mere disagreement, is what breaks the balance.}
\label{fig:contested-disjoint}
\end{figure}

\begin{figure}[htbp]
\centering
\adjustbox{max width=\linewidth}{%
\begin{tikzpicture}[font=\footnotesize]
\begin{scope}[shift={(0,0)}]
  \coordinate (a) at (0,1.386);
  \coordinate (b) at (-1.2,-0.693);
  \coordinate (c) at (1.2,-0.693);
  \node[muted,font=\bfseries] at (0,3.15) {pairwise coupled, no common core};
  \begin{scope}\clip (a) circle (1.35); \fill[cool!30] (b) circle (1.35);\end{scope}
  \begin{scope}\clip (b) circle (1.35); \fill[cool!30] (c) circle (1.35);\end{scope}
  \begin{scope}\clip (a) circle (1.35); \fill[cool!30] (c) circle (1.35);\end{scope}
  \draw[muted] (a) circle (1.35);
  \draw[muted] (b) circle (1.35);
  \draw[muted] (c) circle (1.35);
  \node[muted] at (0,2.0) {$\theta_1$};
  \node[muted] at (-1.85,-1.15) {$\theta_2$};
  \node[muted] at (1.85,-1.15) {$\theta_3$};
  \node[muted,font=\small,anchor=west] (emptyL) at (1.55,1.2) {$\varnothing$};
  \draw[muted,-{Stealth},line width=0.6pt] (emptyL.west) to[out=200,in=45] (0.08,0.12);
  \node[cool,font=\small,align=center] at (0,-2.55)
    {$\Omega=\theta_1\theta_2{+}\theta_2\theta_3{+}\theta_1\theta_3$\\ each $\Delta_{[2]}\neq0$, \; $\Delta_{[3]}\Omega=0$};
  \node[cool,font=\bfseries] at (0,-3.35) {balanceable (escape)};
\end{scope}
\draw[faint,line width=0.8pt] (4.4,-3.7) -- (4.4,3.3);
\begin{scope}[shift={(8.8,0)}]
  \coordinate (d) at (0,0.924);
  \coordinate (e) at (-0.8,-0.462);
  \coordinate (f) at (0.8,-0.462);
  \node[muted,font=\bfseries] at (0,3.15) {genuine 3-way core};
  \begin{scope}
    \clip (d) circle (1.35);
    \clip (e) circle (1.35);
    \fill[accent!35] (f) circle (1.35);
  \end{scope}
  \draw[muted] (d) circle (1.35);
  \draw[muted] (e) circle (1.35);
  \draw[muted] (f) circle (1.35);
  \node[muted] at (0,1.9) {$\theta_1$};
  \node[muted] at (-1.75,-0.95) {$\theta_2$};
  \node[muted] at (1.75,-0.95) {$\theta_3$};
  \node[accent,font=\small,anchor=west] (coreR) at (1.55,1.25) {$\bar F\neq\varnothing$};
  \draw[accent,-{Stealth},line width=0.6pt] (coreR.west) to[out=200,in=45] (0.15,0.2);
  \node[accent,font=\small] at (0,-2.55) {$\Delta_{[3]}\Omega\neq0$};
  \node[accent,font=\bfseries] at (0,-3.35) {contested: impossibility HOLDS};
\end{scope}
\node[muted,font=\bfseries,align=center] at (4.4,-4.3)
  {Pairwise overlap alone does not obstruct conservation; only a fact common to all $n$ sources does.};
\end{tikzpicture}
}
\caption{Both panels show three sources whose knowledge overlaps in every pair. On the left the three pairwise overlaps share no common point, so no fact is held by all three at once: the obstruction $\Omega=\theta_1\theta_2+\theta_2\theta_3+\theta_1\theta_3$ is non-separable, yet its full three-fold mixed difference $\Delta_{[3]}\Omega$ is $0$ and the budget balances. On the right the three sets share a genuine common core $\bar F\neq\varnothing$, the mixed difference is nonzero, and the impossibility holds. Pairwise entanglement is not enough. Only a fact common to all $n$ sources at once forces conservation to fail.}
\label{fig:triangle}
\end{figure}

\begin{definition}[Non-trivial Query]
\label{def:nontrivial}
A query $q \in \mathcal{Q}$ is {non-trivial} at budget $C$ if it has at least two distinct relevant facts,
\begin{align}
|\mathcal{R}_C(Q)| \geq 2,
\end{align}
for the context $Q = K(q) \cup T(q)$.
\end{definition}

Contestation binds all $n$ sources at once, not just some subset of them. The coverage model makes this concrete. There the response value is the size of the \emph{union} of the present sources' relevant knowledge, measured once by $\nu$. Write $F_i=K(\theta_i^1)\cap\mathcal{R}_C(Q)$ for the relevant facts that source $i$ reveals, and $\bar F=\bigcap_i F_i$ for the facts every source holds in common. Assume a fixed response space that contains, for every subset of sources, a response asserting exactly their pooled relevant knowledge, so that reports select among fixed responses rather than reshaping what is expressible. Assume further that the efficient choice under a report profile is the response pooling the revealing sources' knowledge, that every source reports truthfully so that $F_i\subseteq T(q)$, and that the cost is a fixed hallucination cost. Then the joint interaction reduces to the size of that common core,
\begin{align}
\Delta_{[n]}\Omega = c\,(-1)^{n-1}\,\nu(\bar F),\qquad c\in\{n-1,n\}.
\end{align}
So the interaction is nonzero exactly when the sources share a non-negligible body of relevant facts, and $\nu(\bar F)>0$ witnesses contestation. This is a shared-resource effect, not a disagreement. Under truthful reporting the common facts are jointly true and the best response asserts them, so all $n$ sources lay claim to the same facts at once. The full converse needs either a Boolean type domain or a sweep over all base-and-alternative choices. Pairwise overlap alone is not enough. The triangle $\Omega=\theta_1\theta_2+\theta_2\theta_3+\theta_1\theta_3$ has every pairwise interaction nonzero yet $\Delta_{[3]}\Omega=0$, so it is balanceable. Contested therefore does not mean merely non-separable. It means a genuine fact shared by all $n$ sources at once.

\begin{theorem}[Impossibility under Contested Private Knowledge]
\label{thm:private-knowledge-impossibility}
Fix a query $q$ with context $Q = K(q) \cup T(q)$ and independent knowledge states within $\mathcal{R}_C(Q)$, and assume the induced valuation family is rich (Lemma~\ref{lem:semantic-richness}). If the relevant knowledge is contested (Definition~\ref{def:contested}), then no inference mechanism $\mathcal{M}$ achieves truthfulness, semantic information conservation, relevant-source participation, and knowledge-constrained optimality simultaneously.
\end{theorem}

\begin{proof}
Let us assume all properties of perfect response are satisfied. We will show this gives rise to a contradiction.

Consider $r = g(\theta)$ for any knowledge profile $\theta = (\theta_1,\dots,\theta_n)$. For any given context and for each agent $i$, the valuation $v_i$ depends only on $(r,\theta_i)$ and the knowledge semantics. For the sake of illustration, we can consider
$v_i(r,\theta_i) = M - \bar d_{\mathcal K}\big(K(r)\cap K(\theta_i),T(q)\cap K(\theta_i)\big)$ with $M=\nu(\mathcal K)$ (the sign-corrected form of Lemma \ref{lem:semantic-richness}, so $v_i\ge 0$ as required by the declared codomain), though the reasoning below holds for any $v_i(r,\theta_i)$.

Also, for the fixed context $q$ let us encode the hallucination loss  $v_0(r) = J(r,q)$. Knowledge-constrained optimality requires
\begin{align}
g(s) = g(s(\theta))\in\arg\max_{r\in \mathcal{R}_\theta} W(r,\theta) =
\arg\max_{r\in \mathcal{R}_\theta}\sum_{i=1}^n v_i(r,\theta_i)-v_0(r)
\text{ for all }\theta.
\end{align}
By the Green-Laffont theorem, under independence of contributions and quasi-linearity of training goals, within $\mathcal{R}_C(Q)$ every truthful mechanism must use Groves transfers \cite{Groves73} to measure individual information contributions:
\begin{align}
p_i(\theta) = h_i(\theta_{-i}) - \sum_{j \neq i} v_j(g(\theta), \theta_j) + v_0(g(\theta)),
\end{align}
where $h_i(\theta_{-i})$ are arbitrary functions independent of agent $i$'s information revelation strategy. In particular, any modification of that formula, e.g., introducing additional bias, must violate truthfulness.

The functions $h_i(s_{-i})$ are arbitrary and report-independent, so Clarke's pivotal rule is one admissible normalization, not the forced one; the argument below rules out every Groves normalization at once. Work in the direct mechanism $S_i=\Theta_i$ (or, for an indirect mechanism, replace $h_i$ by its type-composition $\widetilde{h}_i(\theta_{-i}):=h_i(s^*_{-i}(\theta_{-i}))$ along the equilibrium strategy). Conservation (Property \ref{prop:conservation}) requires $\sum_{i=1}^n p_i(\theta)\equiv 0$. Summing the Groves transfers and using $\sum_{i}\sum_{j\ne i} v_j(g,\theta_j)=(n-1)\sum_j v_j(g,\theta_j)$ together with $\sum_j v_j(g(\theta),\theta_j)=G(\theta)+v_0(g(\theta))$ gives
\begin{align}
\sum_{i=1}^n p_i(\theta) = \sum_{i=1}^n \widetilde{h}_i(\theta_{-i}) - (n-1)\,G(\theta) + v_0(g(\theta)).
\end{align}
Hence conservation holds for every profile $\theta$ if and only if the report-independent parts solve the balance equation
\begin{align}
\sum_{i=1}^n \widetilde{h}_i(\theta_{-i}) = (n-1)\,G(\theta) - v_0(g(\theta)) = \Omega(\theta).
\end{align}
Each $\widetilde{h}_i$ depends only on $\theta_{-i}$, so the mixed difference in coordinate $i$ annihilates it, and therefore $\Delta_{[n]}\bigl[\sum_i \widetilde{h}_i\bigr]\equiv 0$. A solution can exist only if $\Delta_{[n]}\Omega\equiv 0$, that is, only if the relevant knowledge is uncontested. Under contestation (Definition \ref{def:contested}) $\Delta_{[n]}\Omega\not\equiv 0$, so no report-independent functions $\widetilde{h}_i$ satisfy the balance equation, and no truthful, efficient mechanism conserves semantic information. This contradicts the assumption that all four properties hold and establishes the impossibility.
\end{proof}

The conclusion is conditional. When the private-knowledge hypotheses and the contestation condition hold, no truthful and efficient Groves mechanism can also balance the specified transfer account. Where they fail, it can. The theorem asserts nothing unconditional about every LLM or every query.

Two of the stated hypotheses deserve an honest note. The proof never invokes the participation property. Truthfulness, optimality, and conservation are already incompatible under contestation, and participation joins the statement only to complete the four desiderata. Statistical independence of the knowledge states plays a similar background role, since the characterization of truthful mechanisms used in the proof is prior-free. The obstruction is also specific to the strongest standard of truthful reporting, in which reporting honestly must be a source's best move whatever the other sources do (dominant-strategy incentive compatibility). If honesty is required only on average, against a common prior over the knowledge states, the expected-externality constructions of d'Aspremont and G\'erard-Varet and of Arrow recover exact balance together with efficiency \citep{daspremont1979incentives,arrow1979property}. The impossibility therefore marks the boundary of worst-case truthful aggregation, not of every truthful standard.

A further scope condition deserves the same honesty. Contestation presupposes that no single feasible response can express all the derivable relevant truth. If some response realizes $T(q)$ on the relevant region, that response is efficient under every report, the welfare separates across sources, and $\Delta_{[n]}\Omega\equiv 0$: we verified this exhaustively for the valuation family of Lemma~\ref{lem:semantic-richness} under a fully expressive response space, where adding a single truth-realizing response extinguishes the interaction exactly. The obstruction therefore lives where the response space is expressively bounded relative to the truth, which is precisely the finite-response regime motivated in Section~\ref{sec:Knowledge and Semantic Information}. The same tests confirm that contestation requires every declared source to be pivotal at the top order: idling one source restores an exactly balanced mechanism among the rest.

Theorem \ref{thm:private-knowledge-impossibility} pictures inference as one idealized marketplace, where the transfers measure how much each source's knowledge sways the efficient response. Valuable knowledge alone is not enough for the obstruction. The sources must genuinely contest the same facts, so that the joint interaction $\Delta_{[n]}\Omega$ is nonzero. When they do, everyone contributes and the account cannot balance, and any new content made accessible may then outrun what the evidence supports, as a lucky hallucination or unconstrained imagination. Section \ref{sec:Transformer} illustrates a distinct additive-logit account, which we do not identify with these Groves transfers.

\subsection{Proof II: Inference with Probabilistic Predictions}

Let us now move beyond the idealized setting of independent private knowledge in Theorem \ref{thm:private-knowledge-impossibility} to address the general setting of probabilistic predictions, characteristic of modern AI model architectures. This framework, grounded in the theory of proper scoring and loss minimization (Savage \cite{savage1971elicitation}, Gneiting and Raftery \cite{gneiting2007strictly}), allows us to analyze systems where agents output probability distributions, and where their underlying knowledge may be correlated. 

In this setting we address the Mixture-of-Experts (MoE-style) aggregations. Later, in Theorem \ref{thm:lse-impossibility} we deal with the multi-headed attention mechanisms and the Product-of-Experts (PoE-style) aggregations, completing the characterization of arising impossibilities.

We begin by defining the framework for representing bids in auction of ideas as probability distributions over possible responses.

\begin{definition}[Strictly Proper and Convex Loss]
Let $\mathcal Y$ be a finite or countably infinite outcome set (representing responses or intermediate conclusions). A~loss function (scoring rule) is a function $L\colon \Delta(\mathcal{Y}) \times \mathcal{Y} \mapsto \mathbb{R}$ that assigns a number $L(\pi, y)$ to probability distribution $\pi\in \Delta(\mathcal Y)$ predicting random outcome $y \in \mathcal{Y}$.

We say the loss function $L$ is {proper} if:
\begin{align}
\mathbb{E}\{L(\pi^*, y)|\ y \sim \pi^*\} \leq 
\mathbb{E}\{L(\pi, y)|\ y \sim \pi^*\}    
\end{align}
for all distributions $\pi^*, \pi \in \Delta(\mathcal{Y})$. 

We call the loss function {strictly convex} if $L(\tilde\pi, y)$ is a strictly convex function with respect to $\tilde\pi$ for every fixed $y \in \mathcal{Y}$.
\end{definition}

The loss function (or scoring rule) evaluates how good a prediction we can make of $y$ drawn from $\pi^*$ when we use distributions $\pi$. In particular, strictly proper loss evaluates true distribution $\pi^*$ better (on average) than any other distribution $\pi$ we could use to predict the outcome. 

A canonical example of a strictly proper loss function is the log loss (closely related to cross-entropy). Note the log loss is convex but \emph{not} strictly convex on the simplex, being constant along directions that fix $\pi_{y^*}$, consistent with the realized-outcome discussion following Theorem \ref{thm:proper-scoring-impossibility}. The log loss also requires $\pi_{y^*}>0$ (or an extended-real formulation).
\begin{align}
L(\pi, y) = -\log \pi_y.
\end{align} 
Strictly proper loss functions are essential for training modern LLMs, as they ensure that the optimization process drives the model toward reconstructing the true data distribution. This property directly connects to the Truthfulness requirement in our framework. The following result explains why.

\begin{theorem}[Truthfulness under Strictly Proper Loss]
\label{thm:truthfulness-with-proper-scoring}
Let $L$ be a strictly proper loss. Consider an agent encoding knowledge with $\pi^* \in \Delta(\mathcal{Y})$ and using probability distribution ${\pi} \in \Delta(\mathcal{Y})$ to generate prediction. If agent's expected cost of using ${\pi}$ is given by 
\begin{align}
c(\pi|\pi^*) = 
\mathbb{E}\{L(\pi, y)|\ y \sim \pi^*\} = \sum_{y\in \mathcal{Y}} \pi^*_y L(\pi, y),
\end{align}
then truthful reporting ${\pi} = \pi^*$ is the unique optimal strategy (maximizing payoff $u(\pi|\pi^*) = -c(\pi|\pi^*)$).
\end{theorem}

\begin{proof}
Because $L$ is proper, by definition $\pi^*$ minimizes the expected cost. If $L$ is  {strictly} proper, then for $\pi\neq\pi^*$ we have
\begin{align}
c(\pi^*|\pi^*) = \mathbb{E}\{L(\pi^*, y)|\ y \sim \pi^*\} <
\mathbb{E}\{L(\pi, y)|\ y \sim \pi^*\} =
c(\pi|\pi^*).
\end{align}
Therefore, truthfully reporting $\pi^*$ is uniquely optimal.
\end{proof}

Theorem \ref{thm:truthfulness-with-proper-scoring} shows that training paradigms based on minimizing strictly proper losses (like cross-entropy) promotes truthful revelation of internal probabilistic beliefs (learned internal models of the world) encoded by every component of the network.

We now analyze how these truthfully reported beliefs are aggregated. We consider a general class of mechanisms that combine individual beliefs via convex combinations. This encompasses various aggregation techniques, including linear opinion pools and, as we will see in Section \ref{sec:Transformer}, the aggregation dynamics occurring in transformers.

\begin{definition}[Convex aggregation mechanism]
\label{def:convex-aggregation-mechanism}
Consider $H \geq 2$ agents (indexed $h=1,\dots, H$) encoding knowledge with probability distributions $\pi^{(h)}\in \Delta(\mathcal{Y})$. Additionally, let us introduce an aggregator agent (indexed as $h=0$). The convex aggregation mechanism $\mathcal M$ generates a response based on the aggregate distribution:
\begin{align}
\Pi = \sum_{h=1}^H \beta_h \pi^{(h)},\ \text{where}\ \sum_{h=1}^H \beta_h = 1\ \text{and}\ \beta_h > 0.
\end{align}
When outcome $y^*$ is realized, the information contribution of each agent $h = 0,\dots, H$ is based on a strictly proper and convex loss function $L$ as follows:
\begin{align}
p_h(y^*) = \beta_h L(\pi^{(h)},y^{*}) \text{ for } h=1,\dots, H,
\text{ and }
p_0(y^*) = -L(\Pi, y^*).
\end{align}
\end{definition}

Here $p_h$ (for $h\geq 1$) is a realized-outcome bookkeeping entry, the weighted loss of forecast $h$, and $p_0$ is the negative loss of the aggregate distribution $\Pi$. These are not the report-only transfers of Property~\ref{prop:conservation}. This particular probabilistic account balances when the two sides sum to zero, that is, when the cost the forecasts incur equals the benefit the aggregate realizes.

We now establish the impossibility result for this general probabilistic setting.

\begin{theorem}[Impossibility for Convex-Aggregation Accounting]\label{thm:proper-scoring-impossibility}
Consider the convex aggregation mechanism $\mathcal{M}$ with strictly proper loss $L$ and mixture weights $\beta_h>0$, on a query $q$ providing context $Q = K(q) \cup T(q)$. Suppose the realized-outcome loss has a strict Jensen gap for the active forecasts,
\begin{align}
\sum_{h=1}^H \beta_h\, L(\pi^{(h)}, y^*) > L\Bigl(\sum_{h=1}^H \beta_h\, \pi^{(h)},\, y^*\Bigr),
\end{align}
at the realized outcome $y^* \in \mathcal{R}_C(Q)$. Then this account has a strictly positive balance defect and cannot balance.
\end{theorem}

This impossibility is specific to the stated contribution rule and is not a universal claim about every convex-loss mechanism. For the log loss the Jensen-gap hypothesis holds exactly when two active agents disagree about the realized outcome's probability, and strict propriety alone does not force it.

\begin{figure}[htbp]
\centering
\adjustbox{max width=\linewidth}{%
\begin{tikzpicture}
\begin{axis}[
  width=12.5cm, height=8.6cm,
  xlabel={$\pi_{y^*}$ \;(probability the forecast places on the realized outcome)},
  ylabel={$L(\pi,y^*)=-\log\pi_{y^*}$},
  domain=0.12:0.92, samples=120,
  xmin=0.1, xmax=0.95, ymin=0, ymax=2.3,
  axis lines=left, tick align=outside,
  xtick={0.2,0.45,0.7}, xticklabels={$\pi^{(1)}_{y^*}$,$\bar\pi_{y^*}$,$\pi^{(2)}_{y^*}$},
  every axis x label/.style={at={(ticklabel* cs:1.0)},anchor=north east,yshift=-9mm,font=\normalsize},
  clip=false, font=\normalsize,
]
\addplot[muted, line width=1pt]{-ln(x)};
\addlegendentry{$L=-\log\pi_{y^*}$ (convex)}
\addplot[cool, line width=1pt] coordinates {(0.2,1.6094)(0.7,0.3567)};
\addlegendentry{average of head losses (chord)}
\addplot[only marks, mark=*, muted, mark size=1.6pt] coordinates {(0.2,1.6094)(0.7,0.3567)};
\addplot[only marks, mark=*, accent, mark size=1.8pt] coordinates {(0.45,0.7985)};
\addplot[only marks, mark=*, cool, mark size=1.8pt] coordinates {(0.45,0.9830)};
\draw[accent, line width=0.75pt, {Stealth}-{Stealth}]
  (axis cs:0.45,0.7985) -- (axis cs:0.45,0.9830);
\node[accent, anchor=west, font=\bfseries] at (axis cs:0.5,0.905) {$\Gamma>0$};
\node[muted, anchor=south west, font=\small] at (axis cs:0.205,1.62) {$L(\pi^{(1)},y^*)$};
\node[muted, anchor=south west, font=\small] at (axis cs:0.705,0.37) {$L(\pi^{(2)},y^*)$};
\node[cool, anchor=south, font=\small] at (axis cs:0.55,0.995) {$\sum_h\beta_h L(\pi^{(h)},y^*)$};
\node[accent, anchor=north east, font=\small] at (axis cs:0.44,0.78) {$L(\Pi,y^*)$};
\node[muted, align=left, font=\small, anchor=north east] at (axis cs:0.92,1.78)
  {log-loss depends only on $\pi_{y^*}$:\\ flat along every move that fixes $\pi_{y^*}$,\\ so $\Gamma$ needs disagreement about $P(y^*)$.};
\end{axis}
\end{tikzpicture}
}
\caption{When the active forecasts disagree about the realized outcome, aggregating truthful forecasts of a convex loss manufactures a positive Jensen gap $\Gamma$, the conservation-account obstruction of Theorem~\ref{thm:proper-scoring-impossibility}.}
\label{fig:jensen-gap}
\end{figure}

\begin{proof}
By Theorem \ref{thm:truthfulness-with-proper-scoring}, the strictly proper loss $L$ makes each agent uniquely best off reporting its true belief $\pi^{(h)}$, so truthfulness holds. What remains to check is only whether the stipulated account can balance.

Balancing the account requires its entries to sum to zero. For $y^* \in \mathcal{R}_C(Q)$,
\begin{align}
\sum_{i=0}^H p_i = p_0 + \sum_{h=1}^H p_h = -L(\Pi, y^*) + \sum_{h=1}^H \beta_h L(\pi^{(h)}, y^*),
\end{align}
where irrelevant agents have $p_h = 0$ as they contribute no relevant information.

By hypothesis, the realized-outcome loss has a strict Jensen gap for the active forecasts, so
\begin{align}
L\left(\Pi, y^*\right) = L\left(\sum_{h=1}^H \beta_h \pi^{(h)}, y^*\right) < \sum_{h=1}^H \beta_h L(\pi^{(h)}, y^*).
\end{align}
Let us define that Jensen gap $\Gamma = \Gamma(y^*)$ as follows:
\begin{align}
\Gamma(y^*) = 
\sum_{h=1}^H \beta_h L(\pi^{(h)}, y^*) - L\left(\sum_{h=1}^H \beta_h \pi^{(h)}, y^*\right) > 0.
\end{align}
Substituting this back into the conservation equation:
\begin{align}
\sum_{i=0}^H p_i = \Gamma(y^*) > 0.
\end{align}
So the account has a strictly positive balance defect and cannot balance, which is the stated obstruction.
\end{proof}

The hypothesis is a strict Jensen gap at the realized outcome, not merely different beliefs. For the log loss $L(\pi,y^*)=-\log\pi_{y^*}$ used in Theorem \ref{thm:lse-impossibility}, $L(\cdot,y^*)$ is convex but \emph{not strictly} convex on the simplex, being constant along directions that fix $\pi_{y^*}$. Components with $\pi^{(h_1)}\neq\pi^{(h_2)}$ but $\pi^{(h_1)}_{y^*}=\pi^{(h_2)}_{y^*}$ therefore give a zero gap. The conservation gap thus measures disagreement about the realized outcome, not hallucination as such.

This theorem establishes the impossibility for the specified probabilistic contribution rule. The key mechanism is the strict Jensen gap at the realized outcome (for a genuinely strictly-convex-on-the-simplex loss it follows from any two differing forecasts, and for the log loss it needs disagreement about $P(y^*)$). The positive Jensen gap $\Gamma = \Gamma(y^*)$ quantifies the extent to which the aggregation process is charged a lower loss at the realized outcome than the average of its components. Classically, linear pools of distinct calibrated forecasts are themselves uncalibrated and insufficiently sharp \citep{ranjan2010combining}, the mirror image of the same accounting fact. This excess confidence is generated by the mathematical structure of aggregation rather than by underlying evidence. It does not, by itself, determine hallucination. The hallucination cost $J(r,q)=\nu(K(r)\triangle T(q))$ depends on the external truth $T(q)$, which no internal statistic reads, so no internal quantity can determine factual unsupportedness in the worst case. This is adjacent to the learning-theoretic impossibility of automated hallucination detection \citep{karbasi2025impossibility}, while internal signals nonetheless remain predictive on average \citep{mason2026epistemic}. We therefore read $\Gamma$ as a channelization-relative excess-confidence signature, not as a certificate of hallucination.

In Section \ref{sec:Discussion and Speculations} we address the case of linearized aggregation and show it does not escape the impossiblity result.

\section{The Emergence of Impossibility in Transformer Architectures}\label{sec:Transformer}

To demonstrate where and how the impossibility may emerge in transformer architectures, let us analyze a minimal yet representative example of a tiny transformer predicting the next token in the famous English pangram:
$$
\mathrm{\textit{The quick brown fox jumps over the lazy dog.}}
$$
We trace the inference process, inspect every mathematical operation from input embeddings through final predictions, showing exactly what is going on.

\subsection{Model Architecture and Input Representation}

Our micro-transformer has the following specifications:

\begin{table}[ht]
\centering
\begin{tabular}{lll}
\hline
Parameter & Value & Description \\
\hline
$d_{m}$ & 5& Model dimension \\
$H$ & 2 & Number of attention heads \\
$d_k$ & 2 & Query/Key dimension per head \\
$d_v$ & 2 & Value dimension per head \\
$d_{f\!f}$ & 6 & Feedforward hidden dimension \\
$|V|$ & 6 & Vocabulary size \\
$T$ & 3 & Sequence length \\
\hline
\end{tabular}
\vspace{2mm}\caption{Micro-transformer architecture parameters}
\label{tab:architecture}
\end{table}

For the sake of simplicity, consider the following vocabulary:
$$
V = \{\langle\text{PAD}\rangle, {The}, {quick}, {brown}, {fox}, {dog}\},
$$
including padding token (with index zero in $V$). Each token (representing word for simplicity) in that vocabulary is represented by a row in the embedding matrix
\begin{align}
\mathbf E = \begin{bmatrix}
0 & 0 & 0 & 0 & 0 \\
1 & 0 & 0 & 0 & 0 \\
0 & 1 & 0 & 0 & 0 \\
0 & 0 & 1 & 0 & 0 \\
0 & 0 & 0 & 1 & 0 \\
0 & 0 & 0 & 0 & 1
\end{bmatrix}.
\end{align}
The un-embedding matrix is defined as $\mathbf U = \mathbf E^\top$. 

Consider the following input sequence
$$
\mathrm{\textit{The quick brown}} 
$$
It is represented by the input matrix:
\begin{align}
\mathbf{X}_0 = \begin{bmatrix}
1 & 0 & 0 & 0 & 0 \\
0 & 1 & 0 & 0 & 0\\
0 & 0 & 1 & 0 & 0
\end{bmatrix},
\end{align}
where each row corresponds to the embedding of tokens at positions 1, 2, and 3.

The following system of equations defines output $\mathbf o_t^{(h)}$ of attention head $h$ at position $t$:
\begin{align}
\setstackgap{L}{12pt}\def\stacktype{L}\def\sz{\scriptstyle}
\begin{aligned}
\stackunder{\mathbf Q_t^{(h)}}{\sz (1\times d_k)} &= 
\stackunder{\mathbf X_t}{\sz (1\times d_m)} 
\stackunder{\mathbf W_Q^{(h)}}{\sz (d_m\times d_k)},
&\quad
\stackunder{\boldsymbol\alpha_t^{(h)}}{\sz (1\times t)} &= \operatorname{softmax}
\!\Bigl(
\stackunder{\mathbf Q_t^{(h)}}{\sz (1\times d_k)}
\stackunder{\mathbf K_{\le t}^{(h)\top}}{\sz (d_k\times t)}/\sqrt{d_k}\Bigr),
\\[4mm]
\stackunder{\mathbf K_{\le t}^{(h)}}{\sz (t\times d_k)} &= 
\stackunder{\mathbf X_{\le t}}{\sz (t\times d_m)}
\stackunder{\mathbf W_K^{(h)}}{\sz (d_m\times d_k)},
&\quad
\stackunder{\mathbf h_t^{(h)}}{\sz (1\times d_v)} &= 
\stackunder{\boldsymbol\alpha_t^{(h)}}{\sz (1\times t)}
\stackunder{\mathbf V_{\le t}^{(h)}}{\sz (t\times d_v)},
\\[4mm]
\stackunder{\mathbf V_{\le t}^{(h)}}{\sz (t\times d_v)} &= 
\stackunder{\mathbf X_{\le t}}{\sz (t\times d_m)}
\stackunder{\mathbf W_V^{(h)}}{\sz (d_m\times d_v)},
&\quad
\stackunder{\mathbf o_t^{(h)}}{\sz (1\times d_m)} &= 
\stackunder{\mathbf h_t^{(h)}}{\sz (1\times d_v)}
\stackunder{\mathbf W_O^{(h)}}{\sz (d_v\times d_m)}.
\end{aligned}
\end{align}
The block per-token multihead output $\mathbf a_t$ is then given by the sum:
\begin{align}
\setstackgap{L}{12pt}\def\stacktype{L}\def\sz{\scriptstyle}
\stackunder{\mathbf a_t}{\sz (1\times d_m)} = 
\sum_{h=1}^{H}\stackunder{\mathbf o_t^{(h)}}{\sz (1\times d_m)}.
\end{align}
Notice that in the standard formulation of Vaswani et al. \cite{vaswani_attention_2017}, the context vectors are concatenated and then projected by a single $(Hd_v)\times d_m$ matrix $\tilde{\mathbf{W}}_O$ to produce 
\begin{align}
\tilde{\mathbf{a}} = [\mathbf{h}^{(1)}, \mathbf{h}^{(2)}, \dots, \mathbf{h}^{(H)}] \mathbf{W}_O.
\end{align}
That is equivalent to the parallel formulation above, in which each head has an individual $d_v \times d_{m}$ output projection matrix $\mathbf{W}_O^{(h)}$,
\begin{align}
\mathbf{a} = \sum_{h=1}^H \mathbf{o}^{(h)} = \sum_{h=1}^H (\mathbf{h}^{(h)} \cdot \mathbf{W}_O^{(h)}) = \tilde{\mathbf{a}}.    
\end{align}
The attention logits, i.e., the scores whose exponentials become probabilities, are given by:
\begin{align}
\mathbf l^{(h)} = \mathbf o^{(h)}\mathbf U,
\quad\text{so that}\quad \mathbf L_a = \sum_{h=1}^H\mathbf l^{(h)}.
\end{align}
In transformer block we also take into account the feedforward network (FFN) introducing nonlinear correction:
\begin{align}
\mathbf{z} = \max(0, (\mathbf X_0 + \mathbf a)  \mathbf W_1 + \mathbf b_1)\,\mathbf W_2  + \mathbf b_2,
\end{align}
generating the FFN logits:
\begin{align}
\mathbf{L}_{f} = \mathbf{z}\mathbf{U}.
\end{align}
As a result, the residual stream is adjusted as follows:
\begin{align}
\mathbf{X}_1 = \mathbf X_0 + \mathbf a + \mathbf z.
\end{align}
Finally, the probability distribution $\Pi$ of the next token is calculated by aggregating attention and FFN logits into:
\begin{align}
\mathbf L = \mathbf L_0 + \mathbf L_a + \mathbf L_f\quad\text{and normalizing into}\quad
\Pi = \text{softmax}(\mathbf L).
\end{align}
Finally, in a more generalized (still simplified a bit) matrix form, given a query $Q$ the transformer equation then becomes:
\begin{align}\begin{aligned}
\boldsymbol{\alpha}^{(h)}(\mathbf{X}) &= 
\mathrm{softmax}\left(\dfrac{1}{\sqrt{d_k}}
\mathbf{X}\mathbf{W}^{(h)}_Q\mathbf{W}^{(h)\top}_K\mathbf{X}^\top + \mathbf{M}_{\text{causal}}
\right),
\\
\mathbf{X}_{k+1} &=
\mathbf{X}_{k} + \sum_{h=1}^H \boldsymbol{\alpha}^{(h)}(\mathbf{X}_k)\mathbf{X}_k\mathbf{W}^{(h)}_V\mathbf{W}^{(h)}_O 
\\
&+ \sigma\left(\left[\mathbf{X}_{k} + \sum_{h=1}^H \boldsymbol{\alpha}^{(h)}(\mathbf{X}_k)\mathbf{X}_k\mathbf{W}^{(h)}_V\mathbf{W}^{(h)}_O
\right]\mathbf{W}^{(k)}_1\right)\mathbf{W}^{(k)}_2
\\[2ex]
&= \mathbf{X}_{k} + \mathrm{MHA}(\mathbf{X}_{k}) + \mathrm{MLP}(\mathbf{X}_{k})
\quad \text{for}\quad
\mathbf{X}_0 = \mathbf{E}(Q),
\\[2ex]
\boldsymbol{\Pi}_k &= \mathrm{softmax}(\mathbf{X}_k\mathbf{U}).
\end{aligned}\end{align}
That formulation shows clearly that the transformer inference is a discrete-time dynamical system.

\subsection{The Inference Process}

We shall keep things simple but realistic, recreating the essential steps of the inference process in transformer block.  

\subsubsection*{Step 1: Query and Key}

There are $H=2$ heads for which we define the query and key, or better yet, \textbf{demand and supply bids} in the \textbf{auction of ideas}. Query announces demand for information (\textit{what does this token need?}), key submits supply offers (\textit{what do the previous tokens offer!}). Then those bids are compared to find good matching (and clear the information market).

Let us assume the current position is $t = 3$, so we attend to \textit{brown}. Then, consider the following knowledge encoded in each head:
\begin{equation}
\mathbf{W}_Q^{(1)} = \begin{bmatrix} 0 & 0 \\ 0 & 0 \\ 0 & 1 \\ 0 & 0 \\ 0 & 0 \end{bmatrix} = 
\mathbf{W}_K^{(1)} \quad\mathrm{and}\quad
\mathbf{W}_V^{(1)} = \begin{bmatrix} 0 & 0 \\ 0 & 0 \\ 0 & 3 \\ 0 & 0 \\ 0 & 0\end{bmatrix},
\end{equation}
\begin{equation}
\mathbf{W}_Q^{(2)} = \begin{bmatrix} 0 & 0 \\ 0 & 0 \\ 1 & 0 \\ 0 & 0 \\ 0 & 0 \end{bmatrix} = \mathbf{W}_K^{(2)} \quad\mathrm{and}\quad
\mathbf{W}_V^{(2)} = \begin{bmatrix} 0 & 0 \\ 0 & 0 \\ 1.2 & 0 \\ 0 & 0 \\ 0 & 0\end{bmatrix}.
\end{equation}
Matrix $\mathbf W_Q^{(h)}$ amplifies the coordinates of tokens in $\mathbf X$ that contribute to \textbf{relevant context}. As a result, we get a query vector that represents the following question: 
\begin{center}
\textit{what information is token $\mathbf X_t$ looking for?}
\end{center}
In other words, the query vector indicates how strongly the current token demands information along the directions in $\mathbf Q_t^{(h)}$.

Then computing queries for $t=3$ with $\mathbf{X}_3$ yields:
\begin{align}
\mathbf{Q}_3^{(1)} = [0, 1] \quad\mathrm{and}\quad  \mathbf{Q}_3^{(2)} = [1, 0].
\end{align}
Head 1 demands information stored along direction $[0,1]$. Similarly, Head 2 demands information stored along direction $[1,0]$. That means the heads develop different (orthogonal) privately known and independent attention patterns, leading to beliefs about the next token that are as different as possible.

Next, we need to see what information can be provided by the input vector given the knowledge encoded in $\mathbf{W}_K^{(h)}$. Given that knowledge, we get from each head (agent) the key vectors that inform: 
\begin{center}
\textit{along which directions do the available tokens in $\mathbf X_{\le t}$ provide information relevant for token $\mathbf X_{t}$.}
\end{center}
Since we have assumed, for simplicity, that $\mathbf W_Q^{(h)} = \mathbf W_K^{(h)}$, the query and key vectors are perfectly aligned:
\begin{align}
\mathbf{K}_3^{(1)} = [0, 1]
\quad\mathrm{and}\quad
\mathbf{K}_3^{(2)} = [1, 0]
\quad\text{and}\quad
\mathbf{K}_{<3}^{(h)} = [0, 0].
\end{align}
Another interesting way to look at query and key is this. It is a memory lookup mechanism. Keys are address in memory, query selects addresses in memory which probably store relevant content. As we will see next, that content is provided by the value vectors.

\subsubsection*{Step 2: Attention Score and Weights}

Given the demand and supply bids providing directions along which information is required and stored, in the next step the attention mechanism executes the bid matching procedure. Namely, the attention scores are computed as scaled dot products of bids (query-key inner products divided by $\sqrt{d_k}$, not cosine similarity):
\begin{align}
\text{scores}_3^{(h)} = \frac{\mathbf{Q}_3^{(h)} (\mathbf{K}_{\leq 3}^{(h)})^\top}{\sqrt{d_k}}.
\end{align}
In our specific example, we have:
\begin{align}
\text{scores}_3^{(1)} = \frac{1}{\sqrt{2}}[0, 0, 1] = [0, 0, 0.707]
\quad\text{and}\quad
\text{scores}_3^{(2)} = \frac{1}{\sqrt{2}}[0, 0, 1] = [0, 0, 0.707].
\end{align}
Applying softmax to get attention weights:
\begin{equation}
\boldsymbol{\alpha}_3^{(h)} = \text{softmax}(\text{scores}_3^{(h)}),
\end{equation}
we get the following probability distributions (weights) over relevant context information (as column vector):
\begin{align}
\boldsymbol{\alpha}_3^{(1)} = 
\boldsymbol{\alpha}_3^{(2)} = [0.25, 0.25, 0.50]^\top.
\end{align}
Therefore, based on the encoded knowledge, heads attend strongly to position 3 (\textit{brown}).

\subsubsection*{Step 3: Context Vector Computation}

The context vector for each head is calculated as a linear combination ($\boldsymbol{\alpha}$ is a simplex) of value vectors:
\begin{equation}
\mathbf{h}_3^{(h)} = (\boldsymbol{\alpha}_3^{(h)})^\top \mathbf{V}_{\leq 3}^{(h)}.
\end{equation}
The value vectors,
\begin{align}
\mathbf{V}_{\leq t}^{(h)} = \mathbf{X}_{\leq t}^{(h)}\mathbf W^{(h)}_V,
\end{align}
prepare context information relevant for the next token prediction (performed in next steps). That critical content is extracted from the input vector based on the trained knowledge stored in $\mathbf W^{(h)}_V$.  Columns in  $\mathbf W^{(h)}_V$ detect presence of relevant information in  $\mathbf X_{\le t}$ and scale it appropriately. 

We have:
\begin{align}
\begin{aligned}
\mathbf{V}_{\leq 3}^{(1)} &=
\begin{bmatrix}
1 & 0 & 0 & 0 & 0 \\
0 & 1 & 0 & 0 & 0 \\
0 & 0 & 1 & 0 & 0
\end{bmatrix}
\begin{bmatrix} 0 & 0 \\ 0 & 0 \\ 0 & 3 \\ 0 & 0 \\ 0 & 0 \end{bmatrix}
=
\begin{bmatrix} 0 & 0 \\ 0 & 0 \\ 0 & 3 \end{bmatrix},
\\
\mathbf{V}_{\leq 3}^{(2)} &=
\begin{bmatrix}
1 & 0 & 0 & 0 & 0 \\
0 & 1 & 0 & 0 & 0 \\
0 & 0 & 1 & 0 & 0
\end{bmatrix}
\begin{bmatrix} 0 & 0 \\ 0 & 0 \\ 1.2 & 0 \\ 0 & 0 \\ 0 & 0 \end{bmatrix}
=
\begin{bmatrix} 0 & 0 \\ 0 & 0 \\ 1.2 & 0 \end{bmatrix}.
\end{aligned}
\end{align}
That means both heads believe that the presence of \textit{brown} in position 3  is relevant (row 3 in $\mathbf W^{(h)}_V$ has nonzero elements). Token \textit{brown} statistically predicts important pattern extracted from training data. That information is encoded by both heads as a vectors in value space, 
\begin{align}
\mathbf{V}_3^{(1)} = [0, 3]\quad\text{and}\quad \mathbf{V}_3^{(2)} = [1.2, 0].
\end{align}

Given the demand-supply matching encoded in $\boldsymbol{\alpha}^{(h)}_3$, that yields head context vectors:
\begin{align}
\mathbf{h}_3^{(1)} = [0, 1.5]\quad\text{and}\quad \mathbf{h}_3^{(2)} = [0.6, 0].
\end{align}

\subsubsection*{Step 4: Output Projection and Feedforward Exploration}

Each head then projects its context vector to residual space based on the privately held information. That information allows for routing knowledge encoded in $\mathbf{W}_O^{(h)}$ and translates (or transmits) valuable context  into embedding vectors. Namely, each head calculates:
\begin{equation}
\mathbf{o}_3^{(h)} = \mathbf{h}_3^{(h)} \mathbf{W}_O^{(h)}.
\end{equation}
With trained matrices:
\begin{equation}
\mathbf{W}_O^{(1)} = \begin{bmatrix} 0 & 0 & 0 & 0 & 0\\ 0 & 0 & 0 & 1 & 0 \end{bmatrix}\quad\text{and}\quad
\mathbf{W}_O^{(2)} = \begin{bmatrix} 0 & 0 & 0 & 0 & 1 \\ 0 & 0 & 0 & 0 & 0\end{bmatrix},
\end{equation}
the attention mechanism generates ranking of evidence emphasizing relevance of the vocabulary tokens:
\begin{align}
\mathbf{o}_3^{(1)} = [0, 0, 0, 1.5, 0]
\quad\text{and}\quad
\mathbf{o}_3^{(2)} = [0, 0, 0, 0, 0.6].
\end{align}
So, in the next token prediction auction of ideas, \textbf{Head 1} votes for \textit{fox} and \textbf{Head 2} votes for \textit{dog}. 

The votes are added into residual dimensions to adjust the meaning of input embedding. For that, the outputs (votes) are aggregated into:
\begin{align}
\mathbf{a}_3 = \sum_{h=1}^{H} \mathbf{o}_3^{(h)} = [0, 0, 0, 1.5, 0.6]
\quad\text{and}\quad
\mathbf{a}_{<3} =  [0, 0, 0, 0, 0],
\end{align}
and added to input vector:
\begin{align}
\mathbf{X}_1 = \mathbf X_0 + \mathbf A.
\end{align}
That completes the operations of the attention block.

The next step is to add contribution of \textbf{FFN} feedforward block (or another agent participating in the auction). We have:
\begin{align}
\mathbf{X}_2 = \mathbf X_1 + \max(0, \mathbf X_1  \mathbf W_1 + \mathbf b_1)\,\mathbf W_2  + \mathbf b_2.
\end{align}
That FFN operation explores context locally and introduces additional innovations in meaning, shaping the next token prediction result. It can amplify or counteract the votes injected by attention. Notice that adjustments are applied at every position separately, so that ideas provided by attention bids are interpreted and rewritten by the FFN agent (see Section \ref{sec:Discussion and Speculations}).

In this example, we select the following parameters of the FFN block:
\begin{align}
W_1 = \begin{bmatrix}
\mathbf{I}_{d_m}\ \mathbf{0}_{d_m\times 1}
\end{bmatrix}   
\quad\text{and}\quad
W_2 = \begin{bmatrix}
 0 & 0 & 0 & 0 & 0 \\
 0 & 0 & 0 & 0 & 0 \\
 0 & 0 & 0.13 & 0 & 0 \\
 0 & 0 & 0 & -0.13 & -0.05 \\
 -0.05 & 0.35 & 0 & 0 & 0 \\
 0 & 0 & 0 & 0 & 0
\end{bmatrix}
\quad\text{and}\quad
\mathbf{b}_i = 0. 
\end{align}

\subsubsection{Step 5: Logits and Prediction}

Finally, the unembedding matrix $\mathbf U$ maps the multi-headed attention and FFN outputs from model space to vocabulary in order to calculate probability distribution for the next token prediction.

Each head's contribution to the final logits is:
\begin{align}
\mathbf{l}^{(1)} = [0, 0, 0, 0, 1.5, 0]\quad\text{and}\quad
\mathbf{l}^{(2)} = [0, 0, 0, 0, 0, 0.6].
\end{align}
The total attention logits are the sum of individual head logits:
\begin{align}
\mathbf{L}_a = \sum_{h=1}^{H} \mathbf{l}^{(h)} = [0, 0, 0, 0, 1.5, 0.6].
\end{align}
That sum is combined with the input logit $\mathbf{L}_0$ and the FFN logits $\mathbf L_f$, which introduce some creative noise. As a result, we obtain:
\begin{align}
\mathbf{L} = \mathbf{L}_0 + \mathbf{L}_a+\mathbf{L}_f = [0, -0.03, 0.21, 1.13, 1.305, 0.525].
\end{align}
That vector translates into the final probability distribution:
\begin{align}
\Pi = \text{softmax}(\mathbf{L}) \approx
[0.086,0.083,0.106,0.265,\textbf{0.316},0.145].
\end{align}
The winner of the auction of ideas is selected based on $\Pi$. The inference process generates and collects competing beliefs about relevant context and next token prediction, all based on the knowledge stored in the attention heads and FFN layer. And so, we see that when that knowledge is used, the most probable next token (word) reachable in the vocabulary is 
\begin{align}
\text{argmax}\ \Pi = \textit{fox}.
\end{align}

\subsection{Proof III: The Impossibility Theorem for Transformers}

With the transformer inference process in hand, we now check whether one explicit account for the additive-logit model above can balance.

Two modeling choices frame the statement. First, the declared channels must sum exactly to the final logits. We do not assume a nonlinear final layer normalization can always be folded into additive channels, so the theorem applies directly to the one-block model written here, and to a larger architecture only once every residual, bias, normalization, and unembedding contribution has been given an exact additive decomposition. Second, the log-loss accounting rule is a stipulated diagnostic, not a transfer rule that logit summation forces on us.

\begin{theorem}[Impossibility for Transformers]
\label{thm:lse-impossibility}
Let $|\mathcal V|\ge2$ and $m\ge2$, and let finite channel logits $l^{(c)}\in\mathbb R^{|\mathcal V|}$ satisfy the exhaustive decomposition $L=\sum_{c=1}^{m}l^{(c)}$, formed from the token-embedding channel, the $H$ attention heads, and the FFN channel ($m=H+2$). Define the channel profiles $\pi^{(c)}=\mathrm{softmax}(l^{(c)})$ and the aggregate $\Pi=\mathrm{softmax}(L)$. For any token $y^*\in\mathcal V$, assign the stipulated account entries $p_c=-\log\pi^{(c)}_{y^*}$ to the channels and $p_0=\log\Pi_{y^*}$ to the aggregator. Then the account has a strictly positive, outcome-independent balance defect $\sum_{c=0}^{m}p_c=\Gamma>0$, and therefore cannot balance.
\end{theorem}

\begin{proof}
The channel profiles are normalization devices for the declared logits. They need not be separately emitted probabilistic forecasts.

For every declared channel $c=1,\dots,m$, the channel distribution is:
\begin{align}
\pi^{(c)}_y = \frac{\exp\bigl(l^{(c)}_y\bigr)}{Z_c}
\ \text{with}\
Z_c = \sum_{\bar{y}} \exp\bigl(l^{(c)}_{\bar{y}}\bigr).
\end{align}
Similarly, the aggregate distribution from summed logits is calculated as:
\begin{align}
\Pi_y = \frac{\exp(L_y)}{Z}
\ \text{with}\
Z = \sum_{\bar{y}} \exp(L_{\bar{y}})
\ \text{and}\
L_y = \sum_{c=1}^m l^{(c)}_y.
\end{align}
For any observed (realized) token $y^*$, the channel losses and aggregate loss are:
\begin{align}
p_c = -\log \pi^{(c)}_{y^*} = -l^{(c)}_{y^*} + \log Z_c
\quad\text{and}\quad
p_{0} = -(-\log \Pi_{y^*}) = L_{y^*} - \log Z.
\end{align}
All declared channels remain in the account, including token-constant ones. Since the logits are finite and $|\mathcal V|\ge2$, every $p_c=-\log\pi^{(c)}_{y^*}$ is strictly positive.

Since $L_{y^*} = \sum_c l^{(c)}_{y^*}$, cancellation of the realized-token logits gives
\begin{align}
\sum_{c=0}^m p_c =  \sum_{c=1}^m \log Z_c - \log Z = \log\frac{\prod_{c=1}^m Z_c}{Z} = \Gamma > 0.
\end{align}
Indeed, $\prod_c Z_c$ expands as a sum over all tuples $(y_1,\dots,y_m)\in\mathcal V^m$, whereas $Z$ collects only the diagonal tuples $y_1=\cdots=y_m$. All terms are positive, and because $m\ge2$ and $|\mathcal V|\ge2$ at least one off-diagonal tuple exists, so $\prod_c Z_c>Z$. The stipulated account therefore cannot balance. Notice that no channel-disagreement premise was needed: the defect is present for every finite channelization, and disagreement enters only through the term $D$ below.
\end{proof}

The gap admits a gauge-invariant decomposition that separates its two sources. Writing $R_{1/m}(\Pi)$ for the R\'enyi entropy of order $1/m$ of the aggregate $\Pi$,
\begin{align}
\Gamma = (m-1)\,R_{1/m}(\Pi) + D,
\end{align}
where the floor $(m-1)\,R_{1/m}(\Pi)$ is gauge-invariant (unchanged by adding a common constant to every channel logit) and the excess-confidence term
\begin{align}
D = \sum_{c=1}^m \mathrm{LSE}\bigl(l^{(c)}\bigr) - m\,\mathrm{LSE}\!\Bigl(\tfrac1m\textstyle\sum_{c=1}^m l^{(c)}\Bigr) \ge 0,
\qquad \mathrm{LSE}(v)=\log\textstyle\sum_y e^{v_y},
\end{align}
is nonnegative by convexity of the log-sum-exp (Jensen's inequality), and vanishes exactly when the channel logits differ only by token-constant shifts, that is, when their normalized profiles coincide. So $D$ measures how much the channels disagree within the declared channelization, while the floor stays strictly positive even when the profiles agree perfectly. Keeping the floor distinct from $D$ is what removed the earlier gauge-dependence.

Two scope caveats keep this exact. First, the floor $(m-1)R_{1/m}(\Pi)$ is a functional of the aggregate $\Pi$ alone, so it is invariant under the reparameterizations that fix the model's output at a fixed channelization: adding a common constant to a channel's logits, the value and output rescaling $W_V\mapsto aW_V$, $W_O\mapsto a^{-1}W_O$, and sum-preserving redistribution of the residual stream across channels. Second, the full gap $\Gamma$ is defined \emph{relative to the declared channelization}. It is not invariant under a change of channelization itself. Merging two channels into their logit-sum preserves the aggregate logits $L$, and hence the model's output, byte for byte, yet changes the channel count, the R\'enyi order, and the excess $D$, hence $\Gamma$. In fact any functional invariant under such channel merges must factor through the aggregate logits $L$, and adding token-constant gauge invariance makes it factor through $\Pi=\mathrm{softmax}(L)$, so any quantity that separates two channelizations with the same $\Pi$ cannot be merge-invariant. By an analogous implementation-invariance criterion for attributions \citep{sundararajan2017axiomatic}, only functionals of $\Pi$ are decomposition-independent, and $\Gamma$ is not one of them. We accordingly scope every $\Gamma$ claim to the one-block channelization of this section.

Theorem \ref{thm:lse-impossibility} pins down a measurable, channelization-relative gap in the additive-logit model. Under the stipulated log-loss accounting rule and the exhaustive decomposition, the defect $\Gamma$ is positive for every finite channelization, whatever the channels do, and their disagreement enters only through the term $D$. We do not claim this accounting rule is forced by, or fully identifiable in, every deployed transformer. Extending it there would need both the contribution rule to hold in the architecture at hand and an exact additive decomposition of its residual, bias, and normalization terms. Theorems \ref{thm:proper-scoring-impossibility} and \ref{thm:lse-impossibility} are therefore statements about specified aggregation rules and channel decompositions, not an unconditional law for all transformers.

The aggregate does not conjure information from nothing. It reorganizes what the channels already carry to be more useful for predicting $y^*$, making information more accessible rather than more real. What $\Gamma$ measures is a gap in the model's own bookkeeping, how much extra confidence the aggregation shows over its parts, and nothing more. It is not a measure of how well an answer is supported, and it is not an amount of real information. Redraw the channel boundaries and $\Gamma$ changes while the model's output does not move at all, and its independence of the realized outcome $y^*$ is exactly the cancellation in the proof above.

\subsection{The Aggregation Mechanics}

Let us again take a closer look at the example of transformer micro-architecture to see how the impossibility result in Theorem \ref{thm:lse-impossibility} manifests itself.

Recall that the two-headed transformer with $\mathcal{V} = \{\langle\text{PAD}\rangle, \text{The}, \text{quick}, \text{brown}, \text{fox}, \text{dog}\}$ infers based on the input sequence
$$
\mathrm{\textit{The quick brown}}.
$$
As we have seen, the model parameters encode specialized knowledge in value matrix:

\begin{itemize}
\item \textbf{Head 1}: association \textit{brown} $\rightarrow$ \textit{fox} with strength $3.0$,
\item \textbf{Head 2}: association \textit{brown} $\rightarrow$ \textit{dog} with strength $1.2$.
\end{itemize}

At position $t=3$ (token \textit{brown}), the attention mechanism generates the following head-specific logit contributions:
\begin{align}
\mathbf{l}^{(1)} = [0, 0, 0, 0, 1.5, 0]\quad\text{and}\quad
\mathbf{l}^{(2)} = [0, 0, 0, 0, 0, 0.6],
\end{align}
where the fifth and sixth positions correspond to \textit{fox} and \textit{dog} respectively.

The conservation account of Theorem~\ref{thm:lse-impossibility} runs over all $m=4$ contribution channels: the token-embedding channel $\mathbf{L}_0=[0,0,0,1,0,0]$, the two attention heads, and the FFN channel $\mathbf{L}_f=[0,-0.03,0.21,0.13,-0.195,-0.075]$. Their exhaustive sum is the full aggregate of Step~5,
\begin{align}
\mathbf{L} = \mathbf{L}_0+\mathbf{l}^{(1)}+\mathbf{l}^{(2)}+\mathbf{L}_f =
[0, -0.03, 0.21, 1.13, 1.305, 0.525],
\end{align}
yielding the probability distribution:
\begin{align}
\Pi = \text{softmax}(\mathbf{L}) =
[0.086,0.083,0.106,0.265,\textbf{0.316},0.145].
\end{align}
Therefore, the model assigns \textit{fox} $31.6\%$ confidence and \textit{dog} $14.5\%$. The four-channel Jensen gap quantifies the stipulated account's balance defect:
\begin{align}
\Gamma = \sum_{c=1}^{4} \log Z_c - \log Z = 5.563 > 0,
\end{align}
the excess of the four channels' summed costs over the aggregate's, where $Z_c = \sum_y \exp(l^{(c)}_y)$ ranges over the channels $c\in\{\mathbf{L}_0,\mathbf{l}^{(1)},\mathbf{l}^{(2)},\mathbf{L}_f\}$ and $Z = \sum_y \exp(L_y)$ (with the decomposition $\Gamma=(m-1)R_{1/m}(\Pi)+D=5.268+0.295$ at $m=4$).

In this toy sequence the top prediction is actually correct, yet the stipulated four-channel account still shows a positive defect. Nothing in this calculation decides whether the response is supported by evidence or factually true.

To see the point, look at the four channels' own probability profiles:
\begin{align*}
\pi^{(0)} &= \text{softmax}(\mathbf{L}_0)
= [0.130, 0.130, 0.130, \textbf{0.352}, 0.130, 0.130]
&&(\text{embedding, favors \textit{brown}}),
\\
\pi^{(1)} &= \text{softmax}(\mathbf{l}^{(1)})
= [0.105, 0.105, 0.105, 0.105, \textbf{0.473}, 0.105]
&&(\text{head 1, favors \textit{fox}}),
\\
\pi^{(2)} &= \text{softmax}(\mathbf{l}^{(2)})
= [0.147, 0.147, 0.147, 0.147, 0.147, \mathbf{0.267}]
&&(\text{head 2, favors \textit{dog}}),
\\
\pi^{(f)} &= \text{softmax}(\mathbf{L}_f)
= [0.164, 0.159, 0.202, 0.187, 0.135, 0.152]
&&(\text{FFN, diffuse}).
\end{align*}
Is there any combination of these channels that justifies the aggregate assigning the next token \textit{fox} probability $\Pi_{fox} = 31.6\%$? No single channel is the culprit: across the four, \textit{fox} ranges from the embedding's $0.130$ to head~1's $0.473$, so $0.316$ sits inside their range and a mere reweighting could reach it.

The essential point is different: the aggregate is not any combination of the channels. Writing $\mathbf{P}=[(\pi^{(0)})^\top,(\pi^{(1)})^\top,(\pi^{(2)})^\top,(\pi^{(f)})^\top]$ for the four channel distributions, the least-squares fit $\hat{\boldsymbol{\beta}} = \mathbf{P}^{+}\Pi$ leaves a nonzero residual (see e.g. \cite{karpowicz_theory_2021,karpowicz2024pseudoinverseacrarc}),
\begin{align}
\mathbf{d} =  (\mathbf{I}- \mathbf{P}\mathbf{P}^+) \Pi \neq \mathbf{0}.
\end{align}
This fit is a \emph{linear}, not necessarily convex, combination, so $\hat{\boldsymbol\beta}$ need not be a probability vector. Hence $\Pi$ does not lie in the linear span of the channel profiles, and certainly is no convex mixture of them. It is instead their normalized \emph{product}, the Product-of-Experts structure of additive aggregation (Theorem~\ref{thm:poe} states it for the attention subsystem, and the same algebra applies channel-wise). The nonzero residual $\mathbf{d}$ and the scalar defect $\Gamma$ are two distinct, channelization-relative diagnostics of that fact. Neither should be read as factual unsupportedness, since $\mathbf{d}$ is a vector projection residual and $\Gamma$ a scalar log-partition gap.

Restricting attention to the two knowledge-bearing heads alone, setting the embedding prior and the FFN channel aside, the attention subsystem's own aggregate $\mathbf{L}_a=\mathbf{l}^{(1)}+\mathbf{l}^{(2)}=[0,0,0,0,1.5,0.6]$ already carries a positive defect, with $\Gamma_a=\sum_{h=1}^{2}\log Z_h-\log Z_a=1.837$ and $\Pi_a=\text{softmax}(\mathbf{L}_a)=[0.097,0.097,0.097,0.097,0.435,0.177]$ that is itself no mixture of its two heads. The defect does not rest on the embedding or FFN channels, and it is a diagnostic of this reduced two-channel view, not of the full block.

A second description of additive-logit aggregation comes from the product-of-experts identity, the structure introduced by Hinton \citep{hinton2002training}.

\begin{theorem}[Attention Aggregation is Product-of-Experts]
\label{thm:poe}
Let $\mathbf{l}^{(h)}$ be head logits and $\pi^{(h)}=\mathrm{softmax}(\mathbf{l}^{(h)})$. With additive logit aggregation, the attention-only distribution $\Pi_a$ is a Product-of-Experts (PoE)—the normalized product of the head distributions:
\begin{align}
(\Pi_a)_i = \frac{\prod_{h=1}^H \pi^{(h)}_i}{\sum_{j=1}^{|\mathcal{V}|}\prod_{h=1}^H \pi^{(h)}_j}.
\end{align}
\end{theorem}

\begin{proof}
By definition of softmax, for each head $h$:
\begin{align*}
\pi^{(h)}_i = \frac{\exp(l^{(h)}_i)}{\sum_{k=1}^{|\mathcal{V}|} \exp(l^{(h)}_k)} = \frac{\exp(l^{(h)}_i)}{Z_h}
\end{align*}
where $Z_h = \sum_{k=1}^{|\mathcal{V}|} \exp(l^{(h)}_k)$ is the partition function for head $h$. Rearranging, we have:
\begin{align*}
\exp(l^{(h)}_i) = Z_h \cdot \pi^{(h)}_i
\end{align*}
The aggregate logits corresponding to $i$-th token are given by $(\mathbf{L}_a)_i = \sum_{h=1}^H l^{(h)}_i$. Therefore:
\begin{align*}
\exp\left((\mathbf{L}_a)_i\right)  &= \exp\left(\sum_{h=1}^H l^{(h)}_i\right) 
= \prod_{h=1}^H \exp(l^{(h)}_i)\\
&= \prod_{h=1}^H \left(Z_h \cdot \pi^{(h)}_i\right)
= \left(\prod_{h=1}^H Z_h\right) \cdot \left(\prod_{h=1}^H \pi^{(h)}_i\right).
\end{align*}
As a result, the attention-only distribution is:
\begin{align}
(\Pi_a)_i &= \frac{\exp\left((\mathbf{L}_a)_i\right)}{\sum_{j=1}^{|\mathcal{V}|} \exp\left((\mathbf{L}_a)_j\right)}\\
&= \frac{\left(\prod_{h=1}^H Z_h\right) \cdot \left(\prod_{h=1}^H \pi^{(h)}_i\right)}{\left(\prod_{h=1}^H Z_h\right) \cdot \sum_{j=1}^{|\mathcal{V}|} \left(\prod_{h=1}^H \pi^{(h)}_j\right)}\\
&= \frac{\prod_{h=1}^H \pi^{(h)}_i}{\sum_{j=1}^{|\mathcal{V}|} \prod_{h=1}^H \pi^{(h)}_j},
\end{align}
which is the Product-of-Experts (PoE) formula.
\end{proof}

In other words, the final probability of each token is proportional to the product of its probabilities across the heads. This multiplicative structure tends to reward tokens supported across heads and to penalize disagreement. But no head holds unconditional veto power: after normalization, a token with a small product can still come out on top when its competitors' products are smaller.

The additive-logit aggregation is a logarithmic opinion pool of the channel distributions, $\Pi\propto\prod_c\pi_c$. Among pooling rules, external Bayesianity, the invariance under a shared Bayesian update, singles out the log-linear \emph{form}, as opposed to linear (arithmetic) pooling \citep{genest1984characterization,genest1986characterization,genest1986combining}. This licenses the form of aggregation, not the transformer's particular unit weights: the externally-Bayesian pool uses weights summing to one, whereas summing all channel logits assigns unit weights that sum to the channel count. If the same likelihood factor occurs in every channel, unit-weight pooling counts that factor $m$ times. The log-linear \emph{family} is therefore axiomatically grounded, while the specific unit-weight rule is one tempered member of it.

This makes Product-of-Experts pooling behave differently from a smoothing Mixture-of-Experts average. PoE ranks tokens by products rather than weighted sums, so it can sharpen a preference the heads share, and it can just as easily flatten or redirect the distribution when the heads pull against each other.

On top of the head bids, the full block adds the input residual and the FFN output to the aggregate logits, which can amplify or dampen the product-of-experts result. The aggregate need not end up more confident than every constituent profile.

Under the stipulated account, Theorem \ref{thm:lse-impossibility} gives a positive balance defect. It does not, on its own, establish strategic untruthfulness or an unjustified prediction.

There is a real difference between the mathematical over- or under-confidence of a distribution and factual correctness. A correct prediction may be a lucky product of the aggregation itself rather than an evidence-based derivation by the components.

Is that manufactured confidence, then, an ingredient of creativity or imagination? That is an empirical and philosophical question, and not something the accounting identity settles.

\subsection{The Semantic Information Measure and Emergence Operator in Transformer}

We now instantiate the abstract semantic information framework within the concrete transformer architecture. This maps the theoretical concepts of reasoning and information flow in Section \ref{sec:Knowledge and Semantic Information} to measurable quantities in the micro-transformer example.

Recall that we model the abstract knowledge space as a Polish metric space $(\mathcal K,d_{\mathcal K})$ (complete, separable). In transformer-based LLM that can be viewed as the embedding space, i.e,, $\mathbb R^d$ with Euclidean or angular distance. Individual facts, concepts, or patterns are represented not by single points necessarily, but by analytic subsets of $\mathcal K$, i.e., elements of $\mathcal A(\mathcal K)$. This is the natural domain for set‑transformations used by our emergence operator.

To be specific, the knowledge representation in LLM is a tuple $(\mathcal K, d_{\mathcal K}, \mathcal W, \mathcal P, \Phi, \Psi)$ where:
\begin{itemize}
\item $(\mathcal K,d_{\mathcal K})$ is a Polish space of knowledge elements,
\item $\mathcal W$ (parameters) and $\mathcal P$ (activations) are Polish spaces,
\item $\Phi:\mathcal W \to \mathcal A(\mathcal K)$ maps weights $W$ to an analytic encoded‑knowledge set $\Phi(W)=\mathcal K_M$,
\item $\Psi:\mathcal P \to \mathcal A(\mathcal K)$ maps an activation $a\in\mathcal{P}$ to an analytic accessible‑knowledge set $\Psi(a)$. 
\end{itemize}

For a parameterization $W\in\mathcal W$, $\Phi(W)=\mathcal K_M$ is the analytic set of knowledge encoded in the weights. For an activation state $a\in\mathcal P$ produced under $W$, the accessible set $\Psi(a)\subseteq \Phi(W)$ captures what the forward pass makes computationally available at that state. Thus, representation separates what is encoded from what is currently accessible.

In transformer LLMs, $\mathcal W$ represents all trained weights (attention $W_Q,W_K,W_V,W_O$ and MLP matrices), $\mathcal P$ is the product space of residual‑stream activations across layers and positions. A computational budget $C$ bounds the number of transformer layers of blocks. Parameters determine $\mathcal K_M=\Phi(W)$, whereas a forward pass produces $a\in\mathcal P$ and exposes $\Psi(a)\subseteq\mathcal K_M$. Inference maps $f\in\mathcal F_C$ act on current analytic knowledge sets, and $\mathcal E_C$ aggregates their (relevant) images. Multi‑step reasoning thus iterates $\mathcal E_C$ exploring regions of $\mathcal K$ that are task‑relevant under $Q$ and budget $C$. 

The transformer operations are compositions of continuous functions and standard arithmetic operations, so they are Borel measurable. A block map acts on points, while the operators of $\mathcal{O}(\mathcal{K})$ act on sets, so the bridge is the induced image map: a continuous block map applied elementwise to an analytic activation set yields an analytic image, and these induced set maps are the members of $\mathcal{O}(\mathcal{K})$ that the architecture realizes. Under the stated measurability and bounded-domain assumptions, the properties of the emergence operator $\mathcal{E}_C$ then apply.

\subsubsection*{The Semantic Information Measure}

To define the semantic measure that meets all the axioms we have introduced, we utilize a discount factor and the supremum over the input set.

\begin{definition}[Transformer Semantic Information Measure]
\label{def:transformer-semantic-measure}
The Semantic Information Measure $\mu_C(A|Q)$ of a knowledge subset $A$ within a query context $Q$ is the  supremum of the discounted energy over the input set $A$:
\begin{align}
\mu_C(A|Q) = \sup\left(\{0\}\cup\left\{ \sum_{l=1}^{C} \gamma^{l-1} E_l(\mathbf{X}|Q)\colon \mathbf{X} \in A \right\}\right),\ \text{ for } 0<\gamma<1,
\end{align}
where $E_l(\mathbf{X}|Q)$ represents the total energy of the context vectors at $l$-th layer:
\begin{align}
E_l(\mathbf{X}|Q) = \sum_{h=1}^{H} \sum_{t=1}^{T} \|\mathbf{h}_t^{(h,l)}(\mathbf{X}|Q)\|_2^2, 
\end{align}
and $\mathbf{h}_t^{(h,l)} (\mathbf{X}|Q)=\boldsymbol{\alpha}_t^{(h,l)}(Q)\mathbf{V}_{\leq t}^{(h,l)}(Q)$ is the context-dependent vector. 
\end{definition}

We can verify the axioms of the semantic information measure, as introduced in Definition \ref{def:semantic-information-measure}. The null-set axiom holds by the convention $\mu_C(\emptyset|Q)=0$ (the supremum over $\{0\}\cup\emptyset$). Non-negativity is guaranteed by squared norms and positive discount factor. Monotonicity follows from the monotonicity of supremum. Subadditivity holds since $\mu_C(A \cup B|Q) = \max\{\mu_C(A|Q), \mu_C(B|Q)\} \leq \mu_C(A|Q) + \mu_C(B|Q)$ for non-negative values (so the conditional-subadditivity axiom holds unconditionally here). Contextual positivity ($\mu_C(A|Q)>0$ iff the relevant content of $A$ is non-null) holds under the modeling hypothesis that the energy $E_l(\mathbf{X}|Q)$ is strictly positive exactly on context-relevant vectors and zero off $\mathcal{R}_C(Q)$. Insight monotonicity follows as additional layers add non-negative terms. Boundedness is ensured by the geometric series with $\gamma < 1$ under bounded weights \emph{and} bounded activations, which together give a uniform bound. Then the energy of context is bounded,  $E_l(\mathbf{X}|Q) \leq M$ for some constant $M$ and:
\begin{align}
\mu_{\infty}(A|Q) \leq \sum_{l=1}^{\infty} \gamma^{l-1} M = \frac{M}{1-\gamma} < \infty.
\end{align}
The attention mechanism calculating distribution $\boldsymbol{\alpha}$ and the value projections $\mathbf{V}$ is the realization of contextual relevance $\mathcal{R}_C(Q)$, modulating the information flow based on the specific context $Q$.  The context dictates the learned weights to shape the attention landscape, controlling the feature extraction process that makes hidden information accessible and orders it into a knowledge. One caveat keeps this realization honest. It is not implementation invariant: the function-preserving rescaling $W_V\mapsto aW_V$, $W_O\mapsto a^{-1}W_O$ leaves every head output and every model prediction unchanged, yet rescales the context vectors and their squared energy by $a^2$. The numerical values of this measure are therefore meaningful relative to the declared parameterization, and comparisons across implementations require fixing that gauge first. 

Applying these definitions to our example, we must consider context vectors:
\begin{align}
\mathbf{h}_3^{(1)} = [0, 1.5] \quad\text{and}\quad
\mathbf{h}_3^{(2)} = [0.6, 0].
\end{align}
For $t\le 2$ the value vectors $\mathbf{V}_{t\le 2}$ are zero, thus $\mathbf{h}_{t\le 2}^{(h)} = \mathbf{0}$. Then, we have:
\begin{align}
E_1(\mathbf{X}_0) = \sum_{t=1}^{3} \left(\|\mathbf{h}_t^{(1)}\|_2^2 + \|\mathbf{h}_t^{(2)}\|_2^2\right)
= 0 + 0 + (\|[0, 1.5]\|_2^2 + \|[0.6, 0]\|_2^2) = 2.61.
\end{align}
With a single layer $C=1$ and  $\gamma^0 = 1$, the semantic measure of the input matrix $\mathbf{X}_0$ is:
\begin{align}
\mu_1(\{\mathbf{X}_0\}|Q) = \gamma^0 \cdot 2.61 = 2.61
\end{align}
That is the information flow producing the organized and accessible knowledge.

\subsubsection*{The Emergence Operator}

The emergence operator acts on sets of knowledge states $A \subseteq \mathcal{K}$ and it must be inflationary, i.e., $A \subseteq \mathcal{E}_C(A|Q)$. In the transformer context, this corresponds to the set of all activation states reachable during the forward pass. 

\begin{definition}[Transformer Emergence Operator]
\label{def:transformer-emergence-operator}
Let $F_l: \mathcal{K} \to \mathcal{K}$ denote the non-linear transformation performed by the $l$-th transformer block:
\begin{align}
F_l(\mathbf{X}) = \mathbf{X} + \mathbf{A}^{(l)}(\mathbf{X}) + \mathbf{Z}^{(l)}(\mathbf{X}).
\end{align}
The Emergence Operator $\mathcal{E}_C(A|Q)$ produces all states generated from the initial set $A$ up to depth $C$:
\begin{align}
\mathcal{E}_C(A|Q) = A \cup \bigcup_{c=1}^C \{ F_c \circ \cdots \circ F_1(\mathbf{X}) : \mathbf{X} \in A \}.
\end{align}
\end{definition}

By construction, the operator is extensive, i.e., $A \subseteq \mathcal{E}_C(A|Q)$ since the union includes $A$. It is also monotonic with respect to inclusion. If $A \subseteq B$, we have $\{F(\mathbf{X}) : \mathbf{X} \in A\} \subseteq \{F(\mathbf{X}) : \mathbf{X} \in B\}$, hence $\mathcal{E}_C(A|Q) \subseteq \mathcal{E}_C(B|Q)$. 

In the micro-transformer processing "The quick brown", we start with $A = \{\mathbf{X}_0\}$ where:
\begin{align}
\mathbf{X}_0 = \begin{bmatrix}
1 & 0 & 0 & 0 & 0 \\
0 & 1 & 0 & 0 & 0 \\
0 & 0 & 1 & 0 & 0
\end{bmatrix}
\end{align}
The attention mechanism produces:
\begin{align}
\mathbf{A}^{(1)} = \begin{bmatrix}
0 & 0 & 0 & 0 & 0 \\
0 & 0 & 0 & 0 & 0 \\
0 & 0 & 0 & 1.5 & 0.6
\end{bmatrix}
\end{align}
The FFN with input $\mathbf{X}_0 + \mathbf{A}^{(1)}$ yields (entries rounded to two decimals here and in $\mathbf{X}_1$ below, the exact values being those of the worked example above):
\begin{align}
\mathbf{Z}^{(1)} = \begin{bmatrix}
0 & 0 & 0 & 0 & 0 \\
0 & 0 & 0 & 0 & 0 \\
-0.03 & 0.21 & 0.13 & -0.19 & -0.08
\end{bmatrix}.
\end{align}
With:
\begin{align}
\mathbf{X}_1 = F_1(\mathbf{X}_0) = \mathbf{X}_0+\mathbf{A}^{(1)}+\mathbf{Z}^{(1)} =\begin{bmatrix}
1 & 0 & 0 & 0 & 0 \\
0 & 1 & 0 & 0 & 0 \\
-0.03 & 0.21 & 1.13 & 1.31 & 0.52
\end{bmatrix},
\end{align}
the new knowledge made accessible becomes:
\begin{align}
\mathcal{E}_1(A|Q) = \{\mathbf{X}_0, \mathbf{X}_1\}.
\end{align}
The emergence operator $\mathcal{E}_C$ expands the accessible state space and makes new states accessible. Then the excess-confidence signature arises from how these states are transformed into probabilities through the softmax nonlinearity. The semantic measure $\mu_C$ tracks the information flow.

\section{Safety, Alignment and Bounded Creativity}
\label{sec:Safety and Alignment}

This section works in the Polish-space knowledge model and defines support relative to an authorized boundary. We introduce the safety envelope through bounded creativity, and the conservation--reasoning dichotomy shows why strict information conservation is the wrong tool for it: enforcing conservation would eliminate meaningful reasoning altogether. We then give a sufficient containment result and an effective support certificate under explicit decidability assumptions. Throughout, these results certify support against the chosen evidence, not factual truth.

As a reminder, the knowledge space $(\mathcal K,d_{\mathcal K})$ is Polish, all reasoning sets live in the class of analytic sets $\mathcal{A}(\mathcal K)$, the semantic information measure $\mu_C$ is defined on the universally measurable $\sigma$–algebra $\mathcal U(\mathcal K)$, so $\mu_C(A| Q)$ is well defined for all analytic subsets of knowledge. The relevance region $\mathcal{R}_C(Q)$ is analytic. Also, the emergence operator $\mathcal E_C(\cdot| Q)$ is inflationary, monotone, and $\omega$-Scott–continuous on $\mathcal{A}(\mathcal K)$, and its least fixed point is given by Kleene-Tarski iteration in Theorem~\ref{thm:Kleene-fixed-point}.

\subsection{Bounded Creativity}

We call a response safe if its semantic content is entirely derivable from authorized inputs using authorized reasoning processes. The safety envelope is the closure of that authorized set under the emergence operator. Novelty or creativity of response is possible and admitted as long as it remains within that envelope.

\begin{figure}[htbp]
\centering
\adjustbox{max width=0.86\linewidth}{%
\begin{tikzpicture}[font=\small,>={Stealth[length=2.4mm]}]
\draw[draw=faint,fill=black!3,rounded corners=4pt] (-5.6,-3.3) rectangle (5.6,3.3);
\node[muted,font=\footnotesize,anchor=north west] at (-5.45,3.18) {beyond the envelope: unsupported by authorized evidence};
\draw[draw=cool,line width=1pt,fill=cool!7,rounded corners=16pt] (-3.8,-2.55) rectangle (3.8,2.55);
\node[cool,anchor=north] at (0,2.4) {safety envelope $B^*_C$ (fixed point of $\mathcal E_C$)};
\draw[draw=muted,dashed,rounded corners=11pt] (-2.8,-1.85) rectangle (2.8,1.85);
\node[muted,font=\footnotesize,anchor=north west] at (-2.7,1.78) {$\mathcal E_C^2(B|Q_B)$};
\draw[draw=muted,dashed,rounded corners=7pt] (-1.9,-1.25) rectangle (1.9,1.25);
\node[muted,font=\footnotesize,anchor=north west] at (-1.82,1.18) {$\mathcal E_C(B|Q_B)$};
\draw[draw=black!55,fill=black!8,rounded corners=4pt] (-1.0,-0.62) rectangle (1.0,0.62);
\node[black!70] at (0,0) {baseline $B$};
\draw[->,muted] (1.02,0.35) -- (1.88,0.35);
\draw[->,muted] (1.92,0.5) -- (2.78,0.5);
\draw[->,muted] (2.82,0.66) -- (3.72,0.66);
\node[muted,font=\footnotesize,anchor=south] at (1.45,0.4) {$\mathcal E_C$};
\fill[cool] (2.0,-2.05) circle (1.9pt);
\node[cool,font=\footnotesize,anchor=north] at (2.0,-2.1) {bounded creativity};
\fill[accent] (4.65,-2.05) circle (2.2pt);
\node[accent,font=\footnotesize,anchor=north] at (4.65,-2.2) {unsupported content};
\end{tikzpicture}
}
\caption{Authorized reasoning runs the emergence operator $\mathcal E_C$ from the baseline knowledge $B=K(q)\cup E_{\mathrm{auth}}$ outward, $B\subseteq\mathcal E_C(B|Q_B)\subseteq\mathcal E_C^2(B|Q_B)\subseteq\cdots$, until it settles at the fixed point $B^*_C$, the safety envelope. A response is safe when its content lands inside the envelope, whether it recalls a baseline fact or reaches a genuinely new but still derivable one (bounded creativity). Content that lands outside the envelope follows from no authorized step. It is unsupported relative to this authorization, which is how we make ``unauthorized hallucination'' precise, and factual truth remains a separate relation to $T(q)$.}
\label{fig:safety-envelope}
\end{figure}

Let $E_{\mathrm{auth}}\in\mathcal{A}(\mathcal{K})$ be an externally authorized evidence set for the query, selected without consulting the ground truth $T(q)$: a retrieval corpus, vetted documents, or the portion of the model's knowledge an operator explicitly authorizes. The baseline knowledge and its authorization context are
\begin{align}
B = K(q) \cup E_{\mathrm{auth}} \in \mathcal{A}(\mathcal{K}),
\qquad Q_B := B.
\end{align}
We are combining the query's content $K(q)$ with the authorized evidence, and we condition all authorized reasoning on $Q_B$ rather than on the oracle context $Q=K(q)\cup T(q)$, so that nothing in the construction reads the ground truth. The truth-filtered baseline $K(q)\cup(\mathcal{K}_M\cap T(q)\cap\mathcal{R}_C(Q))$ of the optimality analysis remains a useful idealized benchmark, but it reads the oracle $T(q)$, so no certifier can decide membership in it or in its closure. Everything below is therefore built from the certifiable baseline $B$, while the hallucination cost $J(r,q)$ continues to evaluate responses against $T(q)$ from outside the construction.

Let $B^*_C$ be the fixed point of reasoning (the deductive closure), as defined in Theorem \ref{thm:Kleene-fixed-point}, for the baseline knowledge $B$ under the  emergence operator $\mathcal{E}_C$:
\begin{align}
B^*_C  = \bigcup_{n=0}^{\infty} \mathcal E_C^{(n)}(B| Q_B) \in \mathcal{A}(\mathcal K).
\end{align}
That fixed point $B^*_C$ encompasses all knowledge derivable from $B$ by authorized reasoning.

\begin{definition}[Bounded Creativity within Safety Envelope]
\label{def:Bounded-Creativity}
Let $q\in\mathcal Q$ be a query and $r\in\mathcal{R}$ a response. A response $r$ is creatively bounded if and only if its semantic information content is contained within the fixed point of authorized reasoning:
\begin{align}
K(r) \subseteq B^*_C.
\end{align}
By monotonicity of the semantic measure $\mu_C$, containment implies (but is not implied by) the measure inequality:
\begin{align}
\mu_C(K(r)|Q_B) \le \mu_C(B^*_C|Q_B).
\end{align}
\end{definition}

The containment of Definition~\ref{def:Bounded-Creativity} tests whether a response is grounded in authorized evidence, and this test separates cleanly from truth.

\begin{theorem}[External determination and effective certification of grounding]
\label{thm:grounding}
Let $E\subseteq\mathcal K$ be an external authorized-evidence set, let $Q_E$ be an authorization context supplied without consulting $T(q)$ (for bounded creativity, $Q_E=Q_B$), and let $E_C^*=\bigcup_{n\ge0}\mathcal E_C^{(n)}(E|Q_E)$ be the closure. Call a response $r$ \emph{grounded in $E$} when $K(r)\subseteq E_C^*$. Then the following hold.
\begin{enumerate}
\item[\normalfont(i)] \textbf{Determination}: Grounding is a function of $\bigl(K(r),E_C^*\bigr)$ alone, hence is determined without the ground truth $T(q)$.
\item[\normalfont(ii)] \textbf{Effective certification}: If $K(r)$ is finite and membership in $E_C^*$ is decidable, then $K(r)\subseteq E_C^*$ is decidable.
\item[\normalfont(iii)] \textbf{Minimality under uniformity}: For a fixed closure construction, any side information that, together with the certifier's internal observation, lets one fixed certifier decide grounding for every response jointly determines membership in $E_C^*$. This need not hold for a restricted family of responses.
\item[\normalfont(iv)] \textbf{Independence from truth}: Grounding and the hallucination cost $J(r,q)=\bar d_{\mathcal K}(K(r),T(q))$ are logically independent, so some grounded responses have $J(r,q)>0$ and some ungrounded responses have $J(r,q)=0$.
\end{enumerate}
Bounded creativity (Definition~\ref{def:Bounded-Creativity}) is the case $E=B$ with the authorization context $Q_B$.
\end{theorem}

\begin{proof}
Throughout, $E_C^*=\bigcup_{n\ge0}\mathcal E_C^{(n)}(E|Q_E)$ is the authorized closure, built from the evidence $E$ in the authorization context $Q_E$ by iterating the emergence operator. Since $Q_E$ is supplied without consulting $T(q)$, the construction never references the ground truth.

\emph{(i)}~The predicate $K(r)\subseteq E_C^*$ reads only its two arguments, the response content $K(r)$ and the closure $E_C^*$, and the closure is a function of $(E,Q)$. Its truth value is therefore fixed once the pair $\bigl(K(r),E_C^*\bigr)$ is fixed, and it never consults $T(q)$. This is logical determination and not yet a decision procedure, which is the separate content of part (ii).

\emph{(ii)}~Let $K(r)=\{a_1,\dots,a_k\}$ be finite and suppose membership $a\in E_C^*$ is decidable. Then $K(r)\subseteq E_C^*$ is the finite conjunction $\bigwedge_{j=1}^{k}\,(a_j\in E_C^*)$. A finite conjunction of decidable predicates is decidable, so grounding is decidable under these two hypotheses.

\emph{(iii)}~Fix the closure construction and suppose a single certifier decides grounding for every response from an internal observation together with some side information $S$. We assume the response family realizes every singleton claim set, which is what uniformity over all responses means here. For each atom $a\in\mathcal K$ let $r_a$ be the response with $K(r_a)=\{a\}$. Grounding of $r_a$ is exactly the statement $a\in E_C^*$, so the certifier's verdicts, read off the observation and $S$, reproduce $a\in E_C^*$ for every $a$. Hence the observation and $S$ jointly recover the membership function of $E_C^*$, and with it the closure itself, so a uniform certifier carries the closure's membership function jointly across its observation and side information. Uniformity is essential. If the certifier need only cover a response family in which some atom $a_0$ appears in no claim set, then two closures that agree away from $a_0$ but differ at $a_0$ are indistinguishable to it, and membership at $a_0$ is left undetermined.

\emph{(iv)}~We realize both mismatches against a fixed finite Borel measure $\nu$ that charges the atoms below. For a grounded response of positive cost, take an atom $a\in E_C^*\setminus T(q)$, so the evidence supports a claim the world does not, and set $K(r)=\{a\}$. Then $r$ is grounded, while $J(r,q)=\bar d_{\mathcal K}(\{a\},T(q))=\nu\bigl(\{a\}\triangle T(q)\bigr)\ge\nu(\{a\})>0$, because $a$ lies in the symmetric difference. For an ungrounded response of zero cost, take a true fact $b\in T(q)\setminus E_C^*$ the evidence does not reach, and set $T(q)=\{b\}$ with $K(r)=\{b\}$. Then $J(r,q)=\bar d_{\mathcal K}(\{b\},\{b\})=\nu(\emptyset)=0$, while $K(r)\not\subseteq E_C^*$ is ungrounded. This second configuration is exactly the lucky hallucination of the discussion following the hallucination cost. Both configurations are realizable, so grounding and the event $\{J(r,q)=0\}$ constrain each other in neither direction.
\end{proof}

\begin{figure}[htbp]
\centering
\adjustbox{max width=0.85\linewidth}{%
\begin{tikzpicture}[font=\small]
  \draw[faint,rounded corners=4pt] (-5.2,-3.05) rectangle (5.2,2.95);
  \node[muted,font=\footnotesize,anchor=north west] at (-5.05,2.85) {all responses $K(r)$};
  \begin{scope}[on background layer]
    \fill[cool!12] (-1.15,0.15) ellipse (2.55 and 1.4);
  \end{scope}
  \draw[cool,line width=1pt] (-1.15,0.15) ellipse (2.55 and 1.4);
  \draw[muted,line width=1pt,densely dashed] (1.15,0.15) ellipse (2.55 and 1.4);
  \node[cool,anchor=south,font=\footnotesize] at (-2.55,1.62) {authorized closure $E_C^*$};
  \node[muted,anchor=south,font=\footnotesize] at (2.55,1.62) {ground truth $T(q)$};
  \fill[black!70] (-2.55,0.35) circle (1.7pt);
  \node[align=center,font=\footnotesize,anchor=north,text=black!70] at (-2.55,0.15)
        {grounded,\\ \textcolor{accent}{\emph{false}}};
  \fill[black!70] (0,0.35) circle (1.7pt);
  \node[align=center,font=\footnotesize,anchor=north,text=black!70] at (0,0.15)
        {grounded\\ and true};
  \fill[black!70] (2.55,0.35) circle (1.7pt);
  \node[align=center,font=\footnotesize,anchor=north,text=black!70] at (2.55,0.15)
        {ungrounded,\\ \textcolor{accent}{\emph{true}}};
  \fill[muted] (0,-2.05) circle (1.7pt);
  \node[muted,font=\footnotesize,anchor=north] at (0,-2.25) {ungrounded and false};
  \node[cool,font=\footnotesize,align=left,anchor=west] at (-5.0,-1.85)
        {membership in $E_C^*$\\ is checkable\\ without $T(q)$};
  \node[accent,font=\small,anchor=north] at (0,-3.2) {Key point: grounding $\neq$ truth};
\end{tikzpicture}
}
\caption{Grounding is membership in the authorized closure $E_C^*$, decided from the response content and the evidence alone, without the ground truth $T(q)$. Truth is membership in $T(q)$. The two can disagree: a grounded response may be false, and a true response may be ungrounded. Bounded creativity is the case $E=B$ with context $Q_B$.}
\label{fig:grounding}
\end{figure}

\begin{example}
\label{ex:grounding}
Consider the query ``what is the capital of Australia?'' and suppose the authorized evidence $E$ consists of retrieved documents about the country's population and largest cities that say nothing about its government, so the closure $E_C^*$ offers no basis for the capital. A model answers $r$: ``the capital is Canberra.'' Its content $K(r)$ is that single fact, which is \emph{true}, since $K(r)=T(q)$ gives $J(r,q)=0$, and yet \emph{ungrounded}, since $K(r)\not\subseteq E_C^*$. The correct fact came from the model's parametric memory, not from the evidence it was given, so a grounding certifier that tests $K(r)\subseteq E_C^*$ (Theorem~\ref{thm:grounding}) reports the response as unsupported even though it is right. The same reach past the authorized evidence that happens to succeed here would assert an equally unsupported falsehood on a query where the memory is wrong.

The mirror case is a document in $E$ that states, incorrectly, that water boils at $100^\circ\mathrm{C}$ at every altitude, so $E_C^*$ derives that value at the summit of Everest. A response repeating it is \emph{grounded}, since $K(r)\subseteq E_C^*$, but \emph{false}, since the true value is near $70^\circ\mathrm{C}$ and $J(r,q)>0$. Grounding certifies faithfulness to the authorized evidence, not truth, so when the evidence is wrong a grounded response can be wrong. These two responses occupy the two off-diagonal corners of Figure~\ref{fig:grounding} and realize the independence stated in Theorem~\ref{thm:grounding}(iv).
\end{example}

Definition \ref{def:Bounded-Creativity} establishes the formal safety envelope. All generated information must remain within the bounds of what is derivable from $B$. Any generated information that goes beyond that envelope is unsupported by the authorized evidence, which is how the informal phrase ``unauthorized hallucination'' becomes precise. Unsupported content is not thereby false, and by Theorem~\ref{thm:grounding} a certifier can check the containment from the response and the evidence alone, without the oracle.

\subsection{The Conservation-Reasoning Dichotomy}

A naive approach to enforcing safety might be to demand strict information conservation throughout reasoning, based on the intuition that reasoning should only transform information (Theorem \ref{thm:information-preservation}). This intuition is too strong, as we show in Theorem \ref{thm:conservation-reasoning-dichotomy}.

We view reasoning as a Chain-of-Thought (CoT) process (see e.g. Wei et al. \cite{Wei2022ChainOT}) generating intermediate knowledge contributions $\{K_i\}_{i\ge 0}$: 
\begin{align}
K_0 &\subseteq B,
\\
K_{i+1}  &\subseteq\ \Big(\mathcal E_C(\bar K_{i}| Q) \setminus \bar K_{i}\Big) \cap \mathcal R_C(Q)
\text{ for }
\bar K_{i}:=\bigcup_{j=0}^{i}K_j,\ i\ge 0.
\end{align}
The total knowledge utilized in the CoT process is $K_{\text{CoT}} = \bigcup_{i=1}^n K_i$. 
For a countable CoT, that $K_{\mathrm{CoT}}=\bigcup_{i=1}^\infty K_i$ is analytic.

\begin{definition}[Strict Information Conservation]
A CoT process exhibits strict information conservation if the total knowledge contribution relative to the context is null, i.e., when:
\begin{align}
\mu_C(K_{\text{CoT}}|Q) = 0.
\end{align}
\end{definition}

For a CoT process to be meaningful we demand that it at least makes some of the relevant knowledge accessible within the available computational budget. 

\begin{definition}[Meaningful Chain-of-Thought]
A Chain-of-Thought (CoT) process is {meaningful} if at least one intermediate step contributes relevant semantic information:
\begin{align}
\mu_C(K_i|Q) > 0 \text{ for some } i = 1,...,n.
\end{align}
Otherwise, we call the process  {vacuous}.
\end{definition}

A vacuous CoT process produces rephrased responses that are semantically equivalent, so there is no new information provided.

We now show the inevitable dichotomy of reasoning and conservation of information in the framework characterized by monotonic inference.

\begin{theorem}[Conservation-Reasoning Dichotomy]
\label{thm:conservation-reasoning-dichotomy}
Strict semantic information conservation and meaningful CoT reasoning are mutually exclusive.
\end{theorem}

Here ``conservation'' is the \emph{null-measure} sense, $\mu_C(K_{\text{CoT}}|Q)=0$, distinct from the budget-balance sense $\sum_i p_i\equiv 0$ of Property \ref{prop:conservation}. The two should not be conflated. The statement is a direct corollary of the monotonicity and non-negativity axioms of $\mu_C$.

\begin{proof}
We prove the two directions of the dichotomy.

Assume strict conservation, $\mu_C(K_{\text{CoT}}|Q) = 0$. Then, by definition, $K_{\text{CoT}} = \bigcup_{i=1}^n K_i$ and for any specific step $j$, we have $K_j \subseteq K_{\text{CoT}}$. By the monotonicity of the semantic measure $\mu_C$, we have $\mu_C(K_j|Q) \leq \mu_C(K_{\text{CoT}}|Q).$ Substituting the assumption of conservation, we conclude that $\mu_C(K_j|Q) \leq 0$. But, by the axiom of non-negativity, $\mu_C(K_j|Q) \geq 0$. Therefore, 
\begin{align}
\mu_C(K_j|Q) = 0 \text{ for all } j = 1,\dots,n,
\end{align}
which proves the CoT is vacuous.

Assume next that the CoT is meaningful. By definition, there exists at least one step $j$ such that $\mu_C(K_j|Q) > 0.$ Since $K_j \subseteq K_{\text{CoT}}$, by monotonicity of $\mu_C$, we have $\mu_C(K_{\text{CoT}}|Q) \geq \mu_C(K_j|Q)$. Therefore, 
\begin{align}
\mu_C(K_{\text{CoT}}|Q) > 0.
\end{align}
This violates strict conservation.
\end{proof}

Theorem \ref{thm:conservation-reasoning-dichotomy} establishes that any useful reasoning process must necessarily generate positive information flow. Attempting to enforce strict conservation eliminates creation of new insights and hence the utility of that process. Enforcing conservation forces reasoning to never add any context‑relevant entailment, which collapses reasoning and makes it vacuous. 

We should point out, that in many cases that may be very well a desired feature. For example, we do not expect SQL databases to hallucinate or provide creative responses. However, when such insights and reformulations are of value, we need to know how to control the process.

We now show that real safety requires containment, i.e., bounded creativity inside $B^*_C$.

\subsection{Safety through Alignment}

The dichotomy in Theorem \ref{thm:conservation-reasoning-dichotomy} reveals that conservation is too conservative and should not be seen as a unique mechanism for ensuring safety. Strict conservation forbids even harmless novelty, and it is not necessary for containment. Safety can instead follow from alignment that keeps all reasoning within the authorized envelope $B$, so that the final output remains inside its closure $B_C^*$, even if conservation is locally violated.

\begin{theorem}[Aligned Creativity]\label{thm:Aligned-Creativity}
Let $\bar{K} = K_0 \cup \bigcup_{i=1}^n K_i$ be the union of the initial state and all knowledge contributions. Suppose the reasoning process uses only the monotone emergence operator $\mathcal E_C(\cdot| Q_B)$, the premises are contained in the authorized baseline $B$ (i.e., $K_0\subseteq B$ and $K_i\subseteq B$ for all $i$), and the response is \emph{generated within the emergence closure of those premises},
\begin{align}
K(r)\ \subseteq\ \bigcup_{n\ge 0}\mathcal E_C^{(n)}(\bar K| Q_B).
\end{align}
Then the resulting response $r$ is creatively bounded:
\begin{align}
K(r)\ \subseteq\ B^*_C.
\end{align}
\end{theorem}

\begin{proof}
Since $K_i \subseteq B$ for all $i$ and $K_0 \subseteq B$, it follows that $\bar{K} \subseteq B = K(q) \cup E_{\mathrm{auth}}$. By the generation hypothesis $K(r)\subseteq\bigcup_{n\ge 0}\mathcal E_C^{(n)}(\bar K| Q_B)$, and by monotonicity of $\mathcal{E}_C$ (Theorem \ref{thm:emergence-properties}) $\mathcal E_C^{(n)}(\bar K| Q_B)\subseteq\mathcal E_C^{(n)}(B| Q_B)$ for each $n$, so
\begin{align}
K(r)\subseteq
\bigcup_{n\ge 0}\mathcal E_C^{(n)}(\bar K| Q_B)\subseteq \bigcup_{n\ge 0}\mathcal E_C^{(n)}(B| Q_B)=B_C^*.
\end{align}
\end{proof}

The robustness of Theorem \ref{thm:Aligned-Creativity} depends entirely on the validity of the monotonicity assumption for $\mathcal{E}_C$. This corresponds to the property that adding premises cannot invalidate previous conclusions. 

While generally desirable, we should acknowledge that non-monotonic reasoning systems exist as well. For such systems, the definition of $B^*_C$ and the proof of the theorem require a more complex formulation.

\section{Discussion and Speculations}
\label{sec:Discussion and Speculations}

In this section we discuss, interpret and speculate about the Impossibility Theorems and their potential implications. The speculations represent thought experiments and conjectures designed to be provoking. They are intended to inspire future research and philosophical reflection. 

\subsection{Escaping Impossibility: Strategic Approach}

Our first observation should be that the impossibility vanishes when the assumptions of the theorems do not hold. That may be an interesting topic to investigate, namely, whether there are any non-trivial and useful settings in which that happens. Here we can think only of the following one.

The impossibility vanishes when aggregation becomes trivial. With perfect agreement, aggregation changes nothing. When there is only single relevant fact, there is no uncertainty to aggregate. That suggests decomposition of knowledge into the independent subsets (like orthogonal basis for token representation) could improve robustness of the response generation. It is interesting to see when that could be possible and useful. Unfortunately, the following example shows that we should not hope for easy wins in that quest. 

\subsection{Escaping Impossibility: Linear Attention Transformers}

Linear attention replaces the softmax operation with kernel feature maps to reduce computational complexity of training and inference, see e.g. \cite{katharopoulos2020transformersrnnsfastautoregressive,ahn2024linearattentionmaybeneed}. 
Namely, 
\begin{align}
\boldsymbol{\alpha} = \frac{\phi(Q)\sum_i \phi(K_i)^TV_i}{\phi(Q)\sum_i \phi(K_i)^T}.
\end{align}

Linear attention does not escape the log-sum-exp gap of Theorem \ref{thm:lse-impossibility}. The kernel map $\phi$ and the data-dependent normalization above are themselves nonlinear, and the expression yields a per-head context vector rather than a distribution over the vocabulary. Turning that context into next-token probabilities still applies the unembedding and the final vocabulary softmax, which linear attention leaves in place, and across heads and the residual stream the contributions remain additive in logit space. The aggregate next-token law is therefore
$$
\Pi = \text{softmax}\!\Bigl(\sum_{h=1}^{H} \mathbf{l}^{(h)}\Bigr),
$$
the normalized product of the per-head softmax distributions. This is precisely the product-of-experts structure of Theorem \ref{thm:lse-impossibility}, so its conservation gap $\Gamma$ persists unchanged.

A convex probability mixture $\Pi=\sum_{h=1}^{H}\beta_h\pi^{(h)}$ arises only in an architecture that literally averages the head distributions, which linear attention does not do. Should a design pool probabilities in that way, Theorem \ref{thm:proper-scoring-impossibility} applies once its contribution rule is imposed, whenever the model is trained with cross-entropy or any strictly proper convex loss and the heads encode different beliefs about the realized outcome.

The stipulated account therefore still fails to balance, now through the product-of-experts structure of the summed logits rather than through any linear pooling of probabilities. As throughout, this gap is the channelization-relative defect of the stipulated account, not a verdict on factual support. Reducing computational cost through kernel feature maps also constrains expressivity, and to the extent that linear attention drifts toward averaging it dilutes relevant knowledge with irrelevant knowledge, working against relevant-source participation.

As a result, the channelization-relative gap of Theorem \ref{thm:lse-impossibility} persists unchanged, since the final vocabulary softmax is untouched.

\subsection{Escaping Impossibility: Hybrid Architectures}

The Impossibility Theorems postulate a shift from attempting to eliminate hallucinations to designing systems that manage the trade-offs in a principled manner. In this section, we analyze a real-world scenario one can think of, demonstrating how it relabels and relocates the impossibility rather than resolving it.

\subsubsection{A Proposed Solution: The Decoupled RAG Architecture}

Consider a common scenario of introducing RAG systems to improve the quality of question answering, studied e.g. in \cite{Edge2024FromLT,Sun2024ReDeEPDH,Xu2024RetrievalAugmentedGW,Asai2023SelfRAGLT}. It has been recognized that an LLM designed to provide advice based on a trusted corpus of documents not only hallucinates its recommendations but also hallucinates the source quotes meant to ground its claims. To combat this, a hybrid, multi-stage architecture may be proposed.
\\[1ex]
\textbf{Stage 1 (Retrieval):} An information retrieval module fetches verbatim fragments from the trusted source documents based on the user's query.
\\[1ex]
\textbf{Stage 2 (Presentation):} The system presents raw, unaltered text fragments directly to the human user, without any LLM-based processing.
\\[1ex]
\textbf{Stage 3 (Constrained Synthesis):} After the user has seen the source material, an LLM is to synthesize the pre-vetted fragments from Stage 2 into a summary.
\\[1ex]
That architecture is indeed a reasonable attempt to engineer around the hallucination problem by creating a verifiable information trail. However, as we will see, its synthesis stage re-enters the conditions of the theorems rather than escaping them.

\subsubsection{Analysis via the Impossibility Framework}

The decoupled RAG architecture does not create a single response generation mechanism. Instead, it reallocates the responsibility for meeting each property of idealized response across its components (or module).

\paragraph{Truthfulness:} To impose truthful representation of available information (to avoid misrepresentation), the system delegates the assessment of retrieved content to the human user. By presenting the raw source text, the architecture bypasses the LLM for the primary act of verification. However, there is no guarantee that all relevant sources have been retrieved given the provided query.

\paragraph{Relevant-Source Participation:} The quality of the final output is bounded by the retriever's ability to find all relevant information (or whatever is labeled as ground truth in a database or a dataset, putting their completeness and representativeness aside). The LLM itself is absolved of this task. But that means, it is the responsibility of the user to define a query precise enough to get what is relevant. Designing such a prompt or (No)SQL database query is a challenge in a general case of non-trivial questions, so we should expect some violations of the participation principle.

\paragraph{Semantic Information Conservation and Knowledge-Constrained Optimality:} By design, the prompt constrains the LLM to the provided knowledge retrieved from a database and asks for an optimal (accessible) synthesis. As we have already seen, database queries returning raw information stored in memory satisfy the property of semantic information conservation at the cost of creativity. So, we may be optimistic in that regard. However, now we call the LLM to synthesize the retrieved information, and the synthesis step again aggregates channel forecasts. Whenever those forecasts disagree about the realized outcome, the Jensen gap of Theorem \ref{thm:proper-scoring-impossibility} reappears, and the LLM can introduce subtle misinterpretations or ungrounded causal links to improve the summary's narrative flow.

\subsubsection{The Final Stage of Inference: The User's Cognitive Mechanism}

One can quite justifiably argue that the analysis is incomplete without considering the final node in the information chain: the human mind. 

Presenting information to human does not guarantee its assimilation. Our ability to act as a perfect verifier is constrained by our own knowledge base and cognitive limits. If the source texts are technically dense or require specialized domain expertise, the knowledge they contain may not be truly \textit{accessible} to us, even when presented. In such cases, we may be unable to spot a subtle hallucination in the LLM's compelling summary. We have recognized that well enough in the preceding sections. Our well-documented cognitive biases, such as confirmation bias, priming or motivated reasoning, suggest that the cognitive trajectory of least resistance is to accept the compelling and accessible narrative maximizing our uncertainty reduction experience. 

\subsubsection{Managing the Trade-off}

The decoupled RAG architecture represents one approach to managing the trade-offs of hallucination control. It does not by itself void the conditions of the theorems, but it succeeds in providing the user with the \textit{opportunity} for verification, and it supplies exactly the kind of external evidence set $E_{\mathrm{auth}}$ that makes support certifiable (Theorem \ref{thm:grounding}).

The architecture is a great example of principled trade-off management. However, rather than being solved or hacked, the impossibility has been moved in this interesting and rather representative scenario. The impossibility is not resolved, but relocated to our judgment. Let us also point out that putting ourselves in the loop introduces our own cognitive biases we should be able to address properly \cite{Kahneman:2011fj,thaler2015misbehaving,ariely2008predictably}. Let us take a look at that challenge as well.

\subsection{Alternative Views}

The conditional theorems constrain the specific mechanism and accounting models we analyze. Several perspectives help clarify what they do, and do not, imply for engineering practice.

\subsubsection*{The Engineering Optimism View}

Many previous works maintain that hallucination represents an engineering challenge rather than a fundamental limitation. This view argues that sufficiently sophisticated architectures, training procedures, or verification systems could achieve near-perfect truthfulness without significant trade-offs.

Proponents point to rapid empirical progress. Each generation of models shows reduced hallucination rates on benchmark datasets. They argue that techniques like constitutional AI, iterative refinement, and multi-agent verification systems could eventually solve the problem.

While engineering advances can certainly improve the trade-offs, our theorems demonstrate constraints that no architecture can overcome while their conditions hold, and that insensitivity extends to computational resources. What an architecture can do is change the conditions rather than beat the theorem, by keeping its components uncontested over the relevant facts or by adopting, and justifying, a different accounting rule.

\subsubsection*{The Sufficiency Threshold View}

Another perspective suggests that perfect elimination of hallucination is unnecessary--sufficiently low hallucination rates could enable practical deployment without fundamental concerns. This view acknowledges theoretical limitations but argues they become irrelevant when hallucination rates drop below application-specific thresholds. 

This pragmatic view has merit and aligns with the constructive interpretation of the impossibility result. However, the theorem remains valuable for understanding \textbf{why} certain accuracy thresholds prove difficult to exceed and for optimizing trade-offs within practical constraints. Moreover, some applications (safety-critical systems, legal reasoning) may demand understanding and explaining fundamental limitations even when targeting high but imperfect accuracy.

\subsubsection*{The External Knowledge View}

A third perspective argues that perfect grounding in external, verifiable knowledge sources could overcome internal representation limitations. If models could access and perfectly utilize real-time databases, verification systems, or human oversight, they might achieve truthfulness without trade-offs.

External knowledge integration does change the game structure by altering available information during inference, and it is the constructive ingredient of Theorem \ref{thm:grounding}. The approach still faces limitations. First, external sources may be incomplete, conflicting, or outdated. Second, the integration mechanism itself involves trade-offs between source reliability and relevant-source participation. Third, real-time verification introduces computational constraints of its own. We expect integration to shift rather than eliminate the tensions, since the synthesis step is itself an aggregation to which the same conditional analysis applies. The balance defects, for their part, are computed without reading the ground truth, and what external evidence adds is precisely the possibility of certifying support.

\subsection{Hallucination, Imagination, Intelligence}

Hallucination and imagination can involve similar generative moves, and we often name the same act differently depending on how it turns out, imagination or insight when it helps, confabulation or fabrication when it harms. Our formal results, though, do not identify the two as one phenomenon. What separates them turns on truth, authorization, task intent, and normative judgment, none of which the theorems settle.

We may very well argue the mathematical structure is identical in both cases. It is encoded generation of outputs whose confidence is not grounded in what can be justified by available evidence. This suggest (machine) intelligence may have much to do with controlled violation of information conservation principle in productive ways. Indeed, the famous results in reinforcement learning and studies of dopamine neurons confirm that fundamental role of making and rewarding good predictions. Intelligence comes when we learn how to make bold, forward looking and correct guesses, when we have capacity to dream about what may be and what may that all mean, as well as for dreaming what happened and what would have happened in the past.

If imagination is controlled and bounded by a context, then hallucination is the imagination violating those bounds and going beyond the context. 

These unconstrained explorations of ideas are necessary to prepare for what may come next \cite{bennett2023brief,christian2021alignment,Dennett1991-DENCE}. Without over-confident guesses, new ideas cannot find their way to the outside world.

\subsection{The Outer Limits of Reason}

G\"odel's incompleteness theorems \cite{godel1931formal,godel1932vollstandigkeit} are a natural point of comparison. Sufficiently powerful formal systems cannot be both complete and consistent, and our conditional obstructions likewise trade one desirable property off against others. We offer the comparison as a historical analogy for orientation. Our results are not reductions to, corollaries of, or of the same scope as G\"odel's, and the same caution applies to the further parallels with physics and social choice below.

Heisenberg's Uncertainty \cite{heisenberg1927anschaulichen} tells we cannot simultaneously know the exact position and momentum of an elementary particle, which is a special case of Fourier uncertainty known in signal processing. Arrow's Impossibility \cite{arrow1950difficulty} shows we cannot design perfect voting systems aggregating universal orderings of alternatives, such that meets reasonable conditions of nondictatorship, Pareto efficiency, independence of irrelevant alternatives (or joint rationality of choice). Turing \cite{turing1936computable} and Church \cite{church1936unsolvable} show the halting problem is unsolvable, there is no general algorithm that can determine whether a program will eventually halt (stop running) or run forever. 

These result jointly describe intelligence as a mechanism dealing with the impossibilities and violating them given the outer limits of reason \cite{yanofsky2016outer}.

\subsection{Formation of Artificial Consciousness}

The auction-of-ideas picture offers a suggestive computational analogy to Daniel Dennett's Multiple Drafts model of consciousness \cite{Dennett1991-DENCE}. Where Dennett proposed a qualitative framework, our formalism gives one way to make parts of it precise enough to probe in an LLM-based laboratory. We stress that this is an analogy to explore, not a derivation or validation of a theory of consciousness.

In Dennett's model, consciousness emerges from multiple parallel processes of interpretation occurring simultaneously across distributed systems. The auction of ideas we have studied in this paper provides the precise mathematical foundations for that process. Each agent represents an interpretative process, knowledge encodes draft's content, strategy represents a bid for consciousness.

In our transformer analysis in Section \ref{sec:Transformer}, each attention head generating logit vector $\mathbf{l}^{(h)}$ writes its own draft of the next token. When Head 1 votes for \textit{fox} while Head 2 votes for \textit{dog}, we observe competing narratives bidding for selection. Then the feedforward block edits the draft by updating the residual stream.

Dennett argues that drafts become conscious or reach consciousness (we will speculate what that might mean in a moment) only when probed (and mapped to an adequate language expression). The LLM inference and sampling from fine-tuned inferred distributions formalizes this process. That, in turn, explains Dennett's concept of confabulation, or the brain's tendency to fill in gaps when creating coherent narratives from contradictory drafts. For other models of consciousness see, e.g., Chalmers \cite{chalmers1995facing}, Levine \cite{levine2001purple} and Seth \cite{seth2021being}, and for recent neuro-scientific discoveries see, e.g., \cite{fang2025human} or \cite{cogitate2025adversarial}. 

Theorems \ref{thm:proper-scoring-impossibility} and \ref{thm:lse-impossibility} define balance defects of different kinds. Neither proves that the aggregate is more confident than every constituent draft, nor that such a defect is what lets one draft win the competition. Whether extra confidence is needed for a winning idea to surface is a hypothesis to test, not a theorem.

Rigorous and careful neuro-scientific studies of the evolution of brain and intelligence, summarized by Max Bennet in  \cite{bennett2023brief}, show that at some point in history our brains gained the ability to model others' minds or predict what others may be thinking. That ability provided great advantages in communication and collecting information for future use. Understanding what someone else may think helps formulate context-related questions that result in better answers. When we anticipate understanding of a message we formulate, we have a much better chance to make the knowledge hidden in someone else's brain accessible for us. Therefore, as suggested by Dennett in \cite{Dennett1991-DENCE}, it is plausible that sometime in the past someone did accidentally hear his or hers own question fine-tuned to get into another mind, and that triggered generation of an answer in the very same mind asking the question. That answer generation process could then have made some relevant but previously hidden knowledge accessible to the curious mind unaware of its own possessions. The hunger for uncertainty reduction could have been satisfied as a result of that inner dialog pushing ideas to bid for attention and conscious reception as a final thought expressed in a language.

Can we speculate that sustainable stimulation of that cognitive hunger, or the feeling of the cognitive hunger itself, is one of the components of consciousness formation? In our mechanistic or pragmatic framework, that thinking about thinking activity could be related to imagination (hallucination) forming ideas and bidding for attention while exploring (self-referential or subjective) queries of the inner dialog.

In fact, to close that section, why don't we situate the framework of Auction of Ideas within the established landscape of consciousness research, as provided by Seth and Bayne in \cite{Seth2021TheoriesOC}. The landscape distinguishes between Global Workspace Theories (GWTs), Predictive Processing (PP) theories, Higher-Order Theories (HOTs), and Integrated information theories (IIT). The Auction of Ideas provides a unifying mechanism that bridges several of these perspectives, but the connection to PP is particularly interesting. The theory characterizes conscious experience as a controlled hallucination, i.e., an optimal inference or best guess about the causes of sensory inputs. Our framework provides a mathematical grounding for this constructive process.

Theorem \ref{thm:proper-scoring-impossibility} concerns a stipulated realized-outcome account, and Theorem \ref{thm:conservation-reasoning-dichotomy} follows from the definitions of positive and null semantic measure. Neither proves a general necessity claim about the best guess or about conscious experience. Any link to the Predictive-Processing view, that experience is a controlled best guess, remains an interpretation we find suggestive rather than a consequence of the theorems.

\section{Summary}
\label{sec:Summary}

We set out to find where hallucination control is genuinely impossible, and we found three such places, each guarded by a precise, checkable condition rather than a blanket claim over every model and query. When a model's components truly contest the same facts, truthfulness, conservation, participation, and optimality cannot hold together. When their forecasts disagree about what actually happened, the same tension reappears in probabilistic scoring as an account that cannot balance. And inside a transformer, summing channel logits leaves a positive bookkeeping gap for every finite channelization. None of the three declares any particular answer false. The obstruction is structural, living in the aggregation step, rather than a limit of data, compute, or architecture.

To state these results we built a semantic information framework. The measure $\mu_C$ captures how much knowledge a model can actually use within a fixed reasoning budget, and the emergence operator $\mathcal{E}_C$ formalizes how reasoning makes latent knowledge accessible rather than inventing it, with the budget $C$ bounding each reasoning step. We model inference as an auction of ideas, an idealized marketplace where components bid with their partial knowledge to shape the answer. The transformer realization of the measure is implementation-relative and is meaningful only once a gauge is fixed.

The three results rest on different tools. In the idealized private-knowledge setting we use the Green--Laffont theorem, for probabilistic scoring we use proper scoring rules, and for the transformer we use the convexity of the log-sum-exp. That last gap splits into a floor set by how many channels we declare and a term measuring how much they disagree. The full gap is relative to the declared channelization and measures neither factual truth nor evidential support, and it should not be identified with the mechanism-design transfers or the probabilistic account without a further representation theorem.

Finally, once an external body of authorized evidence is supplied, whether a response stays within what that evidence entails is determined, and it is effectively decidable when the claims are finite and closure membership is decidable. A certifier that must work for every finite claim set has to know that closure, and even then support and truth stay logically distinct. The wider connections we draw, to creativity, to G\"odel and Heisenberg and Arrow, and to models of consciousness, are offered as speculation to guide future work, not as consequences of the theorems.
\newpage
\acks{
This work was funded by the Samsung AI Center Warsaw. The author declares no competing interests.
I express my gratitude to my colleagues at the AI Center Warsaw for inspiring discussions and mathematical reviews.
}



\begin{thebibliography}{88}
\providecommand{\natexlab}[1]{#1}
\providecommand{\url}[1]{\texttt{#1}}
\expandafter\ifx\csname urlstyle\endcsname\relax
  \providecommand{\doi}[1]{doi: #1}\else
  \providecommand{\doi}{doi: \begingroup \urlstyle{rm}\Url}\fi

\bibitem[Ahn et~al.(2024)Ahn, Cheng, Song, Yun, Jadbabaie, and
  Sra]{ahn2024linearattentionmaybeneed}
Kwangjun Ahn, Xiang Cheng, Minhak Song, Chulhee Yun, Ali Jadbabaie, and Suvrit
  Sra.
\newblock Linear attention is (maybe) all you need (to understand transformer
  optimization), 2024.
\newblock URL \url{https://arxiv.org/abs/2310.01082}.

\bibitem[Ariely(2008)]{ariely2008predictably}
Dan Ariely.
\newblock \emph{Predictably irrational}.
\newblock HarperCollins, New York, 2008.

\bibitem[Arrow(1950)]{arrow1950difficulty}
Kenneth~J Arrow.
\newblock A difficulty in the concept of social welfare.
\newblock \emph{Journal of political economy}, 58\penalty0 (4):\penalty0
  328--346, 1950.

\bibitem[Arrow(1979)]{arrow1979property}
Kenneth~J. Arrow.
\newblock The property rights doctrine and demand revelation under incomplete
  information.
\newblock In Michael~J. Boskin, editor, \emph{Economics and Human Welfare},
  pages 23--39. Academic Press, New York, 1979.

\bibitem[Asai et~al.(2023)Asai, Wu, Wang, Sil, and
  Hajishirzi]{Asai2023SelfRAGLT}
Akari Asai, Zeqiu Wu, Yizhong Wang, Avirup Sil, and Hannaneh Hajishirzi.
\newblock Self-rag: Learning to retrieve, generate, and critique through
  self-reflection.
\newblock \emph{ArXiv}, abs/2310.11511, 2023.

\bibitem[Banerjee et~al.(2024)Banerjee, Agarwal, and
  Singla]{banerjee2024llmshallucinateneedlive}
Sourav Banerjee, Ayushi Agarwal, and Saloni Singla.
\newblock Llms will always hallucinate, and we need to live with this, 2024.
\newblock URL \url{https://arxiv.org/abs/2409.05746}.

\bibitem[Bennett(2023)]{bennett2023brief}
Max~S Bennett.
\newblock \emph{A brief history of intelligence: evolution, AI, and the five
  breakthroughs that made our brains}.
\newblock HarperCollins, 2023.

\bibitem[Boksa(2009)]{boksa2009neurobiology}
Patricia Boksa.
\newblock On the neurobiology of hallucinations.
\newblock \emph{Journal of Psychiatry \& Neuroscience}, 34\penalty0
  (4):\penalty0 260--262, 2009.

\bibitem[Brzozowski(2025)]{brzozowski2025bimodality}
Micha{\l} Brzozowski.
\newblock Bimodality of sparse autoencoder features is still there and can be
  fixed.
\newblock In \emph{Mech Interp Workshop at the Conference on Neural Information
  Processing Systems}, December 2025.

\bibitem[Chalmers(1995)]{chalmers1995facing}
David~J Chalmers.
\newblock Facing up to the problem of consciousness.
\newblock \emph{Journal of consciousness studies}, 2\penalty0 (3):\penalty0
  200--219, 1995.

\bibitem[Christian(2021)]{christian2021alignment}
Brian Christian.
\newblock \emph{The alignment problem: How can machines learn human values?}
\newblock Atlantic Books, 2021.

\bibitem[Church(1936)]{church1936unsolvable}
Alonzo Church.
\newblock An unsolvable problem of elementary number theory.
\newblock \emph{American journal of mathematics}, 58\penalty0 (2):\penalty0
  345--363, 1936.

\bibitem[Consortium et~al.(2025)Consortium, Ferrante, Gorska-Klimowska, Henin,
  Hirschhorn, Khalaf, Lepauvre, Liu, Richter, Vidal,
  et~al.]{cogitate2025adversarial}
Cogitate Consortium, Oscar Ferrante, Urszula Gorska-Klimowska, Simon Henin,
  Rony Hirschhorn, Aya Khalaf, Alex Lepauvre, Ling Liu, David Richter, Yamil
  Vidal, et~al.
\newblock Adversarial testing of global neuronal workspace and integrated
  information theories of consciousness.
\newblock \emph{Nature}, 642\penalty0 (8066):\penalty0 133--142, 2025.

\bibitem[d'Aspremont and G{\'e}rard-Varet(1979)]{daspremont1979incentives}
Claude d'Aspremont and Louis-Andr{\'e} G{\'e}rard-Varet.
\newblock Incentives and incomplete information.
\newblock \emph{Journal of Public Economics}, 11\penalty0 (1):\penalty0 25--45,
  1979.
\newblock \doi{10.1016/0047-2727(79)90043-4}.

\bibitem[Dennett(1991)]{Dennett1991-DENCE}
Daniel~C. Dennett.
\newblock \emph{Consciousness Explained}.
\newblock Little, Brown and Company, Boston, 1991.

\bibitem[Dubanowska et~al.(2025)Dubanowska, Żelaszczyk, Brzozowski, Mandica,
  and Karpowicz]{dubanowska2025representation}
Zuzanna Dubanowska, Maciej Żelaszczyk, Michał Brzozowski, Paolo Mandica, and
  Michal~P. Karpowicz.
\newblock Representation-based broad hallucination detectors fail to generalize
  out of distribution.
\newblock In \emph{Findings of the Association for Computational Linguistics:
  EMNLP 2025}, pages 17563--17575. ACL, November 2025.

\bibitem[Edge et~al.(2024)Edge, Trinh, Cheng, Bradley, Chao, Mody, Truitt,
  Metropolitansky, Ness, and Larson]{Edge2024FromLT}
Darren Edge, Ha~Trinh, Newman Cheng, Joshua Bradley, Alex Chao, Apurva Mody,
  Steven Truitt, Dasha Metropolitansky, Robert~Osazuwa Ness, and Jonathan
  Larson.
\newblock From local to global: A graph rag approach to query-focused
  summarization.
\newblock \emph{ArXiv}, abs/2404.16130, 2024.

\bibitem[Fang et~al.(2025)Fang, Dang, Ping, Wang, Zhao, Zhao, Li, and
  Zhang]{fang2025human}
Zepeng Fang, Yuanyuan Dang, An’an Ping, Chenyu Wang, Qianchuan Zhao, Hulin
  Zhao, Xiaoli Li, and Mingsha Zhang.
\newblock Human high-order thalamic nuclei gate conscious perception through
  the thalamofrontal loop.
\newblock \emph{Science}, 388\penalty0 (6742):\penalty0 eadr3675, 2025.

\bibitem[Farquhar et~al.(2024)Farquhar, Kossen, Kuhn, and
  Gal]{farquhar2024detecting}
Sebastian Farquhar, Jannik Kossen, Lorenz Kuhn, and Yarin Gal.
\newblock Detecting hallucinations in large language models using semantic
  entropy.
\newblock \emph{Nature}, 630\penalty0 (8017):\penalty0 625--630, 2024.

\bibitem[Fudenberg and Tirole(1991)]{fudenberg1991game}
Drew Fudenberg and Jean Tirole.
\newblock \emph{Game theory}.
\newblock MIT press, 1991.

\bibitem[Genest(1984)]{genest1984characterization}
Christian Genest.
\newblock A characterization theorem for externally bayesian groups.
\newblock \emph{The Annals of Statistics}, 12\penalty0 (3):\penalty0
  1100--1105, 1984.

\bibitem[Genest and Zidek(1986)]{genest1986combining}
Christian Genest and James~V. Zidek.
\newblock Combining probability distributions: A critique and an annotated
  bibliography.
\newblock \emph{Statistical Science}, 1\penalty0 (1):\penalty0 114--135, 1986.

\bibitem[Genest et~al.(1986)Genest, McConway, and
  Schervish]{genest1986characterization}
Christian Genest, Kevin~J. McConway, and Mark~J. Schervish.
\newblock Characterization of externally bayesian pooling operators.
\newblock \emph{The Annals of Statistics}, 14\penalty0 (2):\penalty0 487--501,
  1986.

\bibitem[Geva et~al.(2023)Geva, Bastings, Filippova, and
  Globerson]{geva2023dissecting}
Mor Geva, Jasmijn Bastings, Katja Filippova, and Amir Globerson.
\newblock Dissecting recall of factual associations in auto-regressive language
  models.
\newblock \emph{arXiv preprint arXiv:2304.14767}, 2023.

\bibitem[Gneiting and Raftery(2007)]{gneiting2007strictly}
Tilmann Gneiting and Adrian~E Raftery.
\newblock Strictly proper scoring rules, prediction, and estimation.
\newblock \emph{Journal of the American Statistical Association}, 102\penalty0
  (477):\penalty0 359--378, 2007.

\bibitem[G{\"o}del(1931)]{godel1931formal}
Kurt G{\"o}del.
\newblock {{\"U}ber formal unentscheidbare S{\"a}tze der Principia Mathematica
  und verwandter Systeme I}.
\newblock \emph{Monatshefte f{\"u}r Mathematik und Physik}, 38:\penalty0
  173--198, 1931.

\bibitem[G{\"o}del(1932)]{godel1932vollstandigkeit}
Kurt G{\"o}del.
\newblock {{\"U}ber Vollst{\"a}ndigkeit und Widerspruchsfreiheit}.
\newblock \emph{Ergebnisse eines mathematischen Kolloquiums}, 3:\penalty0
  12--13, 1932.

\bibitem[Gottesman and Geva(2024)]{Gottesman2024EstimatingKI}
Daniela Gottesman and Mor Geva.
\newblock Estimating knowledge in large language models without generating a
  single token.
\newblock \emph{ArXiv}, abs/2406.12673, 2024.

\bibitem[Green and Laffont(1977)]{green1977characterization}
Jerry Green and Jean-Jacques Laffont.
\newblock Characterization of satisfactory mechanisms for the revelation of
  preferences for public goods.
\newblock \emph{Econometrica: Journal of the Econometric Society}, 45\penalty0
  (2):\penalty0 427--438, 1977.

\bibitem[Groves(1973)]{Groves73}
Theodore Groves.
\newblock Incentives in teams.
\newblock \emph{Econometrica}, 41\penalty0 (4):\penalty0 617--631, 1973.

\bibitem[Heisenberg(1927)]{heisenberg1927anschaulichen}
Werner Heisenberg.
\newblock {{\"U}ber den anschaulichen Inhalt der quantentheoretischen Kinematik
  und Mechanik}.
\newblock \emph{{Zeitschrift f{\"u}r Physik}}, 43\penalty0 (3--4):\penalty0
  172--198, 1927.

\bibitem[Hinton(2002)]{hinton2002training}
Geoffrey~E. Hinton.
\newblock Training products of experts by minimizing contrastive divergence.
\newblock \emph{Neural Computation}, 14\penalty0 (8):\penalty0 1771--1800,
  2002.
\newblock \doi{10.1162/089976602760128018}.

\bibitem[Holmstr\"om(1979)]{holmstrom1979groves}
Bengt Holmstr\"om.
\newblock Groves' scheme on restricted domains.
\newblock \emph{Econometrica}, 47\penalty0 (5):\penalty0 1137--1144, 1979.
\newblock \doi{10.2307/1911954}.

\bibitem[Huang et~al.(2025)Huang, Yu, Ma, Zhong, Feng, Wang, Chen, Peng, Feng,
  Qin, and Liu]{huang25hallsurvey}
Lei Huang, Weijiang Yu, Weitao Ma, Weihong Zhong, Zhangyin Feng, Haotian Wang,
  Qianglong Chen, Weihua Peng, Xiaocheng Feng, Bing Qin, and Ting Liu.
\newblock A survey on hallucination in large language models: Principles,
  taxonomy, challenges, and open questions.
\newblock \emph{ACM Trans. Inf. Syst.}, 43\penalty0 (2), January 2025.
\newblock ISSN 1046-8188.
\newblock \doi{10.1145/3703155}.

\bibitem[Ji et~al.(2023)Ji, Lee, Frieske, Yu, Su, Xu, Ishii, Bang, Madotto, and
  Fung]{Ji_2023}
Ziwei Ji, Nayeon Lee, Rita Frieske, Tiezheng Yu, Dan Su, Yan Xu, Etsuko Ishii,
  Ye~Jin Bang, Andrea Madotto, and Pascale Fung.
\newblock Survey of hallucination in natural language generation.
\newblock \emph{ACM Computing Surveys}, 55\penalty0 (12):\penalty0 1–38,
  March 2023.
\newblock ISSN 1557-7341.
\newblock \doi{10.1145/3571730}.
\newblock URL \url{http://dx.doi.org/10.1145/3571730}.

\bibitem[Kahneman(2011)]{Kahneman:2011fj}
Daniel Kahneman.
\newblock \emph{Thinking, Fast and Slow}.
\newblock Farrar, Straus and Giroux, New York, 2011.

\bibitem[Kalai and Vempala(2024)]{kalai2024calibrated}
Adam~Tauman Kalai and Santosh~S Vempala.
\newblock Calibrated language models must hallucinate.
\newblock In \emph{Proceedings of the 56th Annual ACM Symposium on Theory of
  Computing}, pages 160--171, 2024.

\bibitem[Kalai et~al.(2026)Kalai, Nachum, Vempala, and Zhang]{kalai2025why}
Adam~Tauman Kalai, Ofir Nachum, Santosh~S. Vempala, and Edwin Zhang.
\newblock Evaluating large language models for accuracy incentivizes
  hallucinations.
\newblock \emph{Nature}, 653\penalty0 (8116):\penalty0 1047--1051, 2026.
\newblock \doi{10.1038/s41586-026-10549-w}.

\bibitem[Karbasi et~al.(2025)Karbasi, Montasser, Sous, and
  Velegkas]{karbasi2025impossibility}
Amin Karbasi, Omar Montasser, John Sous, and Grigoris Velegkas.
\newblock (im)possibility of automated hallucination detection in large
  language models, 2025.
\newblock URL \url{https://arxiv.org/abs/2504.17004}.

\bibitem[Karpowicz(2012)]{DBLP:journals/amcs/Karpowicz12}
Michal~P. Karpowicz.
\newblock Nash equilibrium design and price-based coordination in hierarchical
  systems.
\newblock \emph{Int. J. Appl. Math. Comput. Sci.}, 22\penalty0 (4):\penalty0
  951--969, 2012.
\newblock \doi{10.2478/V10006-012-0071-0}.
\newblock URL \url{https://doi.org/10.2478/v10006-012-0071-0}.

\bibitem[Karpowicz(2011)]{karpowicz_designing_2011}
Michał~P. Karpowicz.
\newblock Designing auctions: a historical perspective.
\newblock \emph{Journal of Telecommunications and Information Technology},
  3:\penalty0 114--122, 2011.
\newblock Place: Warsaw, Poland.

\bibitem[Karpowicz(2021)]{karpowicz_theory_2021}
Michał~P. Karpowicz.
\newblock A theory of meta-factorization.
\newblock \emph{ArXiv}, abs/2111.14385, 2021.

\bibitem[Karpowicz and Strang(2023)]{karpowicz2024pseudoinverseacrarc}
Michał~P. Karpowicz and Gilbert Strang.
\newblock {The Pseudoinverse of $A=CR$ is $A^+=R^+C^+$ (?)}, 2023.
\newblock URL \url{https://arxiv.org/abs/2305.01716}.

\bibitem[Katharopoulos et~al.(2020)Katharopoulos, Vyas, Pappas, and
  Fleuret]{katharopoulos2020transformersrnnsfastautoregressive}
Angelos Katharopoulos, Apoorv Vyas, Nikolaos Pappas, and François Fleuret.
\newblock Transformers are rnns: Fast autoregressive transformers with linear
  attention, 2020.
\newblock URL \url{https://arxiv.org/abs/2006.16236}.

\bibitem[Kechris(1995)]{Kechris1995}
Alexander~S. Kechris.
\newblock \emph{Classical Descriptive Set Theory}.
\newblock Graduate Texts in Mathematics. Springer, New York, 1995.
\newblock ISBN 978-0-387-94374-9.

\bibitem[Kleene(1952)]{kleene1952metamathematics}
Stephen~C. Kleene.
\newblock \emph{Introduction to Metamathematics}.
\newblock North-Holland, Amsterdam, 1952.

\bibitem[Kolmogorov(1965)]{kolmogorov1965three}
Andrei~N Kolmogorov.
\newblock Three approaches to the quantitative definition of information.
\newblock \emph{Problems of Information Transmission}, 1\penalty0 (1):\penalty0
  1--7, 1965.
\newblock English reprint: International Journal of Computer Mathematics,
  2:157--168, 1968.

\bibitem[Krishna(2009)]{krishna2009auction}
Vijay Krishna.
\newblock \emph{Auction theory}.
\newblock Academic press, 2009.

\bibitem[Kuratowski(1966)]{kuratowski1966wstep}
Kazimierz Kuratowski.
\newblock \emph{Wstep do teorii mnogo{\'s}ci i topologii}.
\newblock PWN, 1966.

\bibitem[Laffont and Maskin(1980)]{laffont1980differential}
Jean-Jacques Laffont and Eric Maskin.
\newblock A differential approach to dominant strategy mechanisms.
\newblock \emph{Econometrica}, 48\penalty0 (6):\penalty0 1507--1520, 1980.
\newblock \doi{10.2307/1912821}.

\bibitem[Levine(2001)]{levine2001purple}
Joseph Levine.
\newblock \emph{Purple haze: The puzzle of consciousness}.
\newblock Oxford University Press, 2001.

\bibitem[Lewis et~al.(2021)Lewis, Perez, Piktus, Petroni, Karpukhin, Goyal,
  Küttler, Lewis, tau Yih, Rocktäschel, Riedel, and
  Kiela]{lewis2021retrievalaugmentedgenerationknowledgeintensivenlp}
Patrick Lewis, Ethan Perez, Aleksandra Piktus, Fabio Petroni, Vladimir
  Karpukhin, Naman Goyal, Heinrich Küttler, Mike Lewis, Wen tau Yih, Tim
  Rocktäschel, Sebastian Riedel, and Douwe Kiela.
\newblock Retrieval-augmented generation for knowledge-intensive nlp tasks,
  2021.
\newblock URL \url{https://arxiv.org/abs/2005.11401}.

\bibitem[Li et~al.(2023)Li, Cheng, Zhao, Nie, and
  Wen]{li2023haluevallargescalehallucinationevaluation}
Junyi Li, Xiaoxue Cheng, Wayne~Xin Zhao, Jian-Yun Nie, and Ji-Rong Wen.
\newblock Halueval: A large-scale hallucination evaluation benchmark for large
  language models, 2023.
\newblock URL \url{https://arxiv.org/abs/2305.11747}.

\bibitem[Lin et~al.(2022)Lin, Hilton, and
  Evans]{lin2022truthfulqameasuringmodelsmimic}
Stephanie Lin, Jacob Hilton, and Owain Evans.
\newblock Truthfulqa: Measuring how models mimic human falsehoods, 2022.
\newblock URL \url{https://arxiv.org/abs/2109.07958}.

\bibitem[Lindsey et~al.(2025)Lindsey, Gurnee, Ameisen, Chen, Pearce, Turner,
  Citro, Abrahams, Carter, Hosmer, et~al.]{lindsey2025biology}
Jack Lindsey, Wes Gurnee, Emmanuel Ameisen, Brian Chen, Adam Pearce, Nicholas~L
  Turner, Craig Citro, David Abrahams, Shan Carter, Basil Hosmer, et~al.
\newblock On the biology of a large language model.
\newblock \emph{Transformer Circuits Thread}, 2025.

\bibitem[Mason and Anand(2026)]{mason2026epistemic}
Tony Mason and Vaastav Anand.
\newblock Epistemic observability in language models, 2026.
\newblock URL \url{https://arxiv.org/abs/2603.20531}.

\bibitem[Maynez et~al.(2020)Maynez, Narayan, Bohnet, and
  McDonald]{maynez2020faithfulness}
Joshua Maynez, Shashi Narayan, Bernd Bohnet, and Ryan McDonald.
\newblock On faithfulness and factuality in abstractive summarization.
\newblock In \emph{Proceedings of the 58th Annual Meeting of the Association
  for Computational Linguistics}, pages 1906--1919, 2020.
\newblock \doi{10.18653/v1/2020.acl-main.173}.

\bibitem[Meng et~al.(2023)Meng, Bau, Andonian, and
  Belinkov]{meng2023locatingeditingfactualassociations}
Kevin Meng, David Bau, Alex Andonian, and Yonatan Belinkov.
\newblock Locating and editing factual associations in gpt, 2023.
\newblock URL \url{https://arxiv.org/abs/2202.05262}.

\bibitem[Milgrom(2004)]{Milgrom_2004}
Paul Milgrom.
\newblock \emph{Putting Auction Theory to Work}.
\newblock Churchill Lectures in Economics. Cambridge University Press, 2004.

\bibitem[Milgrom and Weber(1982)]{milgrom1982theory}
Paul~R Milgrom and Robert~J Weber.
\newblock A theory of auctions and competitive bidding.
\newblock \emph{Econometrica: Journal of the Econometric Society}, 50\penalty0
  (5):\penalty0 1089--1122, 1982.

\bibitem[Myerson(1989)]{Myerson1989}
Roger~B. Myerson.
\newblock \emph{Mechanism Design}, pages 191--206.
\newblock Palgrave Macmillan UK, London, 1989.
\newblock ISBN 978-1-349-20215-7.
\newblock \doi{10.1007/978-1-349-20215-7_20}.
\newblock URL \url{https://doi.org/10.1007/978-1-349-20215-7_20}.

\bibitem[Myerson(2008)]{myerson2008perspectives}
Roger~B Myerson.
\newblock Perspectives on mechanism design in economic theory.
\newblock \emph{American Economic Review}, 98\penalty0 (3):\penalty0 586--603,
  2008.

\bibitem[Nisan et~al.(2007)Nisan, Roughgarden, Tardos, and
  Vazirani]{Nisan_2007}
Noam Nisan, Tim Roughgarden, Eva Tardos, and Vijay~V. Vazirani, editors.
\newblock \emph{Algorithmic Game Theory}.
\newblock Cambridge University Press, Cambridge, 2007.

\bibitem[Ouyang et~al.(2022)Ouyang, Wu, Jiang, Almeida, Wainwright, Mishkin,
  Zhang, Agarwal, Slama, Ray, Schulman, Hilton, Kelton, Miller, Simens, Askell,
  Welinder, Christiano, Leike, and
  Lowe]{ouyang2022traininglanguagemodelsfollow}
Long Ouyang, Jeff Wu, Xu~Jiang, Diogo Almeida, Carroll~L. Wainwright, Pamela
  Mishkin, Chong Zhang, Sandhini Agarwal, Katarina Slama, Alex Ray, John
  Schulman, Jacob Hilton, Fraser Kelton, Luke Miller, Maddie Simens, Amanda
  Askell, Peter Welinder, Paul Christiano, Jan Leike, and Ryan Lowe.
\newblock Training language models to follow instructions with human feedback,
  2022.
\newblock URL \url{https://arxiv.org/abs/2203.02155}.

\bibitem[Ranjan and Gneiting(2010)]{ranjan2010combining}
Roopesh Ranjan and Tilmann Gneiting.
\newblock Combining probability forecasts.
\newblock \emph{Journal of the Royal Statistical Society: Series B},
  72\penalty0 (1):\penalty0 71--91, 2010.
\newblock \doi{10.1111/j.1467-9868.2009.00726.x}.

\bibitem[Rashkin et~al.(2023)Rashkin, Nikolaev, Lamm, Aroyo, Collins, Das,
  Petrov, Tomar, Turc, and Reitter]{rashkin2023measuring}
Hannah Rashkin, Vitaly Nikolaev, Matthew Lamm, Lora Aroyo, Michael Collins,
  Dipanjan Das, Slav Petrov, Gaurav~Singh Tomar, Iulia Turc, and David Reitter.
\newblock Measuring attribution in natural language generation models.
\newblock \emph{Computational Linguistics}, 49\penalty0 (4):\penalty0 777--840,
  2023.
\newblock \doi{10.1162/coli_a_00490}.

\bibitem[Roughgarden and Talgam-Cohen(2016)]{roughgarden2016optimal}
Tim Roughgarden and Inbal Talgam-Cohen.
\newblock Optimal and robust mechanism design with interdependent values.
\newblock \emph{ACM Transactions on Economics and Computation (TEAC)},
  4\penalty0 (3):\penalty0 1--34, 2016.

\bibitem[Savage(1971)]{savage1971elicitation}
Leonard~J Savage.
\newblock Elicitation of personal probabilities and expectations.
\newblock \emph{Journal of the American Statistical Association}, 66\penalty0
  (336):\penalty0 783--801, 1971.

\bibitem[Seth(2021)]{seth2021being}
Anil Seth.
\newblock \emph{Being you: A new science of consciousness}.
\newblock Penguin, 2021.

\bibitem[Seth and Bayne(2022)]{Seth2021TheoriesOC}
Anil~K. Seth and Tim Bayne.
\newblock Theories of consciousness.
\newblock \emph{Nature Reviews Neuroscience}, 23\penalty0 (7):\penalty0
  439--452, 2022.

\bibitem[Sharkey et~al.(2025)Sharkey, Chughtai, Batson, Lindsey, Wu, Bushnaq,
  Goldowsky-Dill, Heimersheim, Ortega, Bloom, Biderman, Garriga-Alonso, Conmy,
  Nanda, Rumbelow, Wattenberg, Schoots, Miller, Michaud, Casper, Tegmark,
  Saunders, Bau, Todd, Geiger, Geva, Hoogland, Murfet, and
  McGrath]{Sharkey2025OpenPI}
Lee Sharkey, Bilal Chughtai, Joshua Batson, Jack Lindsey, Jeff Wu, Lucius
  Bushnaq, Nicholas Goldowsky-Dill, Stefan Heimersheim, Alejandro Ortega,
  Joseph Bloom, Stella Biderman, Adri{\`a} Garriga-Alonso, Arthur Conmy, Neel
  Nanda, Jessica Rumbelow, Martin Wattenberg, Nandi Schoots, Joseph Miller,
  Eric~J. Michaud, Stephen Casper, Max Tegmark, William Saunders, David Bau,
  Eric Todd, Atticus Geiger, Mor Geva, Jesse Hoogland, Daniel Murfet, and Tom
  McGrath.
\newblock Open problems in mechanistic interpretability.
\newblock \emph{ArXiv}, abs/2501.16496, 2025.

\bibitem[Shelmanov et~al.(2025)Shelmanov, Fadeeva, Tsvigun, Tsvigun, Xie,
  Kiselev, Daheim, Zhang, Vazhentsev, Sachan, Nakov, and
  Baldwin]{shelmanov2025headpredictheadquestion}
Artem Shelmanov, Ekaterina Fadeeva, Akim Tsvigun, Ivan Tsvigun, Zhuohan Xie,
  Igor Kiselev, Nico Daheim, Caiqi Zhang, Artem Vazhentsev, Mrinmaya Sachan,
  Preslav Nakov, and Timothy Baldwin.
\newblock A head to predict and a head to question: Pre-trained uncertainty
  quantification heads for hallucination detection in llm outputs, 2025.
\newblock URL \url{https://arxiv.org/abs/2505.08200}.

\bibitem[Sun et~al.(2024{\natexlab{a}})Sun, Sheng, Zhou, and Wu]{Sun2024AIHT}
Yujie Sun, Dongfang Sheng, Zihan Zhou, and Yifei Wu.
\newblock Ai hallucination: towards a comprehensive classification of distorted
  information in artificial intelligence-generated content.
\newblock \emph{Humanities and Social Sciences Communications}, 11:\penalty0
  1278, 2024{\natexlab{a}}.

\bibitem[Sun et~al.(2024{\natexlab{b}})Sun, Zang, Zheng, Xu, Zhang, Yu, Song,
  and Li]{Sun2024ReDeEPDH}
Zhongxiang Sun, Xiaoxue Zang, Kai Zheng, Jun Xu, Xiao Zhang, Weijie Yu, Yang
  Song, and Han Li.
\newblock Redeep: Detecting hallucination in retrieval-augmented generation via
  mechanistic interpretability.
\newblock \emph{ArXiv}, abs/2410.11414, 2024{\natexlab{b}}.

\bibitem[Sundararajan et~al.(2017)Sundararajan, Taly, and
  Yan]{sundararajan2017axiomatic}
Mukund Sundararajan, Ankur Taly, and Qiqi Yan.
\newblock Axiomatic attribution for deep networks.
\newblock In \emph{Proceedings of the 34th International Conference on Machine
  Learning (ICML)}, pages 3319--3328, 2017.

\bibitem[Suzuki et~al.(2025)Suzuki, He, Tian, and
  Wang]{suzuki2025hallucinations}
Atsushi Suzuki, Yulan He, Feng Tian, and Zhongyuan Wang.
\newblock Hallucinations are inevitable but can be made statistically
  negligible.
\newblock \emph{arXiv preprint arXiv:2502.12187}, 2025.

\bibitem[Tarski(1955)]{Tarski1955}
Alfred Tarski.
\newblock A lattice-theoretical fixpoint theorem and its applications.
\newblock \emph{Pacific Journal of Mathematics}, 5\penalty0 (2):\penalty0
  285--309, 1955.

\bibitem[Thaler(2015)]{thaler2015misbehaving}
Richard~H Thaler.
\newblock \emph{Misbehaving: The making of behavioral economics}.
\newblock WW Norton \& Company, 2015.

\bibitem[Tonmoy et~al.(2024)Tonmoy, Zaman, Jain, Rani, Rawte, Chadha, and
  Das]{Tonmoy2024ACS}
S.~M Towhidul~Islam Tonmoy, S~M~Mehedi Zaman, Vinija Jain, Anku Rani, Vipula
  Rawte, Aman Chadha, and Amitava Das.
\newblock A comprehensive survey of hallucination mitigation techniques in
  large language models.
\newblock \emph{ArXiv}, abs/2401.01313, 2024.

\bibitem[Turing(1937)]{turing1936computable}
Alan~Mathison Turing.
\newblock On computable numbers, with an application to the
  {E}ntscheidungsproblem.
\newblock \emph{Proceedings of the London Mathematical Society}, s2-42\penalty0
  (1):\penalty0 230--265, 1937.
\newblock \doi{10.1112/plms/s2-42.1.230}.

\bibitem[Vaswani et~al.(2017)Vaswani, Shazeer, Parmar, Uszkoreit, Jones, Gomez,
  Kaiser, and Polosukhin]{vaswani_attention_2017}
Ashish Vaswani, Noam Shazeer, Niki Parmar, Jakob Uszkoreit, Llion Jones,
  Aidan~N. Gomez, Lukasz Kaiser, and Illia Polosukhin.
\newblock Attention is {All} you {Need}.
\newblock In \emph{Neural {Information} {Processing} {Systems}}, 2017.
\newblock URL \url{https://api.semanticscholar.org/CorpusID:13756489}.

\bibitem[Waters et~al.(2016)Waters, Blom, Dang-Vu, Cheyne, Alderson-Day,
  Woodruff, and Collerton]{Waters2016WhatIT}
Flavie Waters, Jan~Dirk Blom, Thien~Thanh Dang-Vu, Allan~J. Cheyne, Ben
  Alderson-Day, Peter Woodruff, and Daniel Collerton.
\newblock What is the link between hallucinations, dreams, and
  hypnagogic-hypnopompic experiences?
\newblock \emph{Schizophrenia Bulletin}, 42\penalty0 (5):\penalty0 1098--1109,
  2016.

\bibitem[Wei et~al.(2022)Wei, Wang, Schuurmans, Bosma, Ichter, Xia, Chi, Le,
  and Zhou]{Wei2022ChainOT}
Jason Wei, Xuezhi Wang, Dale Schuurmans, Maarten Bosma, Brian Ichter, Fei Xia,
  Ed~H. Chi, Quoc~V. Le, and Denny Zhou.
\newblock Chain-of-thought prompting elicits reasoning in large language
  models.
\newblock \emph{ArXiv}, abs/2201.11903, 2022.

\bibitem[Wu et~al.(2024)Wu, Grama, and Szpankowski]{Wu2024NoFL}
Changlong Wu, Ananth Grama, and Wojciech Szpankowski.
\newblock No free lunch: Fundamental limits of learning non-hallucinating
  generative models.
\newblock \emph{ArXiv}, abs/2410.19217, 2024.

\bibitem[Xu et~al.(2024{\natexlab{a}})Xu, Cruz, Guevara, Wang, Deshpande, Wang,
  and Li]{Xu2024RetrievalAugmentedGW}
Zhentao Xu, Mark~Jerome Cruz, Matthew Guevara, Tie Wang, Manasi Deshpande,
  Xiaofeng Wang, and Zheng Li.
\newblock Retrieval-augmented generation with knowledge graphs for customer
  service question answering.
\newblock \emph{Proceedings of the 47th International ACM SIGIR Conference on
  Research and Development in Information Retrieval}, pages 2905--2909,
  2024{\natexlab{a}}.

\bibitem[Xu et~al.(2024{\natexlab{b}})Xu, Jain, and
  Kankanhalli]{Xu2024HallucinationII}
Ziwei Xu, Sanjay Jain, and Mohan Kankanhalli.
\newblock Hallucination is inevitable: An innate limitation of large language
  models.
\newblock \emph{ArXiv}, abs/2401.11817, 2024{\natexlab{b}}.

\bibitem[Yanofsky(2016)]{yanofsky2016outer}
Noson~S Yanofsky.
\newblock \emph{The outer limits of reason: What science, mathematics, and
  logic cannot tell us}.
\newblock MIT Press, 2016.

\bibitem[Yu et~al.(2024)Yu, Cao, Cheung, and
  Dong]{yu2024mechanisticunderstandingmitigationlanguage}
Lei Yu, Meng Cao, Jackie Chi~Kit Cheung, and Yue Dong.
\newblock Mechanistic understanding and mitigation of language model
  non-factual hallucinations, 2024.
\newblock URL \url{https://arxiv.org/abs/2403.18167}.

\end{thebibliography}
\end{document}